\documentclass[journal]{IEEEtran}
\usepackage{cite}
\ifCLASSINFOpdf
\else
\fi

\usepackage{dsfont}
\usepackage{enumerate}
\usepackage[inline]{enumitem}
\usepackage{graphicx}
\usepackage{multirow}
\usepackage{amsmath}
\usepackage{xcolor}

\usepackage{tablefootnote}
\usepackage{mathrsfs}
\usepackage{soul}

\begin{document}
\bstctlcite{IEEEexample:BSTcontrol}
%
% paper title
% Titles are generally capitalized except for words such as a, an, and, as,
% at, but, by, for, in, nor, of, on, or, the, to and up, which are usually
% not capitalized unless they are the first or last word of the title.
% Linebreaks \\ can be used within to get better formatting as desired.
% Do not put math or special symbols in the title.
\title{A Survey on Multi-output Learning}
%
%
% author names and IEEE memberships
% note positions of commas and nonbreaking spaces ( ~ ) LaTeX will not break
% a structure at a ~ so this keeps an author's name from being broken across
% two lines.
% use \thanks{} to gain access to the first footnote area
% a separate \thanks must be used for each paragraph as LaTeX2e's \thanks
% was not built to handle multiple paragraphs
%

\author{Donna~Xu,
        Yaxin~Shi,
        Ivor~W.~Tsang,
        Yew-Soon~Ong,
        Chen~Gong,
        and~Xiaobo~Shen% <-this % stops a space
\thanks{D. Xu, Y. Shi and I. W. Tsang are with the Centre for Artificial Intelligence, FEIT, University of Technology Sydney, Ultimo, NSW 2007, Australia (email: doxu2620@gmail.com, yaxin.shi@student.uts.edu.au, ivor.tsang@uts.edu.au).}% <-this % stops a space
\thanks{Y.-S. Ong is with the Data Science \& Artificial Intelligence Research Centre, SCSE, Nanyang Technological University, Singapore 639798 (email: asysong@ntu.edu.sg).}% <-this % stops a space
\thanks{C. Gong and X. Shen are with the School of Computer Science and Engineering, Nanjing University of Science and Technology, Nanjing 210094, China (email: chen.gong@njust.edu.cn, njust.shenxiaobo@gmail.com).}
\thanks{This research is supported by ARC grant LP150100671 and DP180100106, CSC (No: 201706330075), NRFS under its AI Singapore Programme (AISG-RP-2018-004), NSF of China (No: 61602246, 61973162), NSF of Jiangsu Province (No: BK20171430), the FRF for the Central Universities (No: 30918011319), the “Summit of the Six Top Talents” Program (No: DZXX-027), the “Young Elite Scientists Sponsorship Program” by Jiangsu Province, the “Young Elite Scientists Sponsorship Program” by CAST (No: 2018QNRC001), the NNSFC under Grant No. 61906091, the NSF of Jiangsu Province, China (Youth Fund Project) under Grant No. BK20190440, and the FRF for the Central Universities under Grant No. 30919011229.}}

\maketitle

% As a general rule, do not put math, special symbols or citations
% in the abstract or keywords.
\begin{abstract}
The aim of multi-output learning is to simultaneously predict multiple outputs given an input. It is an important learning problem for decision-making, since making decisions in the real world often involves multiple complex factors and criteria. In recent times, an increasing number of research studies have focused on ways to predict multiple outputs at once. Such efforts have transpired in different forms according to the particular multi-output learning problem under study. Classic cases of multi-output learning include multi-label learning, multi-dimensional learning, multi-target regression and others. From our survey of the topic, we were struck by a lack in studies that generalize the different forms of multi-output learning into a common framework. This paper fills that gap with a comprehensive review and analysis of the multi-output learning paradigm. In particular, we characterize the 4 Vs of multi-output learning, i.e., volume, velocity, variety, and veracity, and the ways in which the 4 Vs both benefit and bring challenges to multi-output learning by taking inspiration from big data. We analyze the life cycle of output labeling, present the main mathematical definitions of multi-output learning, and examine the field's key challenges and corresponding solutions as found in the literature. Several model evaluation metrics and popular data repositories are also discussed. Last but not least, we highlight some emerging challenges with multi-output learning from the perspective of the 4 Vs as potential research directions worthy of further studies.
	
\end{abstract}

% Note that keywords are not normally used for peerreview papers.

% For peer review papers, you can put extra information on the cover
% page as needed:
% \ifCLASSOPTIONpeerreview
% \begin{center} \bfseries EDICS Category: 3-BBND \end{center}
% \fi
%
% For peerreview papers, this IEEEtran command inserts a page break and
% creates the second title. It will be ignored for other modes.
\IEEEpeerreviewmaketitle

\section{Introduction}

\IEEEPARstart{T}{raditional} supervised learning is one of the most well established and adopted machine learning paradigms. It offers fast and accurate predictions for today's real-world smart systems and applications. The goal of traditional supervised learning is to learn a function that maps each of the given inputs to a corresponding known output. For prediction tasks, the output comes in the form of a single label. For regression tasks, it is a single value. Traditional supervised learning has been shown to be good at solving these simple single-output problems – classical examples being binary classification, such as filtering spam in an email system, or a regression problem where the daily energy consumption of a machine needs to be predicted based on temperature, wind speed, humidity levels, etc.

However, the traditional supervised learning paradigm is not coping well with the increasing needs of today's complex decision making. As a result, there is a pressing need for new machine learning paradigms. Here, multi-output learning has emerged as a solution. The aim is to simultaneously predict multiple outputs given a single input, which means it is possible to solve far more complex decision-making problems. Compared to traditional single-output learning, multi-output learning is multi-variate nature, and the outputs may have complex interactions that can only be handled by structured inference. Additionally, the potentially diverse data types of the outputs has led to various categories of machine learning problems and corresponding subfields of study. For example, binary output values relate to multi-label classification problems \cite{DBLP:journals/tkde/ZhangZ14,DBLP:conf/aaai/GongTYL16}; nominal output values relate to multi-dimensional classification problems \cite{DBLP:journals/ijar/BielzaLL11}; ordinal output values are studied in label ranking problems \cite{DBLP:books/daglib/p/VembuG10}; and real-valued outputs are considered in multi-target regression problems \cite{DBLP:journals/widm/BorchaniVBL15}. 

Together, all these problems constitute the multi-output paradigm, and the body of literature surrounding this field has grown rapidly. Several works have been presented that provide a comprehensive review of the emerging challenges and learning algorithms in each subfield. For instance, Zhang and Zhou \cite{DBLP:journals/tkde/ZhangZ14} studied the emerging area of multi-label learning;  Borchani et al. \cite{DBLP:journals/widm/BorchaniVBL15} summarized the increasing problems in multi-target regression; and \cite{DBLP:books/daglib/p/VembuG10} Vembu and Gartner presented a review on multi-label ranking. However, little attention has been paid to the global picture of multi-output learning and the importance of the output labels (Section.~\ref{sec:life_cycle_of_label}).
In addition, although the problems in each subfield seem distinctive due to the differences in their output structures (Section.~\ref{sec:structures}), they do share common traits (Section.~\ref{sec:mechanism}) and encounter common challenges brought by the characteristics of the output labels. In this paper, we attempt to provide such a view.

\subsection{The 4 Vs Challenges of Multiple Outputs}
The popular 4 Vs, i.e., \emph{volume}, \emph{velocity}, \emph{variety} and \emph{veracity}, have been well established as the main characteristics of big data. When scholars discuss the 4 Vs in multi-output learning scenarios, they are usually referring to input data; however, the 4 Vs can also be used to describe output labels. Moreover, these 4 Vs bring with them a set of challenges to multi-output learning processes, explained as follows.
\begin{enumerate}
\item Volume refers to explosive growth in output labels, which poses many challenges to multi-output learning. First, output label spaces can grow extremely large, which causes scalability issues. Second, the burden for label annotators is significantly increased and still there are often insufficient annotations in a dataset to adequately train a model. In turn, this may lead to unseen outputs during testing. Third, volume may pose label imbalance issues, especially if not all the generated labels in a dataset have sufficient data instances (inputs).
\item Velocity refers to how rapidly output labels are acquired, which includes the phenomenon of concept drift \cite{widmer1996learning}. Velocity can present challenges due to changes in output distributions, where the target outputs vary over time in unforeseen ways. 
\item Variety refers to the heterogeneous nature of output labels. Output labels are gathered from multiple sources and are of various data formats with different structures. In particular, output labels with complex structures can create multiple challenges in multi-output learning, such as finding an appropriate method of modeling output dependencies, or how to design a multi-variate loss function, or how to design efficient algorithms.
\item Veracity refers to differences in the quality of the output labels. Issues such as noise, missing values, abnormalities, or incomplete data are all characteristics of veracity. 
\end{enumerate}

%\textcolor{blue}{Consequently, we give an overview for multi-output learning based these 4V challenges. We first illustrate the life-cycle of output labels help understand the 4Vs challenges. Then, we summarize the subfields of multi-output learning based on the ``Variety" of the output labels and provide problem definition for each subfield within a unified paradigm. We also give the commonly used datasets and metrics for representative subfields. The emerging challenges caused by the 4Vs are then summarized and discussed with representative works.}

%With the emergence of big data, the explosive growth, various complex structures of output labels bring new challenges to the learning process of multi-output learning.

%we enter the era of big data. It brings new challenges to the learning process of multi-output learning.
%With the emergence of big data, there are a number of challenges encountered for multi-output learning.

%Big data can be defined as the popular 5Vs, volume, velocity, variety, veracity and value.
%The challenges imposed by each of the characteristics of big data are introduced in the following.
%The challenges imposed by big data complicates the learning process 

\subsection{Purpose and Organization of This Survey}
The goal of this paper is to provide a comprehensive overview of the multi-output learning paradigm using the 4 Vs as a frame for the current and future challenges facing this field of study. Multi-output learning has attracted significant attention from many machine learning disciplines, such as part-of-speech sequence tagging, language translation and natural language processing, motion tracking and optical character recognition in computer vision, document categorization and ranking in information retrieval, and so on. We expect this survey to deliver a complete picture of multi-output learning and a summation of the different problems being tackled across multiple communities. Ultimately, we hope to promote further development in multi-output learning, and inspire researchers to pursue worthy and needed future research directions.

The remainder of this survey is structured as follows. Section II illustrates the life cycle of output labels to help understand the challenges presented by the 4 Vs. Section III provides an overview of the myriad output structures along with definitions for the common subproblems addressed in multi-output learning. This section also includes some brief details on the common metrics and publicly-available data used when evaluating models. Section IV presents the challenges in multi-output learning presented by the 4 Vs and their corresponding representative works. Section V concludes the survey.

\section{Life Cycle of Output Labels}\label{sec:life_cycle_of_label}

\begin{figure}
	\centering
	\includegraphics[scale=0.40]{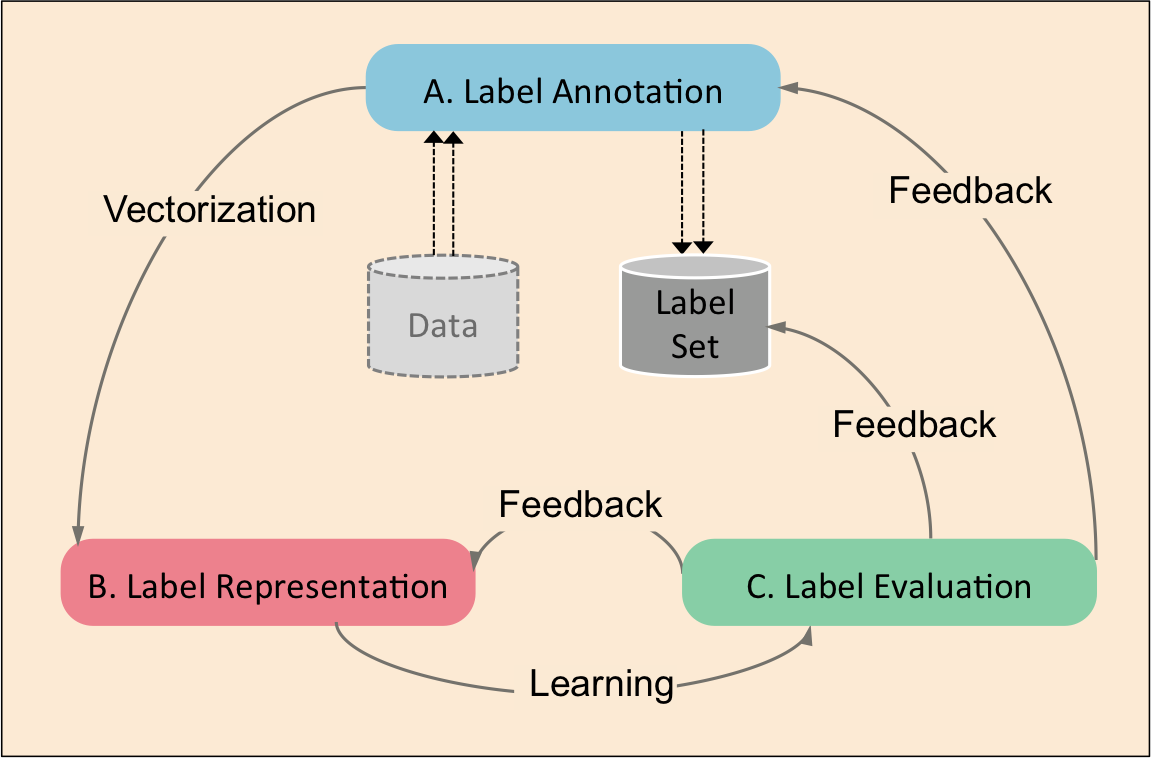}
	\caption{\label{label_life_cycle} The life cycle of the output label.}
\end{figure}

Output labels play an important role in multi-output learning tasks in that how well a model performs a task relies heavily on the quality of those labels. Fig.~\ref{label_life_cycle} depicts the three stages of a label's life cycle: annotation, representation, and evaluation. A brief overview of each stage follows along with the underlying issues that could potentially harm the effectiveness of multi-output learning systems.

\subsection{How is Data Labeled}
Label annotation requires a human to semantically annotate a piece of data and is a crucial step for training multi-output learning models. Data can be used directly with its basic annotations or, once labeled; they can be aggregated into sets for further analysis. Depending on the application and the task, label annotations come in various types. For example, the images for an image classification task should be labeled with tags or keywords, whereas a segmentation task would require each object in the images to be localized with a mask. A captioning task would require the images to be labeled with some textual descriptions, and so on.

Typically, creating large annotated datasets from scratch is time-consuming and labor-intensive no matter the annotation requirement. There are multiple ways to acquire labeled data. Social media provides a platform for researchers to search for labeled datasets - for example, Facebook and Flickr, which allow users to post pictures and comments with tags. Open-source collections, such as WordNet and Wikipedia, can also be useful sources of labeled datasets.

Beyond directly obtaining labeled datasets, crowdsourcing platforms like Amazon Mechanical Turk help researchers solicit labels for unlabeled datasets by recruiting online workers. The annotation type depends on the modeling task and, due to the efficiency of crowdsourcing, this method has quickly become a popular way of obtaining labeled datasets. ImageNet \cite{ILSVRC15} is a popular dataset that was labeled through a crowdsourcing platform. Its database of images is organized into a WordNet hierarchy, and it has been used to help researchers solve problems in a range of areas.

There are also many annotation tools that have been developed to annotate different types of data. LabelMe \cite{DBLP:journals/ijcv/RussellTMF08}, a web-based tool, provides users with a convenient way to label every object in an image and also correct labels annotated by other users. BRAT \cite{DBLP:conf/eacl/StenetorpPTOAT12} is also web-based but is specifically designed for natural language processing tasks, such as named-entity recognition and POS-tagging (part-of-speech tagging). TURKSENT \cite{DBLP:conf/acllaw/EryigitCYTC13} is an annotation tool to support sentiment analysis in social media posts.

\subsection{Forms of Label Representations}
There are many different types of label annotations for different tasks, such as tags, captions, masks, etc., and each type of annotation may have several representations, which are frequently represented as vectors. For example, the most common is the binary vector, whose size equals the vocabulary size of the tags. Annotated samples, e.g., samples with tags, are assigned with a value of 1 and the rest are given a 0. However, binary vectors are not optimal for more complex multi-output tasks because these representations do not preserve all useful information. Details like the semantics or the inherent structure are lost. To tackle this issue, alternative representation methods have been developed. For instance, real-valued vectors of tags \cite{DBLP:conf/cvpr/LampertNH09} indicate the strength and degree of the annotated tags using real values. Binary vectors of the associations between a tag's attributes have been used to convey the characteristics of tags. Hierarchical label embedding vectors \cite{DBLP:conf/cvpr/RohrbachSS11} capture the structure information in tags. Semantic word vectors, such as Word2Vec \cite{DBLP:conf/nips/MikolovSCCD13}, can be used to represent the semantics and/or context of tags and text descriptions. What is key in real-world multi-output applications is to select the label representation that is most appropriate for the given task. 

\subsection{Label Evaluation and Challenges}

Label evaluation is an essential step in guaranteeing the quality of labels and label representations. Thus, label evaluation plays a key role in the performance of multi-output tasks. Different models can be used to evaluate label quality: which to choose depends on the task. Generally, labels can be evaluated in three different respects: 
\begin{enumerate*}
	\item whether the annotation is of good quality (Step~A).
	\item whether the chosen label representation represents the labels well (Step B).
	\item whether the provided label set adequately covers the dataset (Label Set).
\end{enumerate*}
After evaluation, a human expert is generally required to explore any underlying issues and provide feedback to improve different aspects of the labels if needed.

\subsubsection{Issues of Label Annotation}
The aforementioned annotation methods, e.g., crowdsourcing, annotation tools, and social media, help researchers collect annotated data efficiently. But, without experts, these annotations methods are highly likely to result in the so-called noisy label problem, which includes both missing annotations and incorrect annotations. There are various reasons for noisy labels – for example, using crowdsourced workers that lack the required domain knowledge, social media users that include irrelevant tags with their image or post, or ambiguous text in a caption.

\subsubsection{Issues of Label Representation}
Output labels can also have internal structures and, often, this structure information is critical to the performance of the multi-output learning task at hand. Tag-based information retrieval~\cite{deerwester1989computer} and image captioning~\cite{xu2015show} are two examples where structure is crucial. However, incorporating this information into a representation as a labels is a non-trivial undertaking, as the data are usually many and domain knowledge is required to define their structure. In addition, the output label space might contain ambiguity. For example, a bag-of-words (BOW) is traditionally used as a representation of a label space in natural language processing tasks, but BOW contains word sense ambiguity, as two different words may have the same meaning and one word might refer to multiple meanings.

\subsubsection{Issues of the Label Set}
Constructing a label set for data annotation requires a human expert with domain knowledge. Plus, it is common that the provided label set does not contain sufficient labels for the data – perhaps due to fast data growth or the low occurrence of some labels. Therefore, there are likely to be unseen labels in the test data, which leads to open-set \cite{DBLP:journals/pami/ScheirerRSB13}, zero-shot \cite{DBLP:conf/nips/PalatucciPHM09} or concept drift \cite{DBLP:journals/pai/HoensPC12} problems.

\begin{table*}[]
\caption{A summary of subfields of multi-output learning and their corresponding output structures, applications and disciplines.}
\label{summary_of_subfields}
\resizebox{\textwidth}{!}{%
%\begin{scriptsize}
\begin{tabular}{|c|c|c|c|}
	\hline
	\textbf{Subfield} & \textbf{Output Structure} & \textbf{Application} & \textbf{Discipline} \\ \hline
	\multirow{3}{*}{Multi-label Learning} & \multirow{3}{*}{\begin{tabular}[c]{@{}c@{}}Independent \\ Binary Vector\end{tabular}} & Document Categorization \cite{DBLP:conf/cikm/CaiH04} & Natural Language Processing \\ \cline{3-4} 
	&  & Semantic Scene Classification \cite{DBLP:conf/cvpr/XiaoHEOT10} & Computer Vision \\ \cline{3-4} 
	&  & Automatic Video Annotation \cite{DBLP:conf/mm/QiHRTMZ07} & Computer Vision \\ \hline
	\multirow{3}{*}{Multi-target Regression} & \multirow{3}{*}{\begin{tabular}[c]{@{}c@{}}Independent \\ Real-valued\\ Vector\end{tabular}} & River Quality Prediction \cite{DBLP:journals/apin/DzeroskiDG00} & Ecology \\ \cline{3-4} 
	&  & Natural Gas Demand Forecasting \cite{aras2004forecasting} & Energy Meteorology \\ \cline{3-4} 
	&  & Drug Efficacy Prediction \cite{li2017novel} & Medicine \\ \hline
	\multirow{3}{*}{Label Distribution Learning} & \multirow{3}{*}{Distribution} & Head Pose Estimation \cite{DBLP:conf/cvpr/GengX14} & Computer Vision \\ \cline{3-4} 
	&  & Facial Age Estimation \cite{DBLP:conf/aaai/GengSZ10} & Computer Vision \\ \cline{3-4} 
	&  & Text Mining \cite{DBLP:conf/emnlp/ZhouZZZG16} & Data Mining \\ \hline
	\multirow{3}{*}{Label Ranking} & \multirow{3}{*}{Ranking} & Text Categorization Ranking \cite{DBLP:journals/jmlr/CrammerS03a} & Information Retrieval \\ \cline{3-4} 
	&  & Question Answering \cite{DBLP:conf/sigir/KoNS07} & Information Retrieval \\ \cline{3-4} 
	&  & Visual Object Recognition \cite{DBLP:conf/iccv/BucakMJJ09} & Computer Vision \\ \hline
	\multirow{3}{*}{Sequence Alignment Learning} & \multirow{3}{*}{Sequence} & Protein Function Prediction \cite{DBLP:conf/icml/LiuXC05} & Bioinformatics \\ \cline{3-4} 
	&  & Language Translation \cite{koehn2005europarl} & Natural Language Processing \\ \cline{3-4} 
	&  & Named Entity Recognition \cite{DBLP:journals/coling/Shaalan14} & Natural Language Processing \\ \hline
	\multirow{3}{*}{Network Analysis} & Graph & Scene Graph \cite{DBLP:conf/nips/NewellD17} & Computer Vision \\ \cline{2-4} 
	& Tree & Natural Language Parsing \cite{taskar2004max} & Natural Language Processing \\ \cline{2-4} 
	& Link & Link Prediction \cite{DBLP:journals/jasis/Liben-NowellK07} & Data Mining \\ \hline
	\multirow{3}{*}{Data Generation} & Image & Super-resolution Image Reconstruction \cite{park2003super} & Computer Vision \\ \cline{2-4} 
	& Text & Language Generation & Natural Language Processing \\ \cline{2-4} 
	& Audio & Music Generation \cite{DBLP:conf/ismir/YangCY17} & Signal Processing \\ \hline
	\multirow{3}{*}{Semantic Retrieval} & \multirow{3}{*}{\begin{tabular}[c]{@{}c@{}}Independent \\ Real-valued\\ Vector\end{tabular}} & Content-based Image Retrieval \cite{DBLP:journals/pami/SmeuldersWSGJ00} & Computer Vision \\ \cline{3-4} 
	&  & Microblog Retrieval \cite{DBLP:conf/trec/LauLT11} & Data Mining \\ \cline{3-4} 
	&  & News Retrieval \cite{DBLP:conf/sigir/MariaS00} & Data Mining \\ \hline
	\multirow{3}{*}{Time-series Prediction} & \multirow{3}{*}{\begin{tabular}[c]{@{}c@{}} Time Series\\ \end{tabular}} & DNA Microarray Data Analysis \cite{DBLP:journals/titb/ChoongCY09} & Bioinformatics \\ \cline{3-4} 
	&  & Energy Consumption Forecasting \cite{azadeh2008annual} & Energy Meteorology \\ \cline{3-4} 
	&  & Video Surveillance \cite{DBLP:journals/pami/WangMG09} & Computer Vision \\ \hline
\end{tabular}%
%\end{scriptsize}
}
\end{table*}

\section{Multi-output Learning}
In contrast to traditional single-output learning, multi-output learning can concurrently predict multiple outputs. The outputs can be of various types and structures, and the problems that can be solved are diverse. A summary of the subfields that use multi-output learning along with their corresponding output types, structures, and applications is presented in Table \ref{summary_of_subfields}.

We being the section with an introduction to some of the output structures in multi-output learning problems. The different problem definitions common to various subfields are provided next, along with the different constraints placed on the output space. We also discuss some special cases of these problems and give a brief overview of some of the evaluation metrics that are specific to multi-output learning. The section concludes with some insights into the evolution of output dimensions through an analysis of several commonly used datasets.

\subsection{Myriads of Output Structures}\label{sec:structures}
\begin{figure}
	\centering
	\includegraphics[scale=0.245]{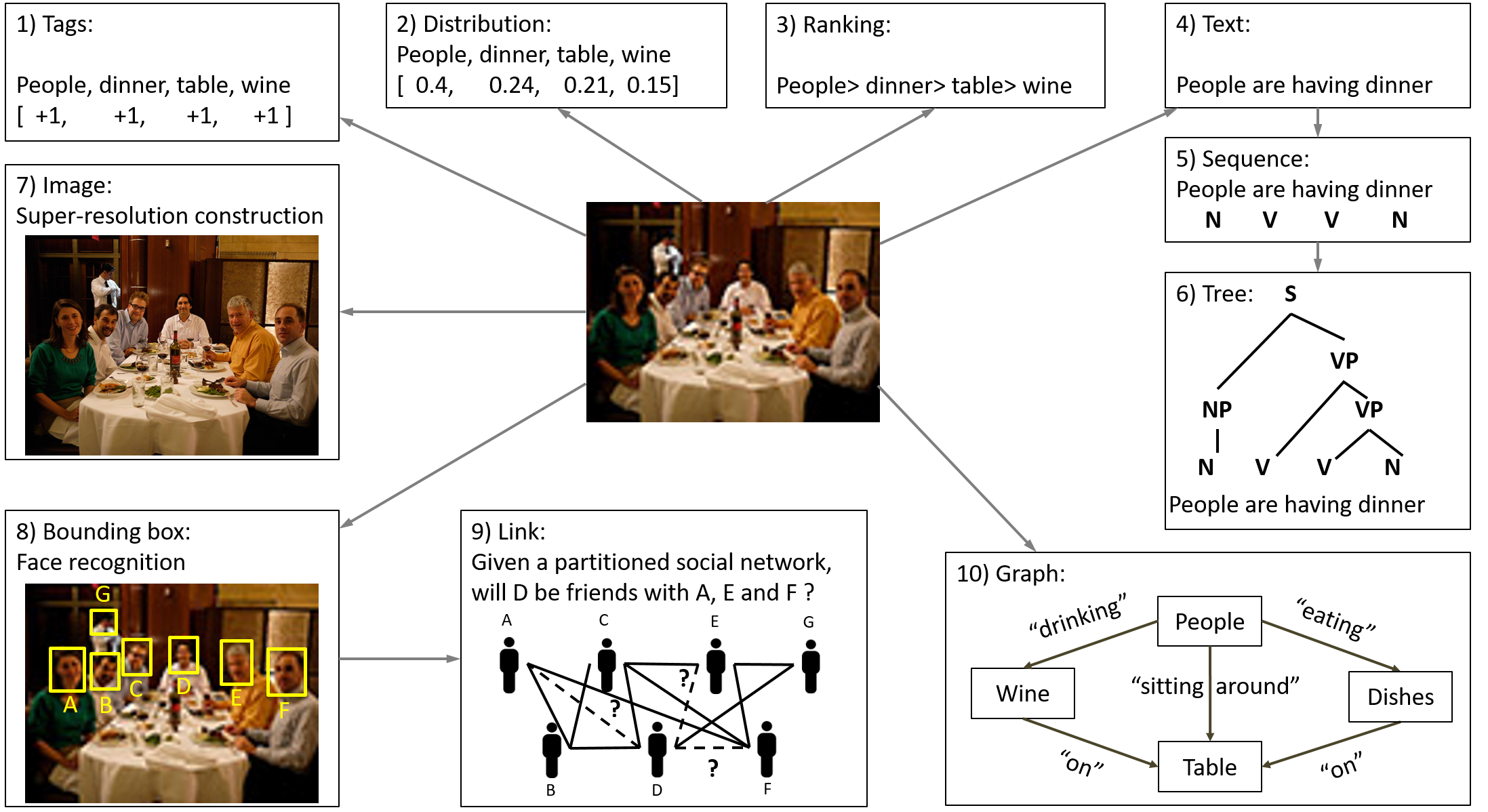}
	\caption{\label{output_structures} An illustration of the myriads of output structures given an input image from a social network.}
\end{figure}

The increasing demand of sophisticated decision-making tasks has led to new creations of outputs, some of which have complex structures. With social media, social networks, and various online services becoming ubiquitous, a wide range of output labels can be stored and then collected by researchers. Output labels can be anything; they could be text, images, audio, or video, or a combination as multimedia. For example, given a long document as input, the output might be a summary of the input in text format. Given some text fragments, the output might be an image with its contents described by the input text. Similarly, audio, such as music and videos, can be generated given different types of inputs. In addition to the different output types, there are also a number of different possible output structures. Here we present several typical output structures given an image as an input using the example in Fig.~\ref{output_structures} as a way to illustrate just how many output structures might be possible across all the different input types.

\subsubsection{Independent Vector}
An independent vector is a vector with separate dimensions (features), where each dimension represents a particular label that does not necessarily depend on other labels. Binary vectors can be used to represent a given piece of data as tags, attributes, BOW, bag-of-visual-words, hash codes, etc. Real-valued vectors provide the weighted dimensions, where the real value represents the strength of the input data against the corresponding label. Applications include annotation or classification of text, images, or video with binary vectors \cite{DBLP:conf/cikm/CaiH04,DBLP:conf/cvpr/XiaoHEOT10,DBLP:conf/mm/QiHRTMZ07}, and demand or energy prediction with real-valued vectors \cite{aras2004forecasting}. An independent vector can be used to represent the tags of an image, as shown in Fig.~\ref{output_structures} (1), where all the tags ``people'', ``dinner'', ``table'' and ``wine'' have equal weight..

\subsubsection{Distribution}
Unlike independent vectors, distributions provide information about the probability that a particular dimension will be associated with a particular data sample. In Fig.~\ref{output_structures} (2), the tag with the largest weight is ``people" and is the main content of the image, while ``dinner" and ``table" have similar distributions. Applications for distribution outputs include head pose estimation \cite{DBLP:conf/cvpr/GengX14}, facial age estimation \cite{DBLP:conf/aaai/GengSZ10} and text mining \cite{DBLP:conf/emnlp/ZhouZZZG16}.

\subsubsection{Ranking}
Outputs might also be in the form of a ranking, which shows the tags ordered from the most to least important. The results from a distribution learning model can be converted into a ranking, but a ranking model is not restricted to only distribution learning models. Text categorization \cite{DBLP:journals/jmlr/CrammerS03a}, question answering \cite{DBLP:conf/sigir/KoNS07} and visual object recognition \cite{DBLP:conf/iccv/BucakMJJ09} are applications where rankings are often used.

\subsubsection{Text}
Text can be in the form of keywords, sentences, paragraphs, or even documents. Fig.~\ref{output_structures} (4) illustrates an example of text output as a caption of the image – ``People are having dinner". Other applications for text outputs are document summarization \cite{DBLP:conf/ijcai/ShenSLYC07} and paragraph generation \cite{DBLP:conf/iccv/LiangHZGX17}.

\subsubsection{Sequence}
Sequence outputs refer to a series of elements selected from a label set. Each element is predicted depending on the input as well as the predicted output(s) from the preceding element. An output sequence often corresponds to an input sequence. For example, in speech recognition, we expect the output to be a sequence of text that corresponds to a given audio signal of speech \cite{DBLP:conf/icassp/GravesMH13}. In language translation, we expect the output to be a sentence transformed into the target language \cite{koehn2005europarl}. In the example shown in Fig.~\ref{output_structures} (5), the input is an image caption, i.e., text, and the outputs are part-of-speech (POS) tags for each word in the sequence.

\subsubsection{Tree}
Tree outputs are essentially outputs in the form of a hierarchy. The outputs, usually labels, have an internal structure where each output has a label that belongs to, or is connected to, its ancestors in the tree. For example, in syntactic parsing \cite{taskar2004max}, as shown in Fig.~\ref{output_structures} (6), each of the outputs for an input sentence is a POS tag and the entire output is a parsing tree. ``people" is labeled as a noun N, but it is also a noun phrase NP as per the tree.

\subsubsection{Image}
Images are a special form of output that consist of multiple pixel values, where each pixel is predicted depending on the input and the pixels around it. Fig.~\ref{output_structures} (7) shows super-resolution construction \cite{park2003super} as one popular application where images are common outputs. Super-resolution construction means constructing a high-resolution image from a low-resolution image. Other image output applications include text-to-image synthesis \cite{DBLP:conf/icml/ReedAYLSL16}, which generates images from natural language descriptions, and face generation \cite{gauthier2014conditional}.

\subsubsection{Bounding Box}
Bounding boxes as outputs are often used to find the exact locations of an object or objects appearing in an image. This is a common task in object recognition and object detection \cite{DBLP:conf/iccv/BucakMJJ09}. In Fig.~\ref{output_structures} (8), each of the faces is localized by a bounding box so that each person can be identified.

\subsubsection{Link}
Links as outputs usually represent the association between two nodes in a network \cite{DBLP:journals/jasis/Liben-NowellK07}. Fig.~\ref{output_structures} (9) illustrates a task to predict whether two currently unlinked users will be friends in the future given a partitioned social network where the edges represent friendships between users.

\subsubsection{Graph}
Graphs are commonly used to model relationships between. They consist of a set of nodes and edges, where a node represents an object and an edge indicates a relationship between two objects. Scene graphs \cite{DBLP:conf/cvpr/JohnsonKSLSBL15}, for example, are often output as a way to describe the content of an image \cite{DBLP:conf/nips/NewellD17}. Fig.~\ref{output_structures} (10) shows that, given an input image, the output is a graph definition where the nodes are the objects appearing in the image, i.e., ``people", ``dinner", ``table", and ``wine", and the edges are the relationships between these objects. Scene graphs are very useful as representations for tasks like image generation \cite{DBLP:conf/cvpr/JohnsonGF18} and visual question answering \cite{DBLP:journals/ijcv/KrishnaZGJHKCKL17}.

\subsubsection{Other Outputs}
Beyond these few types, there are still many other types of output structures. For example, contour and polygon outputs are similar to bounding boxes and can be used as labels for object localization. In information retrieval, the output(s) could be of the list type, say, of data objects that are similar to the given query. In image segmentation, the outputs are usually segmentation masks of different objects. In signal processing, outputs might be audio of speech or music. In addition, some real-world applications may require more sophisticated output structures relating to multiple tasks. For example, one may require that objects be recognized and localized at the same time, such as in co-saliency, i.e., discovering the common saliency of multiple images \cite{DBLP:journals/corr/ZhangFHW16}, simultaneously segmenting similar objects given multiple images in co-segmentation \cite{DBLP:conf/cvpr/JoulinBP10}, or detecting and identifying objects in multiple images in object co-detection~\cite{bao2012object}.

\subsection{Problem Definition of Multi-output Learning}\label{sec:mechanism}
Multi-output learning maps each input (instance) to multiple outputs. Assume $\mathcal{X} = \mathds{R}^d$ is a $d$-dimensional input space, and $\mathcal{Y} = \mathds{R}^m$ is an $m$-dimensional output label space. The aim of multi-output learning is to learn a function $f: \mathcal{X} \rightarrow \mathcal{Y}$ from the training set $\mathcal{D} = \{(\mathbf{x}_i,\mathbf{y}_i) | 1 \le i \le n\}$. For each training example $(\mathbf{x}_i,\mathbf{y}_i)$, $\mathbf{x}_i \in \mathcal{X}$ is a $d$-dimensional feature vector, and $\mathbf{y}_i \in \mathcal{Y}$ is the corresponding output associated with $\mathbf{x}_i$. The general definition of multi-output learning is given as: Finding a function $F: \mathcal{X} \times \mathcal{Y} \rightarrow \mathds{R}$ based on the training sample of input-output pairs, where $F(\mathbf{x},\mathbf{y})$ is a compatibility function that evaluates how compatible the input $\mathbf{x}$ and the output $\mathbf{y}$ are. Then, given an unseen instance $\mathbf{x}$ at the test state, the output is predicted to be the one with the largest compatibility score, namely  $f(\mathbf{x}) = \widetilde{\mathbf{y}} = \arg \max_{\mathbf{y} \in \mathcal{Y}}F(\mathbf{x},\mathbf{y})$~\cite{tsochantaridis2005large}. 
%\textcolor{blue}{This definition provides a general framework for multi-output learning problems within which most of its  sub-fields can be defined with certain constraints on its output label space $\mathcal{Y}$.}

This definition provides a general framework for multi-output learning problems. Although different multi-output learning subfields vary in their output structures, they can be defined within this framework given certain constraints on the output label space $\mathcal{Y}$. 

We selected several popular subfields and present the constraints of their output space in the following sections. Note that multi-output learning is not restricted to these particular scenarios; they are just examples for illustration.

\subsubsection{Multi-label Learning}
The task of multi-label learning is to learn a function $f(\cdot)$ that predicts the proper label sets for unseen instances\cite{DBLP:journals/tkde/ZhangZ14}. In this task, each instance is associated with a set of class labels/tags and is represented by a sparse binary label vector. A value of $+1$ denotes the instance is labeled and $−1$ means unlabeled. Thus, $\mathbf{y}_i \in \mathcal{Y} = \{-1,+1\}^m$. Given an unseen instance $\mathbf{x} \in \mathcal{X}$, the learned multi-label classification function $f(\cdot)$ outputs $f(\mathbf{x}) \in \mathcal{Y}$, where the labels in the output vector with a value of $+1$ are used as the predicted labels for $\mathbf{x}$.

\subsubsection{Multi-target Regression}
The aim of multi-target regression is to simultaneously predict multiple real-valued output variables for one instance \cite{DBLP:journals/widm/BorchaniVBL15,DBLP:journals/kbs/LiuCO18}. Here, multiple labels are associated with each instance, represented by a real-valued vector, where the values represent how strongly the instance corresponds to a label. Therefore, we have the constraint of $\mathbf{y}_i \in \mathcal{Y} = \mathds{R}^m$. Given an unseen instance $\mathbf{x} \in \mathcal{X}$, the learned multi-target regression function $f(\cdot)$ predicts a real-valued vector $f(\mathbf{x}) \in \mathcal{Y}$ as the output.

\subsubsection{Label Distribution Learning} Label distribution learning determines the relative importance of each label in the multi-label learning problem~\cite{geng2016label}.
This is as opposed to multi-label learning, which simply learns to predict a set of labels. But, as illustrated in Fig.\ref{output_structures}, the idea of label distribution learning is to predict multiple labels with a degree value that represents how well each label describes the instance. Therefore, the sum of the degree values for each instance is 1. Thus, the output space for label distribution learning satisfies $\mathbf{y}_i = (y_i^1, y_i^2,...,y_i^m) \in \mathcal{Y} = \mathds{R}^m$ with the constraints $y_i^j \in [0,1], 1  \le j \le m$ and $\sum_{j=1}^{m}y_i^j = 1$. 

\subsubsection{Label Ranking} The goal of label ranking is to map instances to a total order over a finite set of predefined labels~\cite{DBLP:books/daglib/p/VembuG10}.
In label ranking, each instance is associated with the rankings of multiple labels. Therefore, the outputs of the problem are the total order of all the labels for each instance. Let $\mathcal{L} = \{\lambda_1,\lambda_2, ...,\lambda_m\}$ denotes the predefined label set. A ranking can be represented as a permutation $\pi$ on $\{1, 2, ..., m\}$, such that $\pi(j) = \pi(\lambda_j)$ is the position of the label $\lambda_j$ in the ranking. Therefore, given an unseen instance $\mathbf{x} \in \mathcal{X}$, the learned label ranking function $f(\cdot)$ predicts a permutation $f(\mathbf{x}) = (y_i^{\pi(1)}, y_i^{\pi(2)},...,y_i^{\pi(m)}) \in \mathcal{Y}$ as the output.

\subsubsection{Sequence Alignment Learning} Sequence alignment learning aims to identify the regions of relationships between two or more sequences. The outputs in this task are a sequence of multiple labels for the input instance. The output vector has the constraint $\mathbf{y}_i \in \mathcal{Y} = \{0,1,...,c\}^m$, where $c$ denotes the total number of labels. In sequence alignment learning, $m$ may vary depending on the input. Given an unseen instance $\mathbf{x} \in \mathcal{X}$, the learned sequence alignment function $f(\cdot)$ outputs $f(\mathbf{x}) \in \mathcal{Y}$, where all of the predicted labels in the output vector form the predicted sequence for $\mathbf{x}$.

\subsubsection{Network Analysis}
Network analysis explores the relationships and interactions between objects and entities in a network structure, and link prediction is a common task within this subfield. Let $G = (V, E)$ denotes the graph of a network. $V$ is the set of nodes, which represent objects, and $E$ is the set of edges, which represent the relationships between objects. Given a snapshot of a network, the goal of link prediction is to infer whether a connection exists between two nodes. The output vector $\mathbf{y}_i \in \mathcal{Y} = \{-1, +1\}^m$ is a binary vector whose value represents whether there will be an edge $e = (u, v)$ between any pair of nodes $u, v \in V$ and $e \notin E$. $m$ is the number of node pairs that does not appear in the current graph $G$ and each dimension in $\mathbf{y}_i$ represents a pair of nodes that are not currently connected.

\subsubsection{Data Generation}
Data generation is a subfield of multi-output learning that aims to create and then output structured data of a certain distribution. Deep generative models are usually used to generate the data, which may be in the form of text, images, or audio. The multiple output labels in the problem become the different words in the vocabulary, the pixel values, the audio tones, etc. Take image generation as an example. The output vector has the constraint $\mathbf{y}_i \in \mathcal{Y} = \{0,1,...,255\}^{m_w \times m_h \times 3}$, where $m_w$ and $m_h$ are the width and height of the image. Given an unseen instance $\mathbf{x} \in \mathcal{X}$, which is usually a random noise or an embedding vector with some constraints, the learned GAN-based network $f(\cdot)$ outputs $f(\mathbf{x}) \in \mathcal{Y}$, where all of the predicted pixel values in the output vector form the generated image for $\mathbf{x}$.

\subsubsection{Semantic Retrieval} 
Semantic retrieval means finds the meanings within some given information. Here, we consider semantic retrieval in a setting where each input instance has semantic labels that can be used to help retrieval \cite{DBLP:journals/pami/CarneiroCMV07}. Thus, each instance representation comprises semantic labels as the output $\mathbf{y}_i \in \mathcal{Y} = \mathds{R}^m$. Given an unseen instance $\mathbf{x} \in \mathcal{X}$ as the query, the learned retrieval function $f(\cdot)$ predicts a real-valued vector $f(\mathbf{x}) \in \mathcal{Y}$ as the intermediate output result. The intermediate output vector can then be used to retrieve a list of similar data instances from the database by using a proper distance-based retrieval method.

\subsubsection{Time-series Prediction}
The goal in time-series prediction is to predict the future values in a series based on previous observations~\cite{weigend2018time}. The inputs are a series of data vectors for a period of time, and the output is a data vector for a future timestamp. Let $t$ denotes the time index. The output vector at time $t$ is represented as $\mathbf{y}_i^t \in \mathcal{Y} = \mathds{R}^m$. Therefore, the outputs within a period of time from $t=0$ to $t=T$ are $\mathbf{y}_i = (\mathbf{y}_i^0,...,\mathbf{y}_i^t,...\mathbf{y}_i^T)$. Given previously observed values, the learned time-series function outputs predicted consecutive future values. 

\subsection{Special Cases of Multi-output Learning}
\subsubsection{Multi-class Classification}
Multi-class classification can be categorized as a traditional single-output learning paradigm if the output class is represented as either an integer encoding or a one-hot vector.

\subsubsection{Fine-grained Classification}
Fine-grained classification is a challenging multi-classification task where the categories may only have subtle visual differences~\cite{akata2015evaluation}. Although the output of fine-grained classification shares the same vector representation as multi-class classification, the vectors have different internal structures. Also, in its label hierarchy, labels with the same parents tend to be more closely related than labels with different parents.

\subsubsection{Multi-task Learning}
The aim of multi-task learning (MTL) is the subfield that aims to improve generalization performance by learning multiple related tasks simultaneously~\cite{caruana1997multitask,thrun1998clustering}. Each task in the problem outputs one single label or value. This can be thought of as part of the multi-output learning paradigm in that learning multiple tasks is similar to learning multiple outputs. MTL leverages the relatedness between tasks to improve the performance of learning models. One major difference between multi-task learning and multi-output learning is that, in multi-task learning, different tasks might be trained on different training sets or features, while, in multi-output learning, the output variables usually share the same training data or features.

\subsection{Model Evaluation Metrics}

In this section, we presents the conventional evaluation metrics used to assess the multi-output learning models with a test dataset. Let $\mathcal{T} = \{(\mathbf{x}_i,\mathbf{y}_i) | 1 \le i \le N\}$ be the test dataset, $f(\cdot)$ be the multi-output learning model, and $\hat{\mathbf{y}}_i = f(\mathbf{x}_i)$ be the predicted output of $f(\cdot)$ for the testing example $\mathbf{x}_i$. In addition, let $Y_i$ and $\hat{Y}_i$ denote the set of labels corresponding to $\mathbf{y}_i$ and $\hat{\mathbf{y}}_i$, respectively. $\mathds{I}$ is an indicator function, where $\mathds{I}(g) = 1$ if $g$ is true, and $0$ otherwise.
% (which will only be used in classification-based metrics). 
\subsubsection{Classification-based Metrics}
Classification-based metrics evaluate the performance of multi-output learning with respect to classification problems, such as multi-label classification, semantic retrieval, image annotation, label ranking, etc. The outputs are usually in discrete values. The conventional classification metrics fall into three groups: \emph{example-based}, \emph{label-based} and \emph{ranking-based}. 

\begin{enumerate}[label=(\alph*),widest=viii,itemindent=*,leftmargin=0pt]
\item Example-based Metrics: 
\emph{Example-based metrics}~\cite{mao2012objective} evaluate the performance of multi-output learning models with respect to each data instance. Performance is first evaluated on each test instance separately, and then the mean of all the individual results is used to reflect the overall performance of the model.
The evaluation for multi-output classification tasks works under the same mechanism as binary classification (single output) tasks, the classic metrics for binary classification can be extended to evaluate multi-output classification models~\cite{DBLP:journals/tkde/ZhangZ14}. The commonly used metrics are exact match ratio, accuracy, precision, recall and $F_1$ score.

\begin{description}[leftmargin=*]
\item [Hamming loss] is an example-based metric specifically designed for multi-output classification tasks. It computes the average difference between the predicted and actual output, considering both prediction and omission errors, i.e., when the prediction is incorrect or a label is not predicted at all. The Hamming loss averaged overall data instances is defined as:
\begin{equation*}
\begin{split}
Hamming Loss = \frac{1}{N}\sum_{i=1}^{N}\frac{1}{m}|Y_i \Delta \hat{Y}_i|
\end{split}
\end{equation*}
where $m$ is the number of labels and $\Delta$ represents the symmetric difference between two sets.
The lower the hamming loss, the better the performance of the model is.
\end{description}

\item Label-based Metrics: 
\emph{Label-based metrics} evaluate performance with respect to each output label. These metrics aggregate the contributions of all the labels to arrive at an averaged evaluation of the model. There are two techniques for obtaining label-based metrics: macro- and micro-averaging. Macro-based approaches compute the metrics for each label independently and then average over all the labels with equal weights. By contrast, micro-based approaches give equal weight to every data sample. Let $TP_l$, $FP_l$, $TN_l$ and $FN_l$ denote the number of true positives, true negatives, false positives, and false negatives, for each label, respectively. Let $B$ be a binary evaluation metric (accuracy, precision, recall or $F_1$ score) for a particular label. The macro and micro approaches are therefore defined as -
\begin{description}[leftmargin=*]
\item [macro-averageing:]
\begin{equation*}
B_{macro} = \frac{1}{m}\sum_{l=1}^{m}B(TP_l, FP_l, TN_l, FN_l),
\end{equation*}
\item [micro-averaging:] 
\begin{equation*}
B_{micro} = B(\frac{1}{m}\sum_{l=1}^{m}TP_l, \frac{1}{m}\sum_{l=1}^{m}FP_l, \frac{1}{m}\sum_{l=1}^{m}TN_l, \frac{1}{m}\sum_{l=1}^{m}FN_l).
\end{equation*}
\end{description}
\item Ranking-based Metrics: 
\emph{Ranking-based metrics} evaluate the performance in terms of the ordering of the output labels. 
\begin{description}[leftmargin=*]
\item [One-error] is the number of times the top-ranked label is not in the true label set. This approach only considers the most confident predicted label of the model. An averaged one-error over all data instances is computed as:
\begin{equation*}
\text{\emph{One-error}} = \frac{1}{N}\sum_{i=1}^{N}\mathds{I}(\arg\min_{\lambda \in \mathcal{L}}\pi_i(\lambda) \notin Y_i)
\end{equation*}
where $\mathds{I}$ is an indicator function, $\mathcal{L}$ denotes the label set, and $\pi_i(\lambda)$ is the predicted rank of label $\lambda$ for the test instance $\mathbf{x_i}$. The smaller the one-error, the better the performance.
\item [Ranking loss] indicates the average proportion of incorrectly ordered label pairs.
\begin{equation*}
\begin{split}
Ranking Loss = \frac{1}{N}\sum_{i=1}^{N}\frac{1}{|Y_i||\overline{Y_i}|}|E|, where&\\
E = {(\lambda_a, \lambda_b): \pi_i(\lambda_a) > \pi_i(\lambda_b), (\lambda_a, \lambda_b) \in Y_i \times \overline{Y_i}}&
\end{split}
\end{equation*}
where $\overline{Y_i} = \mathcal{L} \setminus Y_i$. The smaller the ranking loss, the better the performance of the model.
\item [Average Precision (AP)] is the proportion of the labels ranked above a particular label in the true label set as an average over all the true labels. The larger the value, the better the performance of the model is. The averaged AP over all test data instances is defined as follows:
\begin{equation*}
\begin{split}
AP = \frac{1}{N}\sum_{i=1}^{N}\frac{1}{|Y_i|}
\sum_{\lambda \in Y_i}\frac{\{\lambda' \in Y_i | \pi_i(\lambda') \le \pi_i(\lambda)\}}{\pi_i(\lambda)}
\end{split}
\end{equation*}
\end{description}
\textbf{\emph{Discussion:}} The metrics listed above are those commonly used with classification-based multi-output learning problems. But the choice of metrics varies according to the different considerations of each application. Take image annotation for example. If the aim of the task is to annotate each image correctly, example-based metrics are optimal for evaluating performance. However, if the objective is keyword-based image retrieval, the macro-averaging metric is preferable~\cite{mao2012objective}. Further, some metrics are more suited to special cases of multi-output learning problems. For instance, for imbalanced learning tasks, geometric mean~\cite{DBLP:journals/pr/TahirKY12} for some classification-based metrics, e.g. the errors, accuracy, F1-scores and etc., are more convincing to be used for evaluation. The minimum sensitivity~\cite{DBLP:journals/npl/AlejoGP15} can help determine the classes that hinder the performance in the imbalanced setting. We do not discuss these metrics in detail as they are not the focus here.

\end{enumerate}
\subsubsection{Regression-based Metrics}
Unsurprisingly, regression-based metrics evaluate multi-output learning performance with regression problems, e.g., object localization or image generation. The outputs are usually real values.

%These metrics are averaged over all the testing instances and all the outputs.
\begin{description}[leftmargin=*]
\item [Mean absolute error (MAE)] is a classic single-output regression metric that computes the absolute difference between the predicted and the actual outputs. It can be extended to evaluate multi-output regression models by simply averaging the metric over all the outputs.
\begin{equation*}
\begin{split}
MAE = \frac{1}{m}\frac{1}{N}\sum_{i=1}^{N}|\mathbf{y}_i - \hat{\mathbf{y}}_i|
\end{split}
\end{equation*}
\item [Mean squared error (MSE)] is a regression metric that computes the average squared difference between the predicted and the actual outputs. Like MAE, it can also be extended to the multi-output setting. However, MSE is more sensitive to the outliers, as it will contribute much higher errors compared to MAE.
\begin{equation*}
\begin{split}
MSE = \frac{1}{m}\frac{1}{N}\sum_{i=1}^{N}(\mathbf{y}_i - \hat{\mathbf{y}_i})^2
\end{split}
\end{equation*}
\item [Average correlation coefficient (ACC)] measures the degree of association between the actual and the predicted outputs.
\begin{equation*}
\begin{split}
ACC = \frac{1}{m}\sum_{l=1}^{m}\frac{\sum_{i=1}^{N}(y_{i}^{l} - \bar{\mathbf{y}}^l) (\hat{y}_{i}^l - \bar{\hat{\mathbf{y}}}^l)}{\sqrt{\sum_{i=1}^{N}(y_{i}^{l} - \bar{\mathbf{y}}^l)^2 \sum_{i=1}^{N}({\hat{y}_i}^l - \bar{\hat{\mathbf{y}}}^l)^2}}
\end{split}
\end{equation*}
where $y_{i}^{m}$ and $\hat{y}_{i}^m$ are the actual and predicted $m$ output of $\mathbf{x_i}$, respectively, and $\bar{\mathbf{y}}^l$ and $\bar{\hat{\mathbf{y}}}^l$ are the vectors of the averages of the actual and predicted outputs for a label $l$ over all samples.
\item [Intersection over union threshold (IoU)] is a specifically-designed metric for assessing object localization or segmentation. It is computed as:
\begin{equation*}
IoU = \frac{\text{Area of Overlap}}{\text{Area of Union}}
\end{equation*}
where area of overlap is the area of intersection between the predicted and the actual bounding boxes/segmentation masks. Similarly, area of union is the union area between the actual and predicted boxes/masks.
\end{description}

\subsubsection{New Metrics}
Data generation is an emerging subfield of multi-out learning that uses generative models to output structured data with certain distributions. Based on the particulars of the task at hand, a model's performance is usually evaluated in two respects: 1). whether the generated data actually follows the desired real data distribution; and 2). the quality of the generated samples. Metrics like average log-likelihood~\cite{goodfellow2014generative}, coverage metric~\cite{tolstikhin2017adagan}, maximum mean discrepancy (MMD)~\cite{gretton2012kernel}, geometry score~\cite{khrulkov2018geometry}, are frequently used to assess the veracity of the distribution. Metrics that quantify the quality of the generated data remain challenging. The commonly used are inception scores (IS)~\cite{salimans2016improved}, mode score (MS)~\cite{che2016mode}, Fréchet inception distance (FID)~\cite{heusel2017gans} and kernel inception distance (KID)~\cite{binkowski2018demystifying}. Precision, recall and F1 score are also employed in GANs to quantify the degree of overfitting in the model~\cite{DBLP:conf/nips/LucicKMGB18}.
\begin{table*}[]
\centering
\caption{Characteristics of the datasets of multi-output learning tasks.}
\label{datasets-stats}
\resizebox{\textwidth}{!}{%
\begin{tabular}{|l|l|l|l|l|l|}
	\hline
	\textbf{\begin{tabular}[c]{@{}l@{}}Multi-output \\ Characteristic\end{tabular}} & \textbf{Challenge} & \textbf{Application Domain} & \textbf{Dataset Name} & \textbf{Statistics} & \textbf{Source} \\ \hline
	\multirow{24}{*}{\textbf{Volume}} & \multirow{10}{*}{\begin{tabular}[c]{@{}l@{}}Extreme Output \\ Dimension\footnotemark\end{tabular}} &  &  & \textbf{Output Dimension} &  \\ \cline{3-6} 
	&  & Review Text & AmazonCat-13K & 13,330 & \cite{DBLP:conf/recsys/McAuleyL13} \\ \cline{3-6} 
	&  & Review Text & AmazonCat-14K & 14,588 & \cite{DBLP:conf/sigir/McAuleyTSH15,DBLP:conf/kdd/McAuleyPL15} \\ \cline{3-6} 
	&  & Text & Wiki10-31 & 30,938 & \cite{DBLP:conf/nips/BhatiaJKVJ15,DBLP:journals/corr/abs-1202-5469} \\ \cline{3-6} 
	&  & Social Bookmarking & Delicious-200K & 205,443 & \cite{DBLP:conf/nips/BhatiaJKVJ15,wetzker2008analyzing} \\ \cline{3-6} 
	&  & Text & WikiLSHTC-325K & 325,056 & \cite{DBLP:conf/kdd/PrabhuV14,DBLP:journals/corr/PartalasKBAPGAA15} \\ \cline{3-6} 
	&  & Text & Wikipedia-500K & 501,070 & Wikipedia \\ \cline{3-6} 
	&  & Product Network & Amazon-670K & 670,091 & \cite{DBLP:conf/nips/BhatiaJKVJ15,DBLP:conf/recsys/McAuleyL13} \\ \cline{3-6} 
	&  & Text & Ads-1M & 1,082,898 & \cite{DBLP:conf/kdd/PrabhuV14} \\ \cline{3-6} 
	&  & Product Network & Amazon-3M & 2,812,281 & \cite{DBLP:conf/sigir/McAuleyTSH15,DBLP:conf/kdd/McAuleyPL15} \\ \cline{2-6} 
	& \multirow{5}{*}{\begin{tabular}[c]{@{}l@{}}Extreme Class \\ Imbalance\end{tabular}} &  &  & \textbf{\begin{tabular}[c]{@{}l@{}}Largest Class \\ Imbalance Ratio\end{tabular}} &  \\ \cline{3-6} 
	&  & Scene Image & WIDER-Attribute & 1:28 & \cite{DBLP:conf/eccv/LiHLT16} \\ \cline{3-6} 
	&  & Face Image & Celeb Faces Attributes & 1:43 & \cite{DBLP:conf/iccv/LiuLWT15} \\ \cline{3-6} 
	&  & Clothing Image & DeepFashion & 1:733 & \cite{DBLP:conf/cvpr/LiuLQWT16} \\ \cline{3-6} 
	&  & Clothing Image & X-Domain & 1:4,162 & \cite{DBLP:conf/cvpr/ChenHFBDY15} \\ \cline{2-6} 
	& \multirow{9}{*}{Unseen Outputs} &  &  & \textbf{Seen / Unseen Labels} &  \\ \cline{3-6} 
	&  & Image & Attribute Pascal abd Yahoo & 20 / 12 & \cite{DBLP:conf/cvpr/FarhadiEHF09} \\ \cline{3-6} 
	&  & Animal Image & Animal with Attributes & 40 / 10 & \cite{DBLP:conf/cvpr/FarhadiEHF09} \\ \cline{3-6} 
	&  & Scene Image & HSUN & 80 / 27 & \cite{DBLP:conf/cvpr/ChoiLTW10} \\ \cline{3-6} 
	&  & Music & MagTag5K & 107 / 29 & \cite{DBLP:conf/ismir/MarquesDLG11} \\ \cline{3-6} 
	&  & Bird Image & Caltech-UCSD Birds 200 & 150 / 50 & \cite{WahCUB_200_2011} \\ \cline{3-6} 
	&  & Scene Image & SUN Attributes & 645 /72 & \cite{DBLP:conf/cvpr/XiaoHEOT10} \\ \cline{3-6} 
	&  & Health & MIMIC II & 3,228 / 355 & \cite{jouhet2012automated} \\ \cline{3-6} 
	&  & Health & MIMIC III & 4,403 / 178 & \cite{johnson2016mimic} \\ \hline
	\multirow{7}{*}{\textbf{Velocity}} & \multirow{7}{*}{\begin{tabular}[c]{@{}l@{}}Change of Output \\ Distribution\end{tabular}} &  &  & \textbf{Time Periods} &  \\ \cline{3-6} 
	&  & Text & Reuters & 365 days & \cite{DBLP:journals/jmlr/LewisYRL04} \\ \cline{3-6} 
	&  & Route & \begin{tabular}[c]{@{}l@{}}ECML/PKDD 15: \\ Taxi Trajectory Prediction\end{tabular} & 365 days & \cite{kaggletaxiroute2015} \\ \cline{3-6} 
	&  & Route & epfl/mobility & 30 days & \cite{CRAWDADdata2009} \\ \cline{3-6} 
	&  & Electricity & Portuguese Electricity Consumption & 365 days & \cite{ElectricityLoadDiagrams20112014} \\ \cline{3-6} 
	&  & Traffic Video & MIT Traffic Data Set & 90 minutes & \cite{DBLP:journals/pami/WangMG09} \\ \cline{3-6} 
	&  & Surveillance Video & VIRAT Video & 8.5 hours & \cite{DBLP:conf/cvpr/OhHPCCLMALDSWJRSVPRYTSFRD11} \\ \hline
	\multirow{13}{*}{\textbf{Variety}} & \multirow{13}{*}{Complex Structures} &  &  & \textbf{Output Structures} &  \\ \cline{3-6} 
	&  & Image & LabelMe & Label, Bounding Box & \cite{DBLP:journals/ijcv/RussellTMF08} \\ \cline{3-6} 
	&  & Image & ImageNet & Label, Bounding Box & \cite{ILSVRC15} \\ \cline{3-6} 
	&  & Image & PASCAL VOC & Label, Bounding Box & \cite{DBLP:journals/ijcv/EveringhamGWWZ10} \\ \cline{3-6} 
	&  & Image & CIFAR100 & Hierarchical Label & \cite{krizhevsky2009learning} \\ \cline{3-6} 
	&  & Lexical Database & WordNet & Hierarchy & \cite{DBLP:journals/cacm/Miller95} \\ \cline{3-6} 
	&  & Wikipedia Network & Wikipedia & Graph, Link & \cite{mahoney2011large} \\ \cline{3-6} 
	&  & Blog Network & BlogCatalog & Graph, Link & \cite{Zafarani+Liu:2009} \\ \cline{3-6} 
	&  & Author Collaboration Network & arXiv-AstroPh & Link & \cite{snapnets} \\ \cline{3-6} 
	&  & Author Collaboration Network & arXiv-GrQc & Link & \cite{snapnets} \\ \cline{3-6} 
	&  & Text & CoNLL-2000 Shared Task & Text Chunks & \cite{DBLP:conf/conll/SangB00} \\ \cline{3-6} 
	&  & Text & Wall Street Journal (WSJ) corpus & POS Tags, Parsing Tree & - \\ \cline{3-6} 
	&  & European Languages & Europarl corpus & Sequence & \cite{koehn2005europarl} \\ \hline
	\multirow{6}{*}{\textbf{Veracity}} & \multirow{6}{*}{Noisy Output Labels} &  &  & \textbf{Noisy Labeled Samples} &  \\ \cline{3-6} 
	&  & Dog Image & AMT & 7,354 & \cite{DBLP:conf/nips/ZhouPBM12} \\ \cline{3-6} 
	&  & Food Image & Food101N & 310K & \cite{DBLP:conf/cvpr/LeeHZY18} \\ \cline{3-6} 
	&  & Clothing Image & Clothing1M & 1M & \cite{DBLP:conf/cvpr/XiaoXYHW15} \\ \cline{3-6} 
	&  & Web Image & WebVision & 2.4M & \cite{DBLP:journals/corr/abs-1708-02862} \\ \cline{3-6} 
	&  & Image and Video & YFCC100M & 100M & \cite{DBLP:journals/corr/ThomeeSFENPBL15} \\ \hline
\end{tabular}%
}
\end{table*}

\subsection{Multi-output Learning Datasets}
Most of the datasets used to experiment with multi-output learning problems have either been constructed or become popular because they reflect, and therefore test, a challenge that needs to be overcome. We have presented these datasets according to the challenges reflected in the 4 Vs. Table~\ref{datasets-stats} lists the datasets, including their multi-output characteristics, the challenge can be tested, the application domain, plus the dataset name, source, and descriptive statistics.

\footnotetext{http://manikvarma.org/downloads/XC/XMLRepository.html}

The large-scale datasets, \emph{i.e.}, the datasets that can be used to test volume, are extremely large. The enormity of their corresponding statistics illustrate the pressing need to overcome the challenges caused by this particular V among the 4.
 
Many studies that have focused on change in output distribution, e.g., concept drift/velocity, rely on synthetic streaming data or static databases in their experiments. We have also included some of the more popular real-world and/or dynamic databases that are used to experiment with these tasks. As shown in the table, the datasets come from various application domains, demonstrating the importance of this challenge.

The datasets designed to test complex multi-output learning problems contain a mix of different output structures. For example, the image datasets listed in the table includes both labels and bounding boxes for the objects. These datasets can be used to test the variety of data.

Lastly, we come to veracity. Many efforts to detail with noisy labels evaluate their methods by beginning with a clean dataset to which artificial noise is then added. This helps researchers control and test different levels of noise. We have also listed several popular real-world datasets with some unknown level of errors in annotation.

\section{The Challenges of Multi-output Learning and Representative Works}
The pressing need for the complex prediction output and the explosive growth of output labels pose several challenges to multi-output learning and have exposed the inadequacies of many learning models that exist to date. In this section, we discuss each of these challenges and review several representative works on how they cope with these emerging phenomena. Further, given the success of artificial neural networks (ANNs), we also present several state-of-the-art examples of multi-output learning using an ANN for each challenge.

\begin{figure*}
	\centering
	\includegraphics[scale=0.46]{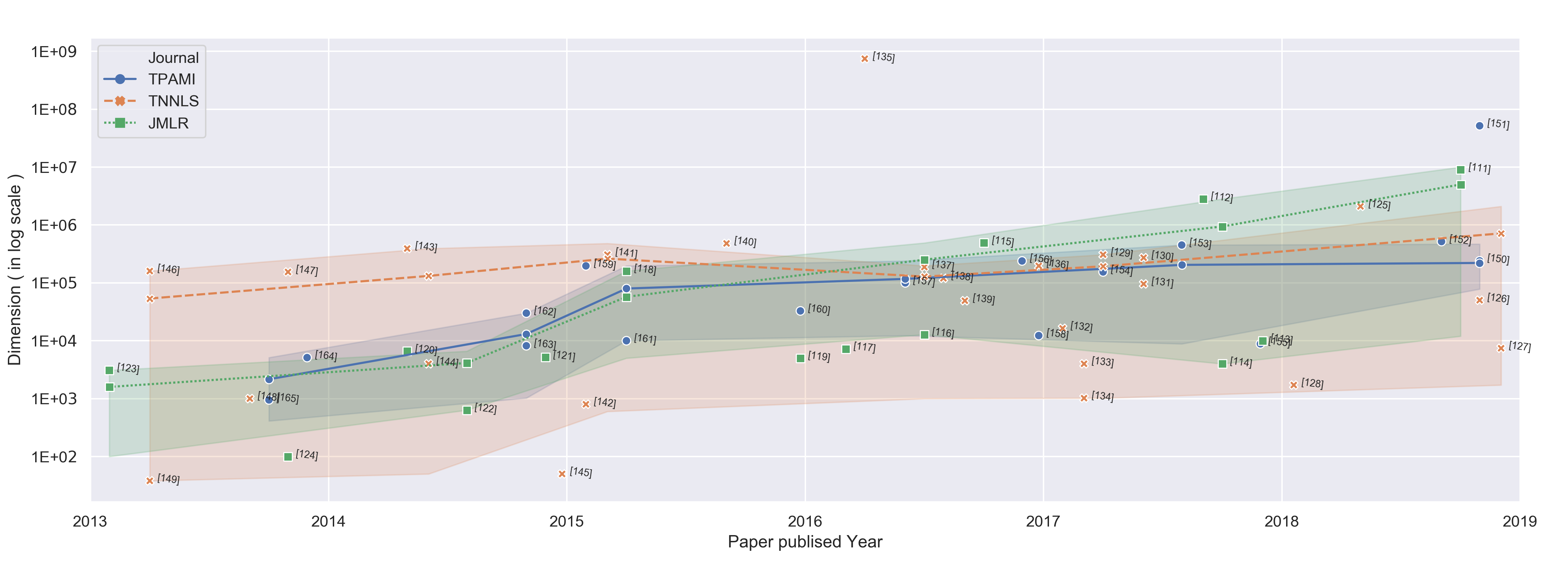}
	\caption{\label{output_dimensions_journal} Output dimension trends from papers published in the journals TPAMI, TNNLS, and JMLR since 2013 \cite{DBLP:journals/jmlr/KummerleS18,DBLP:journals/jmlr/LiuT17,DBLP:journals/jmlr/DupuyB17,DBLP:journals/jmlr/LiuTM17,DBLP:journals/jmlr/BrouardSd16,DBLP:journals/jmlr/ShinLKSYS16,DBLP:journals/jmlr/BabbarPGAA16,DBLP:journals/jmlr/LiZYL15,DBLP:journals/jmlr/HanLL15,DBLP:journals/jmlr/DoppaFT14,DBLP:journals/jmlr/ColomboM14,DBLP:journals/jmlr/GentileO14,DBLP:journals/jmlr/GongYZ13,DBLP:journals/jmlr/TalwalkarKMR13,DBLP:journals/tnn/FuLJZ18,8573154,DBLP:journals/tnn/ZhangWXCW18,DBLP:journals/tnn/LinDHS18,DBLP:journals/tnn/FangLTHC17,DBLP:journals/tnn/ZhangTGLL17,DBLP:journals/tnn/Kim17,DBLP:journals/tnn/CheungLPC17,DBLP:journals/tnn/WangLZ17,DBLP:journals/tnn/LiLXYZ17,DBLP:journals/tnn/LuoZLYXZ16,DBLP:journals/tnn/DengXZTGL16,DBLP:journals/tnn/AlushG16,DBLP:journals/tnn/TaoCSL16,DBLP:journals/tnn/CaoCTZ16,DBLP:journals/tnn/ZhuSLW15,DBLP:journals/tnn/ChenMKP15,DBLP:journals/tnn/LiBKL15,DBLP:journals/tnn/YuGTLZ14,DBLP:journals/tnn/BauerGBMK14,DBLP:journals/tnn/TangLSZ14,DBLP:journals/tnn/DengDLZH13,DBLP:journals/tnn/ZhangWN13,DBLP:journals/tnn/PangJJL13,DBLP:journals/tnn/LuoTXXLW13,zhang2018neural,DBLP:journals/pami/JeongLKKN18,DBLP:journals/pami/RaposoAB18,DBLP:journals/pami/CordtsREFR17,DBLP:journals/pami/XuCGN17,DBLP:journals/pami/FuJCSZ17,DBLP:journals/pami/HasnatAT16,DBLP:journals/pami/GirshickDDM16,DBLP:journals/pami/QinS16,DBLP:journals/pami/KwonKTKT15,DBLP:journals/pami/DjelouahFBCP15,DBLP:journals/access/WangWLZZ18,DBLP:journals/pami/ShitritBFF14,DBLP:journals/pami/ZhouF14,DBLP:journals/pami/PerakisPTK13,DBLP:journals/pami/GongLGP13}.}
\end{figure*}
\begin{figure*}
	\centering
	\includegraphics[scale=0.46]{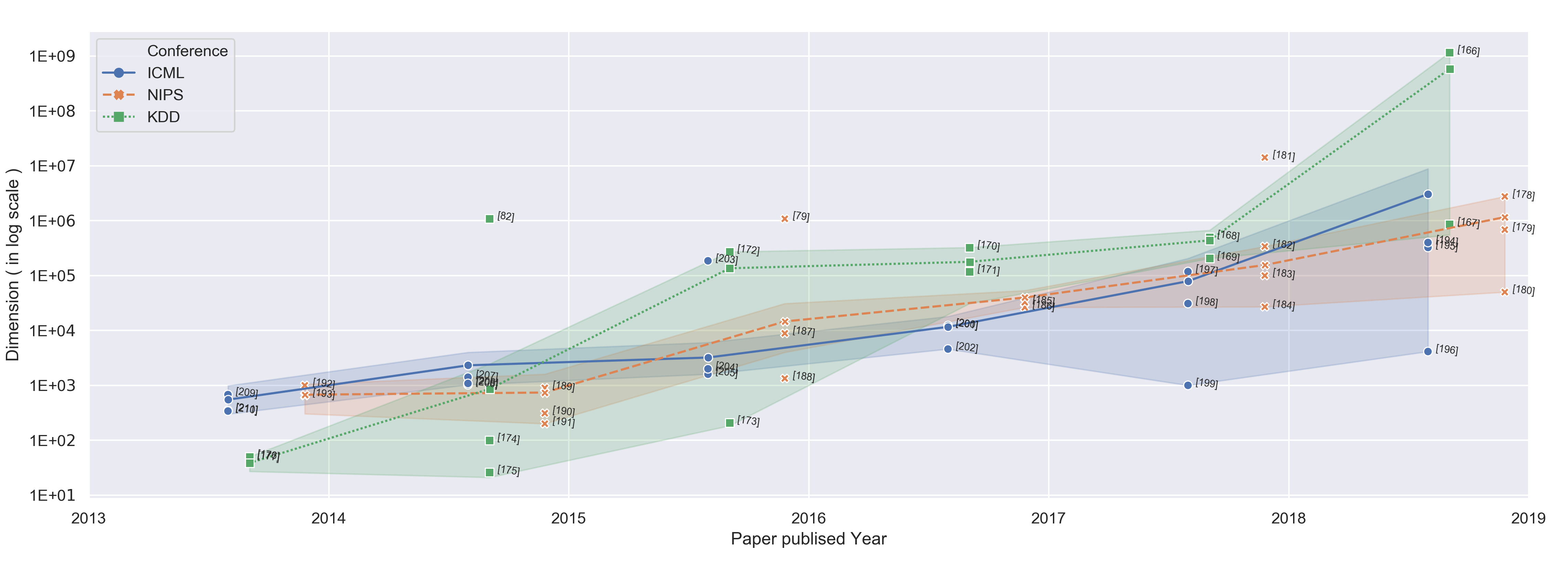}
	\caption{\label{output_dimensions_conf} Output dimension trends from papers published in the conferences ICML, NIPS, and KDD since 2013 \cite{DBLP:conf/kdd/DizajiWH18,DBLP:conf/kdd/LeeGZ18,DBLP:conf/kdd/YenHDRDX17,DBLP:conf/kdd/Tagami17,DBLP:conf/kdd/JainPV16,DBLP:conf/kdd/XuT016,DBLP:conf/kdd/KuoWWCYD15,DBLP:conf/kdd/Papagiannopoulou15,DBLP:conf/kdd/PrabhuV14,DBLP:conf/kdd/WuZTH014,DBLP:conf/kdd/ZhaiXW14,DBLP:conf/kdd/KongCY13,DBLP:conf/kdd/WangS13,DBLP:conf/nips/HongYHL18,DBLP:conf/nips/WydmuchJKBD18,DBLP:conf/nips/PanYLZZCH18,DBLP:conf/nips/RacahBMKPP17,DBLP:conf/nips/HuHS17,DBLP:conf/nips/JoshiAPIM17,DBLP:conf/nips/NamMKF17,DBLP:conf/nips/RosenfeldG16,DBLP:conf/nips/YuRD16,DBLP:conf/nips/BhatiaJKVJ15,DBLP:conf/nips/BengioVJS15,DBLP:conf/nips/RaiHHC15,DBLP:conf/nips/WuPKP14,DBLP:conf/nips/NguyenBRC14,DBLP:conf/nips/HoffmanGTHDGDS14,DBLP:conf/nips/XuJZ13,DBLP:conf/nips/CisseUAG13,DBLP:conf/icml/SibliniMK18,DBLP:conf/icml/YenKYHKR18,DBLP:conf/icml/WehrmannCB18,DBLP:conf/icml/SiZKMDH17,DBLP:conf/icml/JainMR17,DBLP:conf/icml/ZhangZ17,DBLP:conf/icml/LiWPA16,DBLP:conf/icml/YenHRZD16,DBLP:conf/icml/CisseAB16,DBLP:conf/icml/ParkNZSD15,DBLP:conf/icml/Hernandez-Lobato15,DBLP:conf/icml/HuangWSLC15,DBLP:conf/icml/Yu0KD14,DBLP:conf/icml/LinDH014,DBLP:conf/icml/LiZ14,DBLP:conf/icml/BiK13,DBLP:conf/icml/KolmogorovT13,DBLP:conf/icml/XiaoG13}.}
\end{figure*}

\subsection{Volume - Extreme Output Dimensions}
Large-scale datasets are ubiquitous in real-world applications. A dataset is defined to be large-scale if it meets one of three criteria: it has a large number of data instances, the input feature space has high dimensionality, or the output space has high dimensionality. Many studies have sought to solve the scalability issues caused by a large number of data instances, e.g., the instance selection method in \cite{brighton2002advances}, or with high-dimensional feature spaces, such as the feature selection method in \cite{DBLP:journals/cim/ZhaiOT14}. However, the issues associated with high output dimensions have received much less attention.

Consider, for example, that if the label for each dimension of an $m$-dimensional output vectors can be selected from a label set with $c$ different labels, then the number of output outcomes is $c^m$. Hence, these ultra-high-output dimensions/labels result in an extremely large output space and, in turn, high computation costs. Therefore, it is crucial to design multi-output learning models that can handle the immense and ongoing growth in outputs.

An analysis of the current state-of-the-art research on ultra-high-output dimensions revealed some interesting insights. Our analysis was based on the datasets used in studies of multiple disciplines, such as machine learning, computer vision, natural language processing, information retrieval, and data mining. We specifically focused on articles in three top journals and three top international conferences: IEEE Transactions on Pattern Analysis and Machine Intelligence (TPAMI), IEEE Transactions on Neural Networks and Learning Systems (TNNLS), the Journal of Machine Learning Research (JMLR), the International Conference on Machine Learning (ICML), the Conference on Neural Information Processing Systems (NIPS), and the Conference on Knowledge Discovery and Data Mining (KDD). Fig.~\ref{output_dimensions_journal} and Fig.~\ref{output_dimensions_conf} summarize our review. From these two figures, it is evident that the output dimensionality of the under-studied algorithms has continued to increase over time. In addition, the latest papers to address this issue in all selected titles are now dealing with more than a million output dimensions and, in some cases, are approaching billions of outputs. Moreover, the statistics for the conferences with shorter time-lags to publication demonstrate just how rapidly output dimensionality is increasing. From this analysis, we conclude that the explosion in output dimensionality is driving many developments in multi-output learning algorithms.

The studies we reviewed tend to fall into two categories: qualitative and quantitative approaches. The qualitative approaches generally involve generative models, while the quantitative models generally involve discriminative models. The main difference between the two models is that generative models focus on learning the joint probability $P(x,y)$ of the inputs $x$ and the label $y$, while the discriminative models focus on the posterior $P(y|x)$. Note that in a generative model, $P(x,y)$ can be used to generate some data $x$, where, in this case, $x$ is the generated output in this particular case.

\subsubsection{Qualitative Approaches/Generative Models} 

The aim of image synthesis \cite{DBLP:conf/icml/ReedAYLSL16,DBLP:conf/nips/ReedAMTSL16} is to synthesize new images from textual image descriptions of the image.
Some pioneering researchers have synthesized images using a GAN with the image distribution as multiple outputs \cite{DBLP:conf/nips/GoodfellowPMXWOCB14}. But, in real life, GANs can only generate low-resolution images.
However, since the first attempts at this foray, there has been progress in scaling up GANs to generate high-resolution images with sensible outputs. For example, Reed et al. \cite{DBLP:conf/icml/ReedAYLSL16} proposed a GAN architecture that generates visually plausible 64 x 64 pixel images given text descriptions. In a follow-up study, they presented GAWWN \cite{DBLP:conf/nips/ReedAMTSL16}, which scales the synthesized image up to 128 x 128 resolution by leveraging additional annotations. Subsequently, StackGAN \cite{DBLP:conf/iccv/ZhangXL17} was proposed, which is capable of generating photo-realistic images at a 256 x 256 resolution from text descriptions. HDGAN \cite{DBLP:journals/corr/abs-1802-09178} is the current state-of-the-art in image synthesis. It models high-resolution images in an end-to-end fashion at 512 x 512 pixels. Inevitably, the future will see further increases in resolution.

MaskGAN \cite{DBLP:journals/corr/abs-1801-07736} use GAN to generate text (i.e., meaningful word sequences). The label set size accords with the vocabulary size. The output dimension is the length of the word sequence that is generated, which, technically, can be unlimited. However, MaskGAN only handles sentence-level text generation. Document-level and book-level text generations are still challenging.

\subsubsection{Quantitative Approaches/Discriminative Models}

Like instance and feature selection methods, which reduce the number of input instances and, in turn, reduce input dimensionality, it is natural to design models that similarly reduce output dimensionality. Embedding methods can be used to compress a space by projecting the original space onto a lower-dimensional space, with the expected information preserved, such as label correlations and neighborhood structure. Popular methods, such as random projections or canonical correlation analysis projections \cite{DBLP:conf/nips/ChenL12,DBLP:conf/nips/HsuKLZ09,DBLP:journals/neco/TaiL12,DBLP:conf/nips/KapoorVJ12}, can be adopted to reduce the dimensions of the output label space. As a result, these modeling tasks can be performed on a compressed output label space and then the predicted compressed label can be projected back onto the original high-dimensional label space. Recently, several embedding methods have been proposed for extreme output dimensions. Mineiro and Karampatziakis \cite{DBLP:conf/pkdd/MineiroK15} proposed a novel randomized embedding for extremely large output spaces. AnnexML \cite{DBLP:conf/kdd/Tagami17} is another novel embedding method for graphs that captures graph structures in the embedding space. The embeddings are constructed from the k-nearest neighbors of the label vectors, and the predictions are made efficiently through an approximate nearest neighbor search method. Two popular ANN methods for handling extreme output dimensions are fastText learn tree \cite{DBLP:conf/icml/JerniteCS17} and XML-CNN \cite{DBLP:conf/sigir/LiuCWY17}.
FastText learn tree \cite{DBLP:conf/icml/JerniteCS17} jointly learns the data representation and the tree structure, and the learned tree structure is then used for efficient hierarchical prediction. XML-CNN is a CNN-based model that incorporates a dynamic max pooling scheme to capture fine-grained features from regions of the input document. A hidden bottleneck layer is used to reduce the model size.

\subsection{Variety - Complex Structures}
With the increasing abundance of labels, there is a pressing need to understand their inherent structures. Complex output structures can lead to multiple challenges in multi-output learning. For instance, it is common for strong correlations and complex dependencies to exist between labels. Therefore, appropriately modeling output dependencies in the label representation is critical but non-trivial in multi-output learning. In addition, designing a multi-variate loss function and proposing an efficient algorithm to alleviate the high complexity caused by complex structures is also challenging.

\subsubsection{Appropriate Modeling of Output Dependencies}
The simplest method of multi-output learning is to decompose the learning problem into $m$ independent single-output problems with each corresponding to a single value in the output space. A representative approach is binary relevance (BR) \cite{DBLP:journals/pr/BoutellLSB04}, which independently learns binary classifiers for all the labels in the output space. Given an unseen instance $\mathbf{x}$, BR predicts the output labels by predicting each of the binary classifiers and then aggregating the predicted labels. However, such independent models do not consider the dependencies between outputs. A set of predicted output labels might be assigned to the testing instance even though these labels never co-occur in the training set. Hence, it is crucial to model the output dependencies appropriately to obtain better performance for multi-output tasks.

Many classic learning methods have been proposed to model multiple outputs with interdependencies. These include label powersets (LPs) \cite{DBLP:reference/dmkdh/2010}, classifier chains (CC) \cite{DBLP:conf/pkdd/ReadPHF09,DBLP:journals/ml/ReadPHF11}, structured SVMs (SSVM) \cite{DBLP:journals/jmlr/TsochantaridisJHA05}, conditional random fields (CRF) \cite{DBLP:conf/icml/LaffertyMP01} and etc. LPs model the output dependencies by treating each different combination of labels in the output space as a single label, which transforms the problem into one of learning multiple single-label classifiers. The number of single-label classifiers to be trained is the number of label combinations, which grows exponentially with the number of labels. Therefore, LP has the drawback of high computation cost when training with a large number of output labels. Random k-labelsets \cite{DBLP:conf/ecml/TsoumakasV07}, an ensemble of LP classifiers, is a variant of LP that alleviates the computational complexity problem by training each LP classifier on a different random subset of labels.

CC improves BR by taking the output correlations into account. It links all the binary classifiers from BR into a chain via a modified feature space. Given the $j$th label, the instance $\mathbf{x}_i$ is augmented with the 1st, 2nd, ... $(j-1)$th label, i.e., $(\mathbf{x}_i, l_1, l_2,...,l_{j-1})$, as the input, to train the $j$th classifier. Given an unseen instance, CC predicts the output using the 1st classifier, and then augments the instance with the prediction from the 1st classifier as the input to the 2nd classifier for predicting the next output. CC processes values in this way from the 1st classifier to the last and so preserves the output correlations. However, a different order of chains leads to different results. ECC \cite{DBLP:conf/pkdd/ReadPHF09}, an ensemble of CC, was proposed to solve this problem. It trains the classifiers over a set of random ordering chains and averages the results. Probabilistic classifier chains (PCCs) \cite{DBLP:conf/icml/DembczynskiCH10} provide a probabilistic interpretation of CC by estimating the joint distribution of the output labels to capture the output correlations. CCMC \cite{DBLP:journals/jmlr/LiuTM17} is a classifier chain model that considers the order of label difficulties to reduce the degradation in performance caused by ambiguous labels. It is an easy-to-hard learning paradigm that identifies easy and hard labels and uses the predictions for easy labels to help solve the harder labels.

SSVM leverages the idea of large margins to deal with multiple interdependent outputs. The compatibility function is defined as $F(\mathbf{x},\mathbf{y}) = \mathbf{w}^T\Phi(\mathbf{x},\mathbf{y})$, where $\mathbf{w}$ is the weight vector and $\Phi: \mathcal{X} \times \mathcal{Y} \rightarrow \mathds{R}^q$ is the joint feature map over input and output pairs. The SSVM aims to find the classifier $h_{\mathbf{w}}(\mathbf{x}) = \arg \max\limits_{\mathbf{y}\in \mathcal{Y}}  \left<\mathbf{w},\phi(\mathbf{x},\mathbf{y})\right>$ with the following objective
 \begin{equation}\label{0}
 \vspace{-4mm}
\begin{split}
&  \min_{\mathbf{w}\in R^{q},\{ \xi_i \geq 0 \}^n_{i=1}} \frac{\lambda}{2}\|\mathbf{w}\|^{2} \\
&+\frac{C}{n}\sum^n_{i=1} \underbrace{\max\limits_{\mathbf{y}\in \mathcal{Y}}\{\Delta(\mathbf{y}_i,\mathbf{y}) + \mathbf{w}^T\Phi(\mathbf{x}_i,\mathbf{y})\} - \mathbf{w}^T\Phi(\mathbf{x}_i,\mathbf{y}_i)}_{\text{structured hinge loss}}\end{split} \nonumber\vspace{-2mm} 
\end{equation} 
Constraining the structured hinge loss with $\Delta(\mathbf{y}_i,\mathbf{y}) + \mathbf{w}^T\Phi(\mathbf{x}_i,\mathbf{y}) - \mathbf{w}^T\Phi(\mathbf{x}_i,\mathbf{y}_i) \leq \xi_i$, for all $y \in \mathcal{Y}$, the objective can be reformulated as
\begin{equation}\label{1}
\begin{split}
&  \min_{\mathbf{w}\in R^{q},\{ \xi_i \geq 0 \}^n_{i=1}}  \frac{\lambda}{2}\|\mathbf{w}\|^{2} +\frac{C}{n}\sum^n_{i=1} \xi_i^2 \\
&s.t.\ \mathbf{w}^T\Phi(\mathbf{x}_i,\mathbf{y}_i) - \mathbf{w}^T\Phi(\mathbf{x}_i,\mathbf{y}) \geq \Delta(\mathbf{y}_i,\mathbf{y})- \xi_i, \\
&\quad \quad \forall \mathbf{y} \in \mathcal{Y} \setminus \mathbf{y}_i, \forall i.\end{split}
\end{equation} 
where $\Delta: \mathcal{Y} \times \mathcal{Y} \rightarrow \mathds{R}$ is a loss function, $C$ is a positive constant that controls the trade-off between the training error minimization and the margin maximization~\cite{tsochantaridis2005large}, $n$ is the number of training samples and $\xi_i$ is the slack variable. In practice, SSVM is solved with the cutting-plane algorithm \cite{DBLP:journals/ml/JoachimsFY09}.

Apart from the classic models that learn the correlations between output, some of the state-of-the-art multi-output learning models are based on ANNs. For example, models based on convolutional neural networks typically focus on hierarchical multi-labels \cite{DBLP:conf/bionlp/BakerK17} or rankings \cite{DBLP:conf/cvpr/ChenZDLR17}. Recurrent neural network (RNNs) models generally focus on sequence-to-sequence learning \cite{DBLP:conf/nips/SutskeverVL14} and time-series prediction \cite{DBLP:journals/ijon/SmithJ14}. Generative deep neural networks are used to generate output data, such as images, text, and audio~\cite{DBLP:conf/nips/GoodfellowPMXWOCB14}.

\subsubsection{Multivariate Loss Functions}
Various loss functions were defined to compute the difference between the groundtruth and the predicted output. Different loss functions presents different errors given the same dataset, and they greatly affect the performance of the model.

\begin{description}[style=unboxed,leftmargin=0cm]
\item[0/1 loss] is a standard loss function that is commonly used in classification \cite{DBLP:conf/acl/Och03}:	
\begin{equation}\label{zero_one_loss}
\begin{split}
L_{0/1}(\mathbf{y},\mathbf{y}') = \mathds{I}(\mathbf{y} \neq \mathbf{y}')
\end{split}
\end{equation}
where $\mathds{I}$ is the indicator function.
In general, 0/1 loss refers to the number of misclassified training examples.
However, it is very restrictive and does not consider label dependency. Therefore, it is not suitable for large numbers of outputs or for outputs with complex structures. In addition, it is non-convex and non-differentiable, so it is difficult to minimize the loss using standard convex optimization methods. In practice, one typically uses a surrogate loss, which is a convex upper bound of the task loss.  However, a surrogate loss in multi-output learning usually loses the consistency when generalizing single-output methods to deal with multiple outputs \cite{gao2011consistency}. Several works on subfields of multi-output learning study the consistency of different surrogate functions and show that they are consistent under some sufficient conditions
\cite{DBLP:journals/jmlr/TewariB07,DBLP:conf/nips/McAllesterK11}. Yet this is still a challenging aspect of multi-output learning. More exploration on the theoretical consistency of different problems is required.

Below, we describe four popular surrogate losses: hinge loss, negative log loss, perceptron loss, and softmaxmargin loss.

\item[Hinge loss] is one of the most widely used surrogate losses and is usually used in structured SVMs \cite{DBLP:conf/nips/TaskarGK03}. It pushes the score of the correct outputs to be greater than that of the prediction:
\begin{equation}\label{hinge_loss}
\begin{split}
L_{Hinge}(\mathbf{x},\mathbf{y},\mathbf{w}) = \max_{\mathbf{y}' \in \mathcal{Y}}[\Delta(\mathbf{y},\mathbf{y}') + \mathbf{w}^T\Phi(\mathbf{x},\mathbf{y}')] - \mathbf{w}^T\Phi(\mathbf{x},\mathbf{y})
\end{split}
\end{equation}
The margin, $\Delta(\mathbf{y},\mathbf{y}')$, has different definitions based on the output structures and task. For example, for sequence learning or outputs with equal weights, $\Delta(\mathbf{y},\mathbf{y}')$ can be simply defined as the Hamming loss $\sum_{j=1}^{m}\mathds{I}(\mathbf{y}_{(j)} \neq \mathbf{y}'_{(j)})$.
For taxonomic classification with the hierarchical output structure, $\Delta(\mathbf{y},\mathbf{y}')$ can be defined as the tree distance between $\mathbf{y}$ and $\mathbf{y}'$ \cite{DBLP:conf/cikm/CaiH04}.
For ranking, $\Delta(\mathbf{y},\mathbf{y}')$ can be defined as the mean average precision of a ranking $\mathbf{y}'$ compared to the optimal $\mathbf{y}$ \cite{DBLP:conf/sigir/YueFRJ07}.
In syntactic parsing, $\Delta(\mathbf{y},\mathbf{y}')$ is defined as the number of labeled spans where $\mathbf{y}$ and $\mathbf{y}'$ do not agree \cite{taskar2004max}.
Non-decomposable losses, such as the $F_1$ measure, average precision (AP), or intersection over union (IOU), can also be defined as a margin.

\item[Negative log loss] is commonly used in CRFs \cite{DBLP:conf/icml/LaffertyMP01}. Note that minimizing negative log loss is the same as maximizing the log probability of the data.
\begin{equation}\label{log_loss}
\begin{split}
L_{NegativeLog}(\mathbf{x},\mathbf{y},\mathbf{w}) = \log\sum_{\mathbf{y}' \in \mathcal{Y}}\exp&[\mathbf{w}^T\Phi(\mathbf{x},\mathbf{y}')] \\
&- \mathbf{w}^T\Phi(\mathbf{x},\mathbf{y})
\end{split}
\end{equation} 
\item[Perceptron loss] is usually adopted in structured perceptron tasks  \cite{DBLP:conf/emnlp/Collins02} and is the same as hinge loss without the margin.
\begin{equation}\label{perceptron_loss}
\begin{split}
L_{Perceptron}(\mathbf{x},\mathbf{y},\mathbf{w}) = \max_{\mathbf{y}' \in \mathcal{Y}}[\mathbf{w}^T\Phi(\mathbf{x},\mathbf{y}') - \mathbf{w}^T\Phi(\mathbf{x},\mathbf{y})]
\end{split}
\end{equation} 
\item[Softmax-margin loss] is one of the most popular loss functions in multi-output learning models such as SSVMs~\cite{DBLP:conf/icassp/PoveyKKRSV08} and CRFs~\cite{DBLP:conf/naacl/GimpelS10}.
\begin{equation}\label{softmax_margin_loss}
\begin{split}
L_{SoftmaxMargin}(\mathbf{x},\mathbf{y},\mathbf{w}) = &\log\sum_{\mathbf{y}' \in \mathcal{Y}}\exp[\Delta(\mathbf{y},\mathbf{y}') + \\ &\mathbf{w}^T\Phi(\mathbf{x},\mathbf{y}')] - \mathbf{w}^T\Phi(\mathbf{x},\mathbf{y})
\end{split}
\end{equation}
\item[Squared loss] is a popular and convenient loss function that quadratically penalizes the difference between the ground truth and the prediction. It is commonly used in traditional single-output learning and can be easily extended to multi-output learning by summing the squared differences over all the outputs:
\begin{equation}\label{squared_loss}
\begin{split}
L_{Squared}(\mathbf{y},\mathbf{y}') = (\mathbf{y} - \mathbf{y}')^2
\end{split}
\end{equation}
In multi-output learning, it is usually used with continuous valued outputs or continuous intermediate results before converting them into discrete valued outputs. It is also commonly used in neural networks and boosting.
\end{description}

\subsubsection{Efficient Algorithms}
Complex output structures significantly increase the burden on algorithms to formulate a model. Large-scale outputs, complex output dependencies, and/or complex loss functions can all be problematic. Therefore, several algorithms have been proposed specifically to tackle these challenges efficiently. Many leverage classic machine learning models so as to speed up the algorithms and alleviate the burden of complexity. The four most widely used classic models are based on $k$ nearest neighbor ($k$NN), decision trees, $k$-means, and hashing.

\begin{enumerate}
\item $k$NN-based methods are simple yet powerful machine learning models. Predictions are made based on the closest $k$ instances to the test instance vector in terms of Euclidean distance. LMMO-$k$NN~\cite{8263148} is an SSVM-based model involving an exponential number of constraints \emph{w.r.t.} the number of labels. This model imposes $k$NN constraints instantiated by the label vectors from neighboring examples to significantly reduce the training time and make rapid predictions.
\item Decision tree based methods \cite{DBLP:conf/nips/DengSBL11,DBLP:conf/iccv/GaoK11} learn a tree from the training data with a hierarchical output label space. They recursively partition the nodes until each leaf contains a small number of labels. Each novel data point is passed down the tree until it reaches a leaf. This method usually achieves a logarithmic time prediction.
\item $k$-means based methods such as SLEEC \cite{DBLP:conf/nips/BhatiaJKVJ15} cluster the training data using $k$-means clustering. SLEEC learns a separate embedding per cluster and performs classification for a novel instance within its cluster alone. This significantly reduces the prediction time.
\item Hashing methods, such as co-hashing \cite{DBLP:conf/aaai/0001LTSO18,DBLP:journals/tnn/ShenLTSO18} and DBPC \cite{DBLP:conf/ijcai/0001LLOT18}, reduce the prediction time by using hashing on the input or the intermediate embedding space. Co-hashing learns an embedding space to preserve semantic similarity structures between inputs and outputs. Compact binary representations are then generated for the learned embeddings for prediction efficiency. DBPC jointly learns a deep latent Hamming space and binary prototypes while capturing the latent nonlinear structures of the data with an ANN. The learned Hamming space and binary prototypes significantly decrease the prediction complexity and reduce memory/storage costs.
\end{enumerate}

\subsection{Volume - Extreme Class Imbalances}
% Mode collapse problem. E.g. In face generation, the learned systems tend to generate famous faces that dominated by the model. 

Real-world multi-output applications rarely provide data with an equal number of training instances for all labels/classes. Too many instances in one class over another mean the data is imbalanced, and this is common in many applications. Therefore, traditional models learned from such data tend to favor majority classes more. For example, in face generation, a trained model tends to generate the faces of famous people because there are so many more images of celebrities than other people. Though class imbalance problems have been studied extensively in the context of binary classification, this issue still remains a challenge in multi-output learning, especially with extreme imbalances.

Many studies on multi-output learning either create a balanced dataset or ignore the problems introduced by imbalanced data. A natural way to balance class distributions is to resample the dataset. There are two main resampling techniques: undersampling and oversampling~\cite{more2016survey}. Undersampling methods down-size the majority classes. The NearMiss family of methods~\cite{mani2003knn} are representative works of this category. 
The  oversampling methods,  such  as  SMOTE  and  its  variants~\cite{DBLP:journals/jair/ChawlaBHK02},  adopt oversampling  technique  on  minority  classes  to  handle  the imbalanced  class  learning  problem.  However,  all these resampling methods are mainly designed for single output learning problems. There are other techniques to handle class imbalance in multi-output learning tasks with ANN.

For example, Dong et al. \cite{DBLP:journals/corr/abs-1804-10851} combined incremental rectification of mini-batches with a deep neural network. Then a hard sample mining strategy minimizes the dominant effect of the majority classes by discovering the boundaries of sparsely-sampled minority classes. Both of the methods in  \cite{DBLP:journals/corr/abs-1811-10419} and \cite{DBLP:journals/corr/abs-1811-08812} leveraged adversarial training to mitigate imbalance by using a re-weighting technique so that majority classes tend to have a similar impact as minority classes.

\subsection{Volume - Unseen Outputs}
Traditional multi-output learning assumes that the output set in testing is the same as the one in training, i.e., the output labels of a testing instance have already appeared during training. However, this may not be true in real-world applications. For example, a new emerging living species can not be detected using a learned classifier based on existing living animals. Similarly, it is infeasible to recognize the actions or events in a real-time video if no such actions or events with the same labels appeared in the training video set. Nor could a coarse animal classifier provide details of the species of a detected animal, such as whether a dog is a labrador or a shepherd.

Depending on the complexity of the learning task, label annotation is usually very costly. In addition, the enormous growth in the number labels not only leads to high-dimensional output space as a result of computation inefficiency, but also makes supervised learning tasks challenging due to unseen output labels during testing.

\subsubsection{Zero-shot Multi-label Classification}
Multi-label classification is a typical multi-output learning problem. Multi-label classification problems can have various inputs, such as text, images, and video, depending on the application. The output for each input instance is usually a binary label vector, indicating what labels are associated with the input. Multi-label classification problems learn a mapping from the input to the output. However, as the label space increases, it is common to find unseen output labels during testing, where no such labels have appeared in the training set. To study such cases, the zero-shot multi-class classification problem was first proposed in 
\cite{DBLP:conf/nips/PalatucciPHM09,DBLP:conf/icml/Romera-ParedesT15} and most leverage the predefined semantic information, such as attributes \cite{DBLP:conf/cvpr/LampertNH09}, word representations \cite{DBLP:conf/nips/MikolovSCCD13} and etc. This technique was then extended to zero-shot multi-label classification to assign multiple unseen labels to an instance. Similarly, zero-shot multi-label learning leverages the knowledge of the seen and unseen labels and models the relationships between the input features, label representations, and labels. For example, Gaure et al. \cite{gaure2017probabilistic} leverage the co-occurrence statistics of seen and unseen labels and model the label matrix and co-occurrence matrix jointly using a generative model. Rios and Kavuluru \cite{DBLP:conf/emnlp/RiosK18} and Lee et al.  \cite{DBLP:conf/cvpr/LeeFYW18} incorporate knowledge graphs of the label relationships with neural networks.

\subsubsection{Zero-shot Action Localization}
Similar to zero-shot classification problems, localizing human actions in videos without any training video examples is a challenging task. Inspired by zero-shot image classification, many studies into zero-shot action classification predict unseen actions from disjunct training actions based on the prior knowledge of action-to-attribute mappings \cite{DBLP:conf/cvpr/GanYG16,DBLP:conf/cvpr/LiuKS11,DBLP:journals/paa/ZhangWXZL15}.
Such mappings are usually predefined and the seen and unseen actions are linked through a description of the attributes. Thus, they can be used to generalize undefined actions but are unable to localize actions.
More recently, some works are proposed to overcome the issue.
Jain \emph{et al.} \cite{DBLP:conf/iccv/JainGMS15} proposes Objects2action without using any video data or action annotations. It leverages vast object annotations, images and text descriptions that can be obtained from open-source collections such as WordNet and ImageNet. Mettes and Snoek \cite{DBLP:conf/iccv/MettesS17} have subsequently enhanced Objects2action by considering the relationships between actors and objects.

\subsubsection{Open-set Recognition}
Traditional multi-output learning problems, including zero-shot multi-output learning, operate under a closed-set assumption, i.e., where all the testing classes are known at the time of training time either through the training samples or because they are predefined in a semantic label space. However, Scheirer et al.  \cite{DBLP:journals/pami/ScheirerRSB13} proposed a concept called open-set recognition to describe a scenario where unknown classes appear in testing. Open-set recognition presents 1-vs-set machine to classify the known classes as well as deal with the unknown classes. In later studies \cite{DBLP:journals/pami/ScheirerJB14,DBLP:conf/eccv/JainSB14}, they extended this idea into to multi-class settings by formulating a compact abating probability model. Bendale and Boult \cite{DBLP:conf/cvpr/BendaleB16} adapted ANNs for open-set recognition by proposing a new model layer that estimates the probability of an input being an unknown class.

\begin{figure}
	\centering
	\includegraphics[scale=0.43]{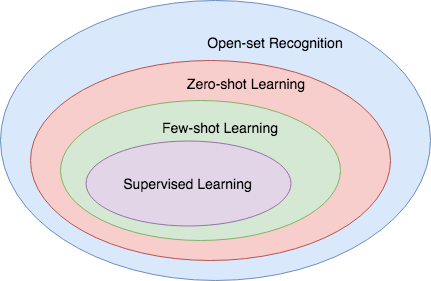}
	\caption{\label{unseen_outputs} Relationship among different levels of unseen outputs. All of these learning problems belong to multi-output learning.}
\end{figure}

Fig.~\ref{unseen_outputs} illustrates the relationships between different levels of unseen outputs in multi-output learning. Open-set recognition is the most generalized problem of all. Few-shot and zero-shot learning have studied with different multi-output learning problems, such as multi-label learning and event localization. However, open-set recognition has only been studied in conjunctions with multi-class classification. Other problems in the context of multi-output learning are still unexplored.

\begin{figure*}[h]
	\centering
	\includegraphics[scale=0.5]{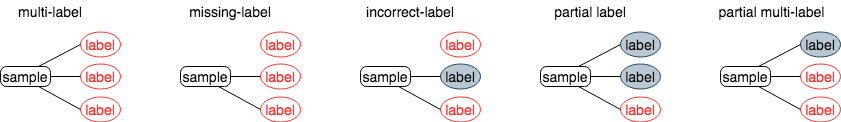}
	\caption{\label{noisy_labels} Range of noisy labels in multi-label classification. Training may be: \textbf{multi-label} (sample associates with multiple labels), \textbf{missing-label} (sample has incomplete label assignment), \textbf{incorrect-label} (sample has at least one incorrect labels and possible incomplete label assignment), \textbf{partial-label} (each sample has multiple labels, only one of which is correct), \textbf{partial multi-label} (each sample has multiple labels, at least one of which is correct). A line connecting a label with the sample represents that the sample associates with the label. The label in red color represents the correct label to the sample. The label in gray box represents an incorrect label to the sample.}
\end{figure*}

\subsection{Veracity - Noisy Output Labels}
Almost all methods of label annotation lead to some amount of noise for various reasons. Associations may be weak, the text may be ambiguous, crowdsourced workers may not be domain experts so labels may be incorrect \cite{DBLP:conf/iccv/LiYSCLL17}. Therefore, it is usually necessary to handle noisy outputs like missing, corrupt, incorrect, and/or partial labels, in real-world tasks.

\subsubsection{Missing Labels}
Often human annotators annotate an image or document with prominent labels but miss some of the less emphasized labels. Additionally, all the objects in an image may not be localized because there are, say, too many objects or the objects are too small. Social media, such as Instagram, allow users to tag uploaded images. But the tags could relate to anything: the type of event, the person's mood, the weather. Plus, no user is likely to tag every object or every aspect of an image. Directly using such labeled datasets in traditional multi-output learning models can not guarantee the performance of the given tasks.
Therefore, handling missing labels is necessary in real-world applications.

In early studies, missing labels were handled by treating them as negative labels  \cite{DBLP:conf/sdm/ChenSWZ08,DBLP:conf/aaai/SunZZ10,DBLP:conf/cvpr/BucakJJ11}. Then modeling tasks are performed based on a fully-labeled dataset. However, this approach can introduce undesirable bias into the learning problem. Therefore, a more widely-used method now is missing value imputation through matrix completion \cite{DBLP:conf/nips/CabralTCB11,DBLP:conf/nips/XuJZ13,DBLP:conf/nips/YuRD16}. Most of these approaches are based on a low-rank assumption and, more recently, on label correlations, which improves learning performance \cite{DBLP:conf/aaai/BiK14,DBLP:conf/eccv/YangZC16}.

\subsubsection{Incorrect Labels}
Many labels in high-dimensional output space are non-informative or simply wrong \cite{DBLP:conf/icdm/GongZYT17}. This is especially common with annotations from crowdsourcing platforms that hire non-expert workers. Labeled datasets from social media networks are also often less than useful. A basic approach for handling incorrect labels is to simply remove those samples \cite{DBLP:conf/sspr/BarandelaG00,DBLP:journals/jair/BrodleyF99}. That said, it is frequently difficult to detect which samples have been mislabeled. Therefore, designing multi-output learning algorithms that learn from noisy datasets is of great practical importance.

Existing multi-output learning methods handling noisy labels  generally fall into two groups. The first group is based on building robust loss functions \cite{DBLP:conf/eccv/JoulinMJV16,DBLP:journals/corr/ReedLASER14,DBLP:conf/cvpr/VeitACKGB17}, which modify the labels in the loss function to alleviate the effect of noise. The second group models latent labels and learns the transition from the latent to the noisy labels \cite{DBLP:conf/icml/MnihH12,DBLP:conf/icdm/JindalNC16,DBLP:journals/corr/abs-1711-00583}.

\textbf{Partial Labels}
A special case of incorrect labels is partial labels \cite{DBLP:conf/apsipa/MaoCLJ16,chai2019tnn,DBLP:journals/tcyb/GongLTYYT18}, where each training instance is associated with a set of candidate labels but only one of them is correct. This is a common problem in real-world applications. For example, a photograph might contain many faces with captions listing who is in the photo but the names are not matched to the face. Many methods for learning partial labels have been developed to recover the ground-truth labels from a candidate set \cite{DBLP:journals/ml/YuZ17,DBLP:journals/tkde/ZhangYT17}.
However, most are based on the assumption of exactly one ground truth for each instance, which may not always hold true by different label annotation methods.
With the use of multiple workers on the crowdsourcing platform to annotate a dataset, the final annotations are usually gathered from the union set of the annotations of all the workers, where each instance might associate with both multiple relevant and irrelevant labels.
Hence, Xie and Huang \cite{DBLP:conf/aaai/XieH18} developed a new learning framework, partial multi-label learning (PML), that relaxes this assumption by leveraging the data structure information to optimize the confidence weighted rank loss. Fig.~\ref{noisy_labels} summarizes all the scenarios with noisy output labels, including multi-label learning, missing labels, incorrect labels, partial label learning, and partial multi-label learning.

\subsection{Velocity - Changes in Output Distribution}
Many real-world applications must deal with data streams, where data arrives continuously and possibly endlessly. In these cases, the output distributions can change over time or concept drift can occur. Streaming data is common in surveillance \cite{DBLP:conf/cvpr/OhHPCCLMALDSWJRSVPRYTSFRD11}, driver route prediction \cite{kaggletaxiroute2015}, demand forecasting \cite{ElectricityLoadDiagrams20112014}, and many other applications.
Take visual tracking \cite{DBLP:conf/cvpr/NamH16} in surveillance video as an example, where the video stream is potentially endless. Data streams come in high velocity as the video keeps generating consecutive frames. The goal is to detect, identify, and locate events or objects in the video. Therefore, the learning model must adapt to possible concept drift while working with limited memory.

Existing multi-output learning methods model changes in output distribution by updating the learning system each time data streams arrive. The update method might be ensemble-based \cite{avidan2007ensemble,DBLP:conf/acml/QuZZQ09,DBLP:conf/kdd/BifetHPKG09,DBLP:conf/colcom/KongY11,DBLP:conf/cikm/BuyukcakirBC18} or ANN-based methods \cite{DBLP:conf/aaai/MilanRD0S17,DBLP:conf/cvpr/NamH16}.
Other strategies to handle concept drift include: the assumption of a fading effect on past data \cite{DBLP:conf/colcom/KongY11}; maintaining a change detector on predictive performance measurements and recalibrating models accordingly \cite{DBLP:conf/kdd/BifetHPKG09,DBLP:journals/ml/ReadBHP12}; and using stochastic gradient descent to update the network and accommodate new data streams with an ANN \cite{DBLP:conf/cvpr/NamH16}.
Notably, the $k$ neareast neighbor ($k$NN) is one of the most classic frameworks in handling multi-output problems, but it cannot be successfully adapted to deal with the challenge of change of output distribution due to the inefficiency issue. Many online hashing and online quantization based methods \cite{DBLP:journals/tnn/HuangYZ18,DBLP:journals/tkde/XuTZ18} are proposed to improve the efficiency of $k$NN while accommodating the changing output distribution. 

\subsection{Other Challenges}
Any two of the aforementioned challenges can be combined to form a more complex challenge. For example, noisy labels and unseen outputs can be combined to form an open-set noisy label problem \cite{DBLP:conf/cvpr/WangLMBZSX18}. In addition, the combination of noisy labels and extreme output dimensions are also worthy of study and further exploration \cite{DBLP:conf/icml/Yu0KD14}. Changes in output distribution together with noisy labels result in online time-series prediction problems with missing values \cite{DBLP:conf/icml/AnavaHZ15}, while changes in distributions combined with dynamic label sets (unseen outputs) lead to open-world recognition problems with incremental labels~\cite{DBLP:conf/cvpr/BendaleB15}. Changing output distribution with extreme class imbalances create the common problem of streaming data with concept drift and class imbalances at the same time~\cite{DBLP:journals/pai/HoensPC12,DBLP:conf/ijcai/XioufisSTV11}. Moreover, the combination of complex output structures with changing output distribution is also frequent in real-world applications~\cite{DBLP:conf/sigir/RenPLDR14}.

\subsection{Open Challenges}
\subsubsection{Output Label Interpretation}
There are different ways to represent output labels and each expresses label information from a specific perspective. Taking label tags as an output for example, binary attributed output embeddings represent what attributes the input relates to. Hierarchical label output embedding conveys the hierarchical structure of the inputs. Semantic word output embeddings reflect the semantic relationships between the outputs. As one can see, each exhibits a certain level of human interpretability. Hence, an emerging approach to label embedding is to incorporate different label information from multiple perspectives and rich contexts to enhance interpretability \cite{DBLP:conf/aaai/ShiXPTP19}. 
This is a challenging undertaking because it is quite difficult to appropriately model the interdependencies between outputs in a way that humans can easily interpret and understand. For example, an image of a centaur is expected to be described with semantic labels like horse and person. Moreover, the image is expected to be described with attributes like head, arm, tail, etc. As such, appropriately modeling the relationships between input and outputs with rich interpretations of the labels is an open challenge that should be explored in future studies.

\subsubsection{Output Heterogeneity}
As the demand for sophisticated decision making increases, so does demand for outputs with more complex structures. Returning to the example of surveillance, people re-identification in traditional approaches usually consists of two steps: people detection, then re-identifying that person if they are input. These steps are essentially two separate tasks that need to be learned together if performance is to be enhanced. Several researchers have recently attempted this demanding challenge, i.e., building a model that can simultaneously learn multiple tasks with different outputs. Mousavian et al. \cite{DBLP:conf/3dim/MousavianPK16} undertook joint people detection in tandem with re-identification, while Van Ranst et al. \cite{van2018fast} tackled image segmentation with depth estimation. However, more exploration and investigation to overcome this challenge is needed. As an example, one worthy undertaking would be to answer the question: Can we simultaneously learn the representation of a new user in a social network as well as their potential links to existing users?

\section{Conclusion}
Multi-output learning has attracted significant attention over the last decade. This paper provides a comprehensive review of the study of multi-output learning using the 4 Vs as a frame. We explore the characteristics of the multi-output learning paradigm beginning with the life cycle of the output labels. We emphasize the issues associated with each step of the learning process. In addition, we provide an overview of the types of outputs, the structures, selected problem definitions, common model evaluation metrics, and the popular data repositories used in experiments, with representative works referenced throughout. The paper concludes with a discussion on the challenges caused by 4 Vs and some future research directions that are worthy of further study.

% if have a single appendix:
%\appendix[Proof of the Zonklar Equations]
% or
%\appendix  % for no appendix heading
% do not use \section anymore after \appendix, only \section*
% is possibly needed

% use appendices with more than one appendix
% then use \section to start each appendix
% you must declare a \section before using any
% \subsection or using \label (\appendices by itself
% starts a section numbered zero.)
%

% %\appendices
% %\section{Proof of the First Zonklar Equation}
% %Appendix one text goes here.

% you can choose not to have a title for an appendix
% if you want by leaving the argument blank
% %\section{}
% %Appendix two text goes here.

% use section* for acknowledgment
% %\section*{Acknowledgment}

% %The authors would like to thank...

% Can use something like this to put references on a page
% by themselves when using endfloat and the captionsoff option.
\ifCLASSOPTIONcaptionsoff
  \newpage
\fi

\bibliographystyle{IEEEtran}

% Generated by IEEEtran.bst, version: 1.13 (2008/09/30)
\begin{thebibliography}{100}
\providecommand{\url}[1]{#1}
\csname url@samestyle\endcsname
\providecommand{\newblock}{\relax}
\providecommand{\bibinfo}[2]{#2}
\providecommand{\BIBentrySTDinterwordspacing}{\spaceskip=0pt\relax}
\providecommand{\BIBentryALTinterwordstretchfactor}{4}
\providecommand{\BIBentryALTinterwordspacing}{\spaceskip=\fontdimen2\font plus
\BIBentryALTinterwordstretchfactor\fontdimen3\font minus
  \fontdimen4\font\relax}
\providecommand{\BIBforeignlanguage}[2]{{%
\expandafter\ifx\csname l@#1\endcsname\relax
\typeout{** WARNING: IEEEtran.bst: No hyphenation pattern has been}%
\typeout{** loaded for the language `#1'. Using the pattern for}%
\typeout{** the default language instead.}%
\else
\language=\csname l@#1\endcsname
\fi
#2}}
\providecommand{\BIBdecl}{\relax}
\BIBdecl

\bibitem{DBLP:journals/tkde/ZhangZ14}
M.~Zhang and Z.~Zhou, ``A review on multi-label learning algorithms,''
  \emph{TKDE}, vol.~26, no.~8, pp. 1819--1837, 2014.

\bibitem{DBLP:conf/aaai/GongTYL16}
C.~Gong, D.~Tao, J.~Yang, and W.~Liu, ``Teaching-to-learn and learning-to-teach
  for multi-label propagation,'' in \emph{AAAI}, 2016, pp. 1610--1616.

\bibitem{DBLP:journals/ijar/BielzaLL11}
C.~Bielza, G.~Li, and P.~Larra{\~{n}}aga, ``Multi-dimensional classification
  with bayesian networks,'' \emph{Int. J. Approx. Reasoning}, vol.~52, no.~6,
  pp. 705--727, 2011.

\bibitem{DBLP:books/daglib/p/VembuG10}
\BIBentryALTinterwordspacing
S.~Vembu and T.~G{\"{a}}rtner, ``Label ranking algorithms: {A} survey,'' in
  \emph{Preference Learning.}, 2010, pp. 45--64. [Online]. Available:
  \url{https://doi.org/10.1007/978-3-642-14125-6\_3}
\BIBentrySTDinterwordspacing

\bibitem{DBLP:journals/widm/BorchaniVBL15}
H.~Borchani, G.~Varando, C.~Bielza, and P.~Larra{\~{n}}aga, ``A survey on
  multi-output regression,'' \emph{Wiley Interdisciplinary Reviews: Data Mining
  and Knowledge Discovery}, vol.~5, no.~5, pp. 216--233, 2015.

\bibitem{widmer1996learning}
G.~Widmer and M.~Kubat, ``Learning in the presence of concept drift and hidden
  contexts,'' \emph{Machine learning}, vol.~23, no.~1, pp. 69--101, 1996.

\bibitem{ILSVRC15}
O.~Russakovsky, J.~Deng, H.~Su, J.~Krause, S.~Satheesh, S.~Ma, Z.~Huang,
  A.~Karpathy, A.~Khosla, M.~Bernstein, A.~C. Berg, and L.~Fei-Fei, ``{ImageNet
  Large Scale Visual Recognition Challenge},'' \emph{IJCV}, vol. 115, no.~3,
  pp. 211--252, 2015.

\bibitem{DBLP:journals/ijcv/RussellTMF08}
B.~C. Russell, A.~Torralba, K.~P. Murphy, and W.~T. Freeman, ``Labelme: {A}
  database and web-based tool for image annotation,'' \emph{IJCV}, vol.~77, no.
  1-3, pp. 157--173, 2008.

\bibitem{DBLP:conf/eacl/StenetorpPTOAT12}
P.~Stenetorp, S.~Pyysalo, G.~Topic, T.~Ohta, S.~Ananiadou, and J.~Tsujii,
  ``brat: a web-based tool for nlp-assisted text annotation,'' in \emph{EACL},
  2012.

\bibitem{DBLP:conf/acllaw/EryigitCYTC13}
G.~Eryigit, F.~S. {\c{C}}etin, M.~Yanik, T.~Temel, and I.~{\c{C}}i{\c{c}}ekli,
  ``{TURKSENT:} {A} sentiment annotation tool for social media,'' in
  \emph{LAW@ACL}, 2013, pp. 131--134.

\bibitem{DBLP:conf/cvpr/LampertNH09}
C.~H. Lampert, H.~Nickisch, and S.~Harmeling, ``Learning to detect unseen
  object classes by between-class attribute transfer,'' in \emph{CVPR}, 2009,
  pp. 951--958.

\bibitem{DBLP:conf/cvpr/RohrbachSS11}
M.~Rohrbach, M.~Stark, and B.~Schiele, ``Evaluating knowledge transfer and
  zero-shot learning in a large-scale setting,'' in \emph{CVPR}, 2011, pp.
  1641--1648.

\bibitem{DBLP:conf/nips/MikolovSCCD13}
T.~Mikolov, I.~Sutskever, K.~Chen, G.~S. Corrado, and J.~Dean, ``Distributed
  representations of words and phrases and their compositionality,'' in
  \emph{NIPS}, 2013, pp. 3111--3119.

\bibitem{deerwester1989computer}
S.~C. Deerwester, S.~T. Dumais, G.~W. Furnas, R.~A. Harshman, T.~K. Landauer,
  K.~E. Lochbaum, and L.~A. Streeter, ``Computer information retrieval using
  latent semantic structure,'' 1989.

\bibitem{xu2015show}
K.~Xu, J.~Ba, R.~Kiros, K.~Cho, A.~Courville, R.~Salakhudinov, R.~Zemel, and
  Y.~Bengio, ``Show, attend and tell: Neural image caption generation with
  visual attention,'' in \emph{International conference on machine learning},
  2015, pp. 2048--2057.

\bibitem{DBLP:journals/pami/ScheirerRSB13}
W.~J. Scheirer, A.~de~Rezende~Rocha, A.~Sapkota, and T.~E. Boult, ``Toward open
  set recognition,'' \emph{TPAMI}, vol.~35, no.~7, pp. 1757--1772, 2013.

\bibitem{DBLP:conf/nips/PalatucciPHM09}
M.~Palatucci, D.~Pomerleau, G.~E. Hinton, and T.~M. Mitchell, ``Zero-shot
  learning with semantic output codes,'' in \emph{NIPS}, 2009, pp. 1410--1418.

\bibitem{DBLP:journals/pai/HoensPC12}
T.~R. Hoens, R.~Polikar, and N.~V. Chawla, ``Learning from streaming data with
  concept drift and imbalance: an overview,'' \emph{Progress in {AI}}, vol.~1,
  no.~1, pp. 89--101, 2012.

\bibitem{DBLP:conf/cikm/CaiH04}
L.~Cai and T.~Hofmann, ``Hierarchical document categorization with support
  vector machines,'' in \emph{CIKM}, 2004, pp. 78--87.

\bibitem{DBLP:conf/cvpr/XiaoHEOT10}
J.~Xiao, J.~Hays, K.~A. Ehinger, A.~Oliva, and A.~Torralba, ``{SUN} database:
  Large-scale scene recognition from abbey to zoo,'' in \emph{CVPR}, 2010, pp.
  3485--3492.

\bibitem{DBLP:conf/mm/QiHRTMZ07}
G.~Qi, X.~Hua, Y.~Rui, J.~Tang, T.~Mei, and H.~Zhang, ``Correlative multi-label
  video annotation,'' in \emph{ACM Multimedia}, 2007, pp. 17--26.

\bibitem{DBLP:journals/apin/DzeroskiDG00}
S.~Dzeroski, D.~Demsar, and J.~Grbovic, ``Predicting chemical parameters of
  river water quality from bioindicator data,'' \emph{Appl. Intell.}, vol.~13,
  no.~1, pp. 7--17, 2000.

\bibitem{aras2004forecasting}
H.~Aras and N.~Aras, ``Forecasting residential natural gas demand,''
  \emph{Energy Sources}, vol.~26, no.~5, pp. 463--472, 2004.

\bibitem{li2017novel}
H.~Li, W.~Zhang, Y.~Chen, Y.~Guo, G.-Z. Li, and X.~Zhu, ``A novel multi-target
  regression framework for time-series prediction of drug efficacy,''
  \emph{Scientific reports}, vol.~7, p. 40652, 2017.

\bibitem{DBLP:conf/cvpr/GengX14}
X.~Geng and Y.~Xia, ``Head pose estimation based on multivariate label
  distribution,'' in \emph{CVPR}, 2014, pp. 1837--1842.

\bibitem{DBLP:conf/aaai/GengSZ10}
X.~Geng, K.~Smith{-}Miles, and Z.~Zhou, ``Facial age estimation by learning
  from label distributions,'' in \emph{AAAI}, 2010.

\bibitem{DBLP:conf/emnlp/ZhouZZZG16}
D.~Zhou, X.~Zhang, Y.~Zhou, Q.~Zhao, and X.~Geng, ``Emotion distribution
  learning from texts,'' in \emph{EMNLP}, 2016, pp. 638--647.

\bibitem{DBLP:journals/jmlr/CrammerS03a}
K.~Crammer and Y.~Singer, ``A family of additive online algorithms for category
  ranking,'' \emph{JMLR}, vol.~3, pp. 1025--1058, 2003.

\bibitem{DBLP:conf/sigir/KoNS07}
J.~Ko, E.~Nyberg, and L.~Si, ``A probabilistic graphical model for joint answer
  ranking in question answering,'' in \emph{SIGIR}, 2007, pp. 343--350.

\bibitem{DBLP:conf/iccv/BucakMJJ09}
S.~S. Bucak, P.~K. Mallapragada, R.~Jin, and A.~K. Jain, ``Efficient
  multi-label ranking for multi-class learning: Application to object
  recognition,'' in \emph{ICCV}, 2009, pp. 2098--2105.

\bibitem{DBLP:conf/icml/LiuXC05}
Y.~Liu, E.~P. Xing, and J.~G. Carbonell, ``Predicting protein folds with
  structural repeats using a chain graph model,'' in \emph{ICML}, 2005, pp.
  513--520.

\bibitem{koehn2005europarl}
P.~Koehn, ``Europarl: A parallel corpus for statistical machine translation,''
  in \emph{MT summit}, vol.~5, 2005, pp. 79--86.

\bibitem{DBLP:journals/coling/Shaalan14}
K.~Shaalan, ``A survey of arabic named entity recognition and classification,''
  \emph{Computational Linguistics}, vol.~40, no.~2, pp. 469--510, 2014.

\bibitem{DBLP:conf/nips/NewellD17}
A.~Newell and J.~Deng, ``Pixels to graphs by associative embedding,'' in
  \emph{NIPS}, 2017, pp. 2168--2177.

\bibitem{taskar2004max}
B.~Taskar, D.~Klein, M.~Collins, D.~Koller, and C.~Manning, ``Max-margin
  parsing,'' in \emph{EMNLP}, 2004.

\bibitem{DBLP:journals/jasis/Liben-NowellK07}
D.~Liben{-}Nowell and J.~M. Kleinberg, ``The link-prediction problem for social
  networks,'' \emph{JASIST}, vol.~58, no.~7, pp. 1019--1031, 2007.

\bibitem{park2003super}
S.~C. Park, M.~K. Park, and M.~G. Kang, ``Super-resolution image
  reconstruction: a technical overview,'' \emph{IEEE signal processing
  magazine}, vol.~20, no.~3, pp. 21--36, 2003.

\bibitem{DBLP:conf/ismir/YangCY17}
L.~Yang, S.~Chou, and Y.~Yang, ``Midinet: {A} convolutional generative
  adversarial network for symbolic-domain music generation,'' in \emph{ISMIR},
  2017, pp. 324--331.

\bibitem{DBLP:journals/pami/SmeuldersWSGJ00}
A.~W.~M. Smeulders, M.~Worring, S.~Santini, A.~Gupta, and R.~C. Jain,
  ``Content-based image retrieval at the end of the early years,''
  \emph{TPAMI}, vol.~22, no.~12, pp. 1349--1380, 2000.

\bibitem{DBLP:conf/trec/LauLT11}
C.~H. Lau, Y.~Li, and D.~Tjondronegoro, ``Microblog retrieval using topical
  features and query expansion,'' in \emph{TREC}, 2011.

\bibitem{DBLP:conf/sigir/MariaS00}
N.~Maria and M.~J. Silva, ``Theme-based retrieval of web news,'' in
  \emph{SIGIR}, 2000, pp. 354--356.

\bibitem{DBLP:journals/titb/ChoongCY09}
M.~K. Choong, M.~Charbit, and H.~Yan, ``Autoregressive-model-based missing
  value estimation for {DNA} microarray time series data,'' \emph{IEEE
  Transactions on Information Technology in Biomedicine}, vol.~13, no.~1, pp.
  131--137, 2009.

\bibitem{azadeh2008annual}
A.~Azadeh, S.~Ghaderi, and S.~Sohrabkhani, ``Annual electricity consumption
  forecasting by neural network in high energy consuming industrial sectors,''
  \emph{Energy Conversion and management}, vol.~49, no.~8, pp. 2272--2278,
  2008.

\bibitem{DBLP:journals/pami/WangMG09}
X.~Wang, X.~Ma, and W.~E.~L. Grimson, ``Unsupervised activity perception in
  crowded and complicated scenes using hierarchical bayesian models,''
  \emph{TPAMI}, vol.~31, no.~3, pp. 539--555, 2009.

\bibitem{DBLP:conf/ijcai/ShenSLYC07}
D.~Shen, J.~Sun, H.~Li, Q.~Yang, and Z.~Chen, ``Document summarization using
  conditional random fields,'' in \emph{IJCAI}, 2007, pp. 2862--2867.

\bibitem{DBLP:conf/iccv/LiangHZGX17}
X.~Liang, Z.~Hu, H.~Zhang, C.~Gan, and E.~P. Xing, ``Recurrent topic-transition
  {GAN} for visual paragraph generation,'' in \emph{ICCV}, 2017, pp.
  3382--3391.

\bibitem{DBLP:conf/icassp/GravesMH13}
A.~Graves, A.~Mohamed, and G.~E. Hinton, ``Speech recognition with deep
  recurrent neural networks,'' in \emph{ICASSP}, 2013, pp. 6645--6649.

\bibitem{DBLP:conf/icml/ReedAYLSL16}
S.~E. Reed, Z.~Akata, X.~Yan, L.~Logeswaran, B.~Schiele, and H.~Lee,
  ``Generative adversarial text to image synthesis,'' in \emph{ICML}, 2016, pp.
  1060--1069.

\bibitem{gauthier2014conditional}
J.~Gauthier, ``Conditional generative adversarial nets for convolutional face
  generation,'' \emph{Class Project for Stanford CS231N: Convolutional Neural
  Networks for Visual Recognition}, vol. 2014, no.~5, p.~2, 2014.

\bibitem{DBLP:conf/cvpr/JohnsonKSLSBL15}
J.~Johnson, R.~Krishna, M.~Stark, L.~Li, D.~A. Shamma, M.~S. Bernstein, and
  F.~Li, ``Image retrieval using scene graphs,'' in \emph{CVPR}, 2015, pp.
  3668--3678.

\bibitem{DBLP:conf/cvpr/JohnsonGF18}
J.~Johnson, A.~Gupta, and L.~Fei{-}Fei, ``Image generation from scene graphs,''
  in \emph{CVPR}, 2018, pp. 1219--1228.

\bibitem{DBLP:journals/ijcv/KrishnaZGJHKCKL17}
R.~Krishna, Y.~Zhu, O.~Groth, J.~Johnson, K.~Hata, J.~Kravitz, S.~Chen,
  Y.~Kalantidis, L.~Li, D.~A. Shamma, M.~S. Bernstein, and L.~Fei{-}Fei,
  ``Visual genome: Connecting language and vision using crowdsourced dense
  image annotations,'' \emph{IJCV}, vol. 123, no.~1, pp. 32--73, 2017.

\bibitem{DBLP:journals/corr/ZhangFHW16}
D.~Zhang, H.~Fu, J.~Han, and F.~Wu, ``A review of co-saliency detection
  technique: Fundamentals, applications, and challenges,'' \emph{CoRR}, vol.
  abs/1604.07090, 2016.

\bibitem{DBLP:conf/cvpr/JoulinBP10}
A.~Joulin, F.~R. Bach, and J.~Ponce, ``Discriminative clustering for image
  co-segmentation,'' in \emph{CVPR}, 2010, pp. 1943--1950.

\bibitem{bao2012object}
S.~Y. Bao, Y.~Xiang, and S.~Savarese, ``Object co-detection,'' in
  \emph{ECCV}.\hskip 1em plus 0.5em minus 0.4em\relax Springer, 2012, pp.
  86--101.

\bibitem{tsochantaridis2005large}
I.~Tsochantaridis, T.~Joachims, T.~Hofmann, and Y.~Altun, ``Large margin
  methods for structured and interdependent output variables,'' \emph{Journal
  of machine learning research}, vol.~6, no. Sep, pp. 1453--1484, 2005.

\bibitem{DBLP:journals/kbs/LiuCO18}
H.~Liu, J.~Cai, and Y.~Ong, ``Remarks on multi-output gaussian process
  regression,'' \emph{Knowledge-Based Systems}, vol. 144, pp. 102--121, 2018.

\bibitem{geng2016label}
X.~Geng, ``Label distribution learning,'' \emph{IEEE Transactions on Knowledge
  and Data Engineering}, vol.~28, no.~7, pp. 1734--1748, 2016.

\bibitem{DBLP:journals/pami/CarneiroCMV07}
G.~Carneiro, A.~B. Chan, P.~J. Moreno, and N.~Vasconcelos, ``Supervised
  learning of semantic classes for image annotation and retrieval,''
  \emph{TPAMI}, vol.~29, no.~3, pp. 394--410, 2007.

\bibitem{weigend2018time}
A.~S. Weigend, \emph{Time series prediction: forecasting the future and
  understanding the past}.\hskip 1em plus 0.5em minus 0.4em\relax Routledge,
  2018.

\bibitem{akata2015evaluation}
Z.~Akata, S.~Reed, D.~Walter, H.~Lee, and B.~Schiele, ``Evaluation of output
  embeddings for fine-grained image classification,'' in \emph{Proceedings of
  the IEEE Conference on Computer Vision and Pattern Recognition}, 2015, pp.
  2927--2936.

\bibitem{caruana1997multitask}
R.~Caruana, ``Multitask learning,'' \emph{Machine learning}, vol.~28, no.~1,
  pp. 41--75, 1997.

\bibitem{thrun1998clustering}
S.~Thrun and J.~O’Sullivan, ``Clustering learning tasks and the selective
  cross-task transfer of knowledge,'' in \emph{Learning to learn}.\hskip 1em
  plus 0.5em minus 0.4em\relax Springer, 1998, pp. 235--257.

\bibitem{mao2012objective}
Q.~Mao, I.~W.-H. Tsang, and S.~Gao, ``Objective-guided image annotation,''
  \emph{IEEE Transactions on Image Processing}, vol.~22, no.~4, pp. 1585--1597,
  2012.

\bibitem{DBLP:journals/pr/TahirKY12}
M.~A. Tahir, J.~Kittler, and F.~Yan, ``Inverse random under sampling for class
  imbalance problem and its application to multi-label classification,''
  \emph{Pattern Recognition}, vol.~45, no.~10, pp. 3738--3750, 2012.

\bibitem{DBLP:journals/npl/AlejoGP15}
\BIBentryALTinterwordspacing
R.~Alejo, V.~Garc{\'{\i}}a, and J.~H. Pacheco{-}S{\'{a}}nchez, ``An efficient
  over-sampling approach based on mean square error back-propagation for
  dealing with the multi-class imbalance problem,'' \emph{Neural Processing
  Letters}, vol.~42, no.~3, pp. 603--617, 2015. [Online]. Available:
  \url{https://doi.org/10.1007/s11063-014-9376-3}
\BIBentrySTDinterwordspacing

\bibitem{goodfellow2014generative}
I.~Goodfellow, J.~Pouget-Abadie, M.~Mirza, B.~Xu, D.~Warde-Farley, S.~Ozair,
  A.~Courville, and Y.~Bengio, ``Generative adversarial nets,'' in
  \emph{Advances in neural information processing systems}, 2014, pp.
  2672--2680.

\bibitem{tolstikhin2017adagan}
I.~O. Tolstikhin, S.~Gelly, O.~Bousquet, C.-J. Simon-Gabriel, and B.~Scholkopf,
  ``Adagan: Boosting generative models,'' in \emph{Advances in Neural
  Information Processing Systems}, 2017, pp. 5424--5433.

\bibitem{gretton2012kernel}
A.~Gretton, K.~M. Borgwardt, M.~J. Rasch, B.~Sch{\"o}lkopf, and A.~Smola, ``A
  kernel two-sample test,'' \emph{Journal of Machine Learning Research},
  vol.~13, no. Mar, pp. 723--773, 2012.

\bibitem{khrulkov2018geometry}
V.~Khrulkov and I.~Oseledets, ``Geometry score: A method for comparing
  generative adversarial networks,'' \emph{arXiv preprint arXiv:1802.02664},
  2018.

\bibitem{salimans2016improved}
T.~Salimans, I.~Goodfellow, W.~Zaremba, V.~Cheung, A.~Radford, and X.~Chen,
  ``Improved techniques for training gans,'' in \emph{Advances in neural
  information processing systems}, 2016, pp. 2234--2242.

\bibitem{che2016mode}
T.~Che, Y.~Li, A.~P. Jacob, Y.~Bengio, and W.~Li, ``Mode regularized generative
  adversarial networks,'' \emph{arXiv preprint arXiv:1612.02136}, 2016.

\bibitem{heusel2017gans}
M.~Heusel, H.~Ramsauer, T.~Unterthiner, B.~Nessler, and S.~Hochreiter, ``Gans
  trained by a two time-scale update rule converge to a local nash
  equilibrium,'' in \emph{Advances in Neural Information Processing Systems},
  2017, pp. 6626--6637.

\bibitem{binkowski2018demystifying}
M.~Bi{\'n}kowski, D.~J. Sutherland, M.~Arbel, and A.~Gretton, ``Demystifying
  mmd gans,'' \emph{arXiv preprint arXiv:1801.01401}, 2018.

\bibitem{DBLP:conf/nips/LucicKMGB18}
M.~Lucic, K.~Kurach, M.~Michalski, S.~Gelly, and O.~Bousquet, ``Are gans
  created equal? {A} large-scale study,'' in \emph{NeualPS}, 2018, pp.
  698--707.

\bibitem{DBLP:conf/recsys/McAuleyL13}
J.~J. McAuley and J.~Leskovec, ``Hidden factors and hidden topics:
  understanding rating dimensions with review text,'' in \emph{Seventh {ACM}
  Conference on Recommender Systems}, 2013, pp. 165--172.

\bibitem{DBLP:conf/sigir/McAuleyTSH15}
J.~J. McAuley, C.~Targett, Q.~Shi, and A.~van~den Hengel, ``Image-based
  recommendations on styles and substitutes,'' in \emph{SIGIR}, 2015, pp.
  43--52.

\bibitem{DBLP:conf/kdd/McAuleyPL15}
J.~J. McAuley, R.~Pandey, and J.~Leskovec, ``Inferring networks of
  substitutable and complementary products,'' in \emph{KDD}, 2015, pp.
  785--794.

\bibitem{DBLP:conf/nips/BhatiaJKVJ15}
K.~Bhatia, H.~Jain, P.~Kar, M.~Varma, and P.~Jain, ``Sparse local embeddings
  for extreme multi-label classification,'' in \emph{NIPS}, 2015, pp. 730--738.

\bibitem{DBLP:journals/corr/abs-1202-5469}
A.~Zubiaga, ``Enhancing navigation on wikipedia with social tags,''
  \emph{CoRR}, vol. abs/1202.5469, 2012.

\bibitem{wetzker2008analyzing}
R.~Wetzker, C.~Zimmermann, and C.~Bauckhage, ``Analyzing social bookmarking
  systems: A del.icio.us cookbook,'' in \emph{Proceedings of the ECAI 2008
  Mining Social Data Workshop}, 2008, pp. 26--30.

\bibitem{DBLP:conf/kdd/PrabhuV14}
Y.~Prabhu and M.~Varma, ``Fastxml: a fast, accurate and stable tree-classifier
  for extreme multi-label learning,'' in \emph{KDD}, 2014, pp. 263--272.

\bibitem{DBLP:journals/corr/PartalasKBAPGAA15}
I.~Partalas, A.~Kosmopoulos, N.~Baskiotis, T.~Arti{\`{e}}res, G.~Paliouras,
  {\'{E}}.~Gaussier, I.~Androutsopoulos, M.~Amini, and P.~Gallinari, ``{LSHTC:}
  {A} benchmark for large-scale text classification,'' \emph{CoRR}, vol.
  abs/1503.08581, 2015.

\bibitem{DBLP:conf/eccv/LiHLT16}
Y.~Li, C.~Huang, C.~C. Loy, and X.~Tang, ``Human attribute recognition by deep
  hierarchical contexts,'' in \emph{ECCV}, 2016, pp. 684--700.

\bibitem{DBLP:conf/iccv/LiuLWT15}
Z.~Liu, P.~Luo, X.~Wang, and X.~Tang, ``Deep learning face attributes in the
  wild,'' in \emph{ICCV}, 2015, pp. 3730--3738.

\bibitem{DBLP:conf/cvpr/LiuLQWT16}
Z.~Liu, P.~Luo, S.~Qiu, X.~Wang, and X.~Tang, ``Deepfashion: Powering robust
  clothes recognition and retrieval with rich annotations,'' in \emph{CVPR},
  2016, pp. 1096--1104.

\bibitem{DBLP:conf/cvpr/ChenHFBDY15}
Q.~Chen, J.~Huang, R.~S. Feris, L.~M. Brown, J.~Dong, and S.~Yan, ``Deep domain
  adaptation for describing people based on fine-grained clothing attributes,''
  in \emph{CVPR}, 2015, pp. 5315--5324.

\bibitem{DBLP:conf/cvpr/FarhadiEHF09}
A.~Farhadi, I.~Endres, D.~Hoiem, and D.~A. Forsyth, ``Describing objects by
  their attributes,'' in \emph{CVPR}, 2009, pp. 1778--1785.

\bibitem{DBLP:conf/cvpr/ChoiLTW10}
M.~J. Choi, J.~J. Lim, A.~Torralba, and A.~S. Willsky, ``Exploiting
  hierarchical context on a large database of object categories,'' in
  \emph{CVPR}, 2010, pp. 129--136.

\bibitem{DBLP:conf/ismir/MarquesDLG11}
G.~Marques, M.~A. Domingues, T.~Langlois, and F.~Gouyon, ``Three current issues
  in music autotagging,'' in \emph{International Society for Music Information
  Retrieval Conference}, 2011, pp. 795--800.

\bibitem{WahCUB_200_2011}
C.~Wah, S.~Branson, P.~Welinder, P.~Perona, and S.~Belongie, ``{The
  Caltech-UCSD Birds-200-2011 Dataset},'' Tech. Rep., 2011.

\bibitem{jouhet2012automated}
V.~Jouhet, G.~Defossez, A.~Burgun, P.~Le~Beux, P.~Levillain, P.~Ingrand,
  V.~Claveau \emph{et~al.}, ``Automated classification of free-text pathology
  reports for registration of incident cases of cancer,'' \emph{Methods of
  information in medicine}, vol.~51, no.~3, p. 242, 2012.

\bibitem{johnson2016mimic}
A.~E. Johnson, T.~J. Pollard, L.~Shen, H.~L. Li-wei, M.~Feng, M.~Ghassemi,
  B.~Moody, P.~Szolovits, L.~A. Celi, and R.~G. Mark, ``Mimic-iii, a freely
  accessible critical care database,'' \emph{Scientific data}, vol.~3, p.
  160035, 2016.

\bibitem{DBLP:journals/jmlr/LewisYRL04}
D.~D. Lewis, Y.~Yang, T.~G. Rose, and F.~Li, ``{RCV1:} {A} new benchmark
  collection for text categorization research,'' \emph{JMLR}, vol.~5, pp.
  361--397, 2004.

\bibitem{kaggletaxiroute2015}
``Kaggle data set ecml/pkdd 15: Taxi trajectory prediction (1).''

\bibitem{CRAWDADdata2009}
\BIBentryALTinterwordspacing
M.~Piorkowski, N.~Sarafijanovic-Djukic, and M.~Grossglauser, ``Crawdad data set
  epfl/mobility (v. 2009-02-24),'' Feb. 2009. [Online]. Available:
  \url{Downloaded from http://crawdad.org/epfl/mobility/}
\BIBentrySTDinterwordspacing

\bibitem{ElectricityLoadDiagrams20112014}
\BIBentryALTinterwordspacing
A.~Trindade, ``Uci maching learning repository -
  electricityloaddiagrams20112014 data set,'' 2016. [Online]. Available:
  \url{http://archive.ics.uci.edu/ml/datasets/ElectricityLoadDiagrams20112014}
\BIBentrySTDinterwordspacing

\bibitem{DBLP:conf/cvpr/OhHPCCLMALDSWJRSVPRYTSFRD11}
S.~Oh, A.~Hoogs, A.~G.~A. Perera, N.~P. Cuntoor, C.~Chen, J.~T. Lee,
  S.~Mukherjee, J.~K. Aggarwal, H.~Lee, L.~S. Davis, E.~Swears, X.~Wang, Q.~Ji,
  K.~K. Reddy, M.~Shah, C.~Vondrick, H.~Pirsiavash, D.~Ramanan, J.~Yuen,
  A.~Torralba, B.~Song, A.~Fong, A.~K. Roy{-}Chowdhury, and M.~Desai, ``A
  large-scale benchmark dataset for event recognition in surveillance video,''
  in \emph{CVPR}, 2011, pp. 3153--3160.

\bibitem{DBLP:journals/ijcv/EveringhamGWWZ10}
M.~Everingham, L.~J.~V. Gool, C.~K.~I. Williams, J.~M. Winn, and A.~Zisserman,
  ``The pascal visual object classes {(VOC)} challenge,'' \emph{IJCV}, vol.~88,
  no.~2, pp. 303--338, 2010.

\bibitem{krizhevsky2009learning}
A.~Krizhevsky and G.~Hinton, ``Learning multiple layers of features from tiny
  images,'' Citeseer, Tech. Rep., 2009.

\bibitem{DBLP:journals/cacm/Miller95}
G.~A. Miller, ``Wordnet: {A} lexical database for english,''
  \emph{Communications of the ACM}, vol.~38, no.~11, pp. 39--41, 1995.

\bibitem{mahoney2011large}
\BIBentryALTinterwordspacing
M.~Mahoney, ``Large text compression benchmark,'' 2011. [Online]. Available:
  \url{http://www.mattmahoney.net/text/text.html}
\BIBentrySTDinterwordspacing

\bibitem{Zafarani+Liu:2009}
\BIBentryALTinterwordspacing
R.~Zafarani and H.~Liu, ``Social computing data repository at {ASU},'' 2009.
  [Online]. Available: \url{http://socialcomputing.asu.edu}
\BIBentrySTDinterwordspacing

\bibitem{snapnets}
J.~Leskovec and A.~Krevl, ``{SNAP Datasets}: {Stanford} large network dataset
  collection,'' \url{http://snap.stanford.edu/data}, Jun. 2014.

\bibitem{DBLP:conf/conll/SangB00}
E.~F. T.~K. Sang and S.~Buchholz, ``Introduction to the conll-2000 shared task
  chunking,'' in \emph{Proceedings of the 2nd workshop on Learning language in
  logic and the 4th conference on Computational natural language learning},
  2000, pp. 127--132.

\bibitem{DBLP:conf/nips/ZhouPBM12}
D.~Zhou, J.~C. Platt, S.~Basu, and Y.~Mao, ``Learning from the wisdom of crowds
  by minimax entropy,'' in \emph{Conference on Neural Information Processing
  Systems}, 2012, pp. 2204--2212.

\bibitem{DBLP:conf/cvpr/LeeHZY18}
K.~Lee, X.~He, L.~Zhang, and L.~Yang, ``Cleannet: Transfer learning for
  scalable image classifier training with label noise,'' in \emph{CVPR}, 2018,
  pp. 5447--5456.

\bibitem{DBLP:conf/cvpr/XiaoXYHW15}
T.~Xiao, T.~Xia, Y.~Yang, C.~Huang, and X.~Wang, ``Learning from massive noisy
  labeled data for image classification,'' in \emph{CVPR}, 2015, pp.
  2691--2699.

\bibitem{DBLP:journals/corr/abs-1708-02862}
W.~Li, L.~Wang, W.~Li, E.~Agustsson, and L.~V. Gool, ``Webvision database:
  Visual learning and understanding from web data,'' \emph{CoRR}, vol.
  abs/1708.02862, 2017.

\bibitem{DBLP:journals/corr/ThomeeSFENPBL15}
B.~Thomee, D.~A. Shamma, G.~Friedland, B.~Elizalde, K.~Ni, D.~Poland, D.~Borth,
  and L.~Li, ``The new data and new challenges in multimedia research,''
  \emph{CoRR}, vol. abs/1503.01817, 2015.

\bibitem{DBLP:journals/jmlr/KummerleS18}
C.~K{\"{u}}mmerle and J.~Sigl, ``Harmonic mean iteratively reweighted least
  squares for low-rank matrix recovery,'' \emph{JMLR}, vol.~19, 2018.

\bibitem{DBLP:journals/jmlr/LiuT17}
W.~Liu and I.~W. Tsang, ``Making decision trees feasible in ultrahigh feature
  and label dimensions,'' \emph{JMLR}, vol.~18, pp. 81:1--81:36, 2017.

\bibitem{DBLP:journals/jmlr/DupuyB17}
C.~Dupuy and F.~Bach, ``Online but accurate inference for latent variable
  models with local gibbs sampling,'' \emph{JMLR}, vol.~18, pp. 126:1--126:45,
  2017.

\bibitem{DBLP:journals/jmlr/LiuTM17}
W.~Liu, I.~W. Tsang, and K.~M{\"{u}}ller, ``An easy-to-hard learning paradigm
  for multiple classes and multiple labels,'' \emph{JMLR}, vol.~18, pp.
  94:1--94:38, 2017.

\bibitem{DBLP:journals/jmlr/BrouardSd16}
C.~Brouard, M.~Szafranski, and F.~d'Alch{\'{e}}{-}Buc, ``Input output kernel
  regression: Supervised and semi-supervised structured output prediction with
  operator-valued kernels,'' \emph{JMLR}, vol.~17, pp. 176:1--176:48, 2016.

\bibitem{DBLP:journals/jmlr/ShinLKSYS16}
H.~Shin, L.~Lu, L.~Kim, A.~Seff, J.~Yao, and R.~M. Summers, ``Interleaved
  text/image deep mining on a large-scale radiology database for automated
  image interpretation,'' \emph{JMLR}, vol.~17, pp. 107:1--107:31, 2016.

\bibitem{DBLP:journals/jmlr/BabbarPGAA16}
R.~Babbar, I.~Partalas, {\'{E}}.~Gaussier, M.~Amini, and C.~Amblard, ``Learning
  taxonomy adaptation in large-scale classification,'' \emph{JMLR}, vol.~17,
  pp. 98:1--98:37, 2016.

\bibitem{DBLP:journals/jmlr/LiZYL15}
X.~Li, T.~Zhao, X.~Yuan, and H.~Liu, ``The flare package for high dimensional
  linear regression and precision matrix estimation in {R},'' \emph{JMLR},
  vol.~16, pp. 553--557, 2015.

\bibitem{DBLP:journals/jmlr/HanLL15}
F.~Han, H.~Lu, and H.~Liu, ``A direct estimation of high dimensional stationary
  vector autoregressions,'' \emph{JMLR}, vol.~16, pp. 3115--3150, 2015.

\bibitem{DBLP:journals/jmlr/DoppaFT14}
J.~R. Doppa, A.~Fern, and P.~Tadepalli, ``Structured prediction via output
  space search,'' \emph{JMLR}, vol.~15, no.~1, pp. 1317--1350, 2014.

\bibitem{DBLP:journals/jmlr/ColomboM14}
D.~Colombo and M.~H. Maathuis, ``Order-independent constraint-based causal
  structure learning,'' \emph{JMLR}, vol.~15, no.~1, pp. 3741--3782, 2014.

\bibitem{DBLP:journals/jmlr/GentileO14}
C.~Gentile and F.~Orabona, ``On multilabel classification and ranking with
  bandit feedback,'' \emph{JMLR}, vol.~15, no.~1, pp. 2451--2487, 2014.

\bibitem{DBLP:journals/jmlr/GongYZ13}
P.~Gong, J.~Ye, and C.~Zhang, ``Multi-stage multi-task feature learning,''
  \emph{JMLR}, vol.~14, no.~1, pp. 2979--3010, 2013.

\bibitem{DBLP:journals/jmlr/TalwalkarKMR13}
A.~Talwalkar, S.~Kumar, M.~Mohri, and H.~A. Rowley, ``Large-scale {SVD} and
  manifold learning,'' \emph{JMLR}, vol.~14, no.~1, pp. 3129--3152, 2013.

\bibitem{DBLP:journals/tnn/FuLJZ18}
K.~Fu, J.~Li, J.~Jin, and C.~Zhang, ``Image-text surgery: Efficient concept
  learning in image captioning by generating pseudopairs,'' \emph{TNNLS},
  vol.~29, no.~12, pp. 5910--5921, 2018.

\bibitem{8573154}
E.~Protas, J.~D. Bratti, J.~F.~O. Gaya, P.~Drews, and S.~S.~C. Botelho,
  ``Visualization methods for image transformation convolutional neural
  networks,'' \emph{TNNLS}, 2018.

\bibitem{DBLP:journals/tnn/ZhangWXCW18}
H.~Zhang, S.~Wang, X.~Xu, T.~W.~S. Chow, and Q.~M.~J. Wu, ``Tree2vector:
  Learning a vectorial representation for tree-structured data,'' \emph{TNNLS},
  vol.~29, no.~11, pp. 5304--5318, 2018.

\bibitem{DBLP:journals/tnn/LinDHS18}
Z.~Lin, G.~Ding, J.~Han, and L.~Shao, ``End-to-end feature-aware label space
  encoding for multilabel classification with many classes,'' \emph{TNNLS},
  vol.~29, no.~6, pp. 2472--2487, 2018.

\bibitem{DBLP:journals/tnn/FangLTHC17}
S.~Fang, J.~Li, Y.~Tian, T.~Huang, and X.~Chen, ``Learning discriminative
  subspaces on random contrasts for image saliency analysis,'' \emph{TNNLS},
  vol.~28, no.~5, pp. 1095--1108, 2017.

\bibitem{DBLP:journals/tnn/ZhangTGLL17}
K.~Zhang, D.~Tao, X.~Gao, X.~Li, and J.~Li, ``Coarse-to-fine learning for
  single-image super-resolution,'' \emph{TNNLS}, vol.~28, no.~5, pp.
  1109--1122, 2017.

\bibitem{DBLP:journals/tnn/Kim17}
M.~Kim, ``Mixtures of conditional random fields for improved structured output
  prediction,'' \emph{TNNLS}, vol.~28, no.~5, pp. 1233--1240, 2017.

\bibitem{DBLP:journals/tnn/CheungLPC17}
Y.~Cheung, M.~Li, Q.~Peng, and C.~L.~P. Chen, ``A cooperative learning-based
  clustering approach to lip segmentation without knowing segment number,''
  \emph{TNNLS}, vol.~28, no.~1, pp. 80--93, 2017.

\bibitem{DBLP:journals/tnn/WangLZ17}
L.~Wang, L.~Liu, and L.~Zhou, ``A graph-embedding approach to hierarchical
  visual word mergence,'' \emph{TNNLS}, vol.~28, no.~2, pp. 308--320, 2017.

\bibitem{DBLP:journals/tnn/LiLXYZ17}
Z.~Li, Z.~Lai, Y.~Xu, J.~Yang, and D.~Zhang, ``A locality-constrained and label
  embedding dictionary learning algorithm for image classification,''
  \emph{TNNLS}, vol.~28, no.~2, pp. 278--293, 2017.

\bibitem{DBLP:journals/tnn/LuoZLYXZ16}
X.~Luo, M.~Zhou, S.~Li, Z.~You, Y.~Xia, and Q.~Zhu, ``A nonnegative latent
  factor model for large-scale sparse matrices in recommender systems via
  alternating direction method,'' \emph{TNNLS}, vol.~27, no.~3, pp. 579--592,
  2016.

\bibitem{DBLP:journals/tnn/DengXZTGL16}
C.~Deng, J.~Xu, K.~Zhang, D.~Tao, X.~Gao, and X.~Li, ``Similarity
  constraints-based structured output regression machine: An approach to image
  super-resolution,'' \emph{TNNLS}, vol.~27, no.~12, pp. 2472--2485, 2016.

\bibitem{DBLP:journals/tnn/AlushG16}
A.~Alush and J.~Goldberger, ``Hierarchical image segmentation using correlation
  clustering,'' \emph{TNNLS}, vol.~27, no.~6, pp. 1358--1367, 2016.

\bibitem{DBLP:journals/tnn/TaoCSL16}
D.~Tao, J.~Cheng, M.~Song, and X.~Lin, ``Manifold ranking-based matrix
  factorization for saliency detection,'' \emph{TNNLS}, vol.~27, no.~6, pp.
  1122--1134, 2016.

\bibitem{DBLP:journals/tnn/CaoCTZ16}
F.~Cao, M.~Cai, Y.~Tan, and J.~Zhao, ``Image super-resolution via adaptive
  {\(\mathscr{l}\)}\({}_{\mbox{p}}\) (0{\textless}p{\textless}1) regularization
  and sparse representation,'' \emph{TNNLS}, vol.~27, no.~7, pp. 1550--1561,
  2016.

\bibitem{DBLP:journals/tnn/ZhuSLW15}
Q.~Zhu, L.~Shao, X.~Li, and L.~Wang, ``Targeting accurate object extraction
  from an image: {A} comprehensive study of natural image matting,''
  \emph{TNNLS}, vol.~26, no.~2, pp. 185--207, 2015.

\bibitem{DBLP:journals/tnn/ChenMKP15}
Y.~Chen, Y.~Ma, D.~H. Kim, and S.~Park, ``Region-based object recognition by
  color segmentation using a simplified {PCNN},'' \emph{TNNLS}, vol.~26, no.~8,
  pp. 1682--1697, 2015.

\bibitem{DBLP:journals/tnn/LiBKL15}
M.~Li, W.~Bi, J.~T. Kwok, and B.~Lu, ``Large-scale nystr{\"{o}}m kernel matrix
  approximation using randomized {SVD},'' \emph{TNNLS}, vol.~26, no.~1, pp.
  152--164, 2015.

\bibitem{DBLP:journals/tnn/YuGTLZ14}
J.~Yu, X.~Gao, D.~Tao, X.~Li, and K.~Zhang, ``A unified learning framework for
  single image super-resolution,'' \emph{TNNLS}, vol.~25, no.~4, pp. 780--792,
  2014.

\bibitem{DBLP:journals/tnn/BauerGBMK14}
A.~Bauer, N.~G{\"{o}}rnitz, F.~Biegler, K.~M{\"{u}}ller, and M.~Kloft,
  ``Efficient algorithms for exact inference in sequence labeling svms,''
  \emph{TNNLS}, vol.~25, no.~5, pp. 870--881, 2014.

\bibitem{DBLP:journals/tnn/TangLSZ14}
K.~Tang, R.~Liu, Z.~Su, and J.~Zhang, ``Structure-constrained low-rank
  representation,'' \emph{TNNLS}, vol.~25, no.~12, pp. 2167--2179, 2014.

\bibitem{DBLP:journals/tnn/DengDLZH13}
Y.~Deng, Q.~Dai, R.~Liu, Z.~Zhang, and S.~Hu, ``Low-rank structure learning via
  nonconvex heuristic recovery,'' \emph{TNNLS}, vol.~24, no.~3, pp. 383--396,
  2013.

\bibitem{DBLP:journals/tnn/ZhangWN13}
H.~Zhang, Q.~M.~J. Wu, and T.~M. Nguyen, ``Incorporating mean template into
  finite mixture model for image segmentation,'' \emph{TNNLS}, vol.~24, no.~2,
  pp. 328--335, 2013.

\bibitem{DBLP:journals/tnn/PangJJL13}
Y.~Pang, Z.~Ji, P.~Jing, and X.~Li, ``Ranking graph embedding for learning to
  rerank,'' \emph{TNNLS}, vol.~24, no.~8, pp. 1292--1303, 2013.

\bibitem{DBLP:journals/tnn/LuoTXXLW13}
Y.~Luo, D.~Tao, C.~Xu, C.~Xu, H.~Liu, and Y.~Wen, ``Multiview vector-valued
  manifold regularization for multilabel image classification,'' \emph{TNNLS},
  vol.~24, no.~5, pp. 709--722, 2013.

\bibitem{zhang2018neural}
B.~Zhang, D.~Xiong, and J.~Su, ``Neural machine translation with deep
  attention,'' \emph{TPAMI}, 2018.

\bibitem{DBLP:journals/pami/JeongLKKN18}
S.~Jeong, J.~Lee, B.~Kim, Y.~Kim, and J.~Noh, ``Object segmentation ensuring
  consistency across multi-viewpoint images,'' \emph{TPAMI}, vol.~40, no.~10,
  pp. 2455--2468, 2018.

\bibitem{DBLP:journals/pami/RaposoAB18}
C.~Raposo, M.~Antunes, and J.~P. Barreto, ``Piecewise-planar stereoscan:
  Sequential structure and motion using plane primitives,'' \emph{TPAMI},
  vol.~40, no.~8, pp. 1918--1931, 2018.

\bibitem{DBLP:journals/pami/CordtsREFR17}
M.~Cordts, T.~Rehfeld, M.~Enzweiler, U.~Franke, and S.~Roth, ``Tree-structured
  models for efficient multi-cue scene labeling,'' \emph{TPAMI}, vol.~39,
  no.~7, pp. 1444--1454, 2017.

\bibitem{DBLP:journals/pami/XuCGN17}
Y.~Xu, E.~Carlinet, T.~G{\'{e}}raud, and L.~Najman, ``Hierarchical segmentation
  using tree-based shape spaces,'' \emph{TPAMI}, vol.~39, no.~3, pp. 457--469,
  2017.

\bibitem{DBLP:journals/pami/FuJCSZ17}
K.~Fu, J.~Jin, R.~Cui, F.~Sha, and C.~Zhang, ``Aligning where to see and what
  to tell: Image captioning with region-based attention and scene-specific
  contexts,'' \emph{TPAMI}, vol.~39, no.~12, pp. 2321--2334, 2017.

\bibitem{DBLP:journals/pami/HasnatAT16}
M.~A. Hasnat, O.~Alata, and A.~Tr{\'{e}}meau, ``Joint color-spatial-directional
  clustering and region merging {(JCSD-RM)} for unsupervised {RGB-D} image
  segmentation,'' \emph{TPAMI}, vol.~38, no.~11, pp. 2255--2268, 2016.

\bibitem{DBLP:journals/pami/GirshickDDM16}
R.~B. Girshick, J.~Donahue, T.~Darrell, and J.~Malik, ``Region-based
  convolutional networks for accurate object detection and segmentation,''
  \emph{TPAMI}, vol.~38, no.~1, pp. 142--158, 2016.

\bibitem{DBLP:journals/pami/QinS16}
Z.~Qin and C.~R. Shelton, ``Social grouping for multi-target tracking and head
  pose estimation in video,'' \emph{TPAMI}, vol.~38, no.~10, pp. 2082--2095,
  2016.

\bibitem{DBLP:journals/pami/KwonKTKT15}
Y.~Kwon, K.~I. Kim, J.~Tompkin, J.~H. Kim, and C.~Theobalt, ``Efficient
  learning of image super-resolution and compression artifact removal with
  semi-local gaussian processes,'' \emph{TPAMI}, vol.~37, no.~9, pp.
  1792--1805, 2015.

\bibitem{DBLP:journals/pami/DjelouahFBCP15}
A.~Djelouah, J.~Franco, E.~Boyer, F.~L. Clerc, and P.~P{\'{e}}rez, ``Sparse
  multi-view consistency for object segmentation,'' \emph{TPAMI}, vol.~37,
  no.~9, pp. 1890--1903, 2015.

\bibitem{DBLP:journals/access/WangWLZZ18}
S.~Wang, Y.~Wei, K.~Long, X.~Zeng, and M.~Zheng, ``Image super-resolution via
  self-similarity learning and conformal sparse representation,'' \emph{TPAMI}.

\bibitem{DBLP:journals/pami/ShitritBFF14}
H.~B. Shitrit, J.~Berclaz, F.~Fleuret, and P.~Fua, ``Multi-commodity network
  flow for tracking multiple people,'' \emph{TPAMI}, vol.~36, no.~8, pp.
  1614--1627, 2014.

\bibitem{DBLP:journals/pami/ZhouF14}
N.~Zhou and J.~Fan, ``Jointly learning visually correlated dictionaries for
  large-scale visual recognition applications,'' \emph{TPAMI}, vol.~36, no.~4,
  pp. 715--730, 2014.

\bibitem{DBLP:journals/pami/PerakisPTK13}
P.~Perakis, G.~Passalis, T.~Theoharis, and I.~A. Kakadiaris, ``3d facial
  landmark detection under large yaw and expression variations,'' \emph{TPAMI},
  vol.~35, no.~7, pp. 1552--1564, 2013.

\bibitem{DBLP:journals/pami/GongLGP13}
Y.~Gong, S.~Lazebnik, A.~Gordo, and F.~Perronnin, ``Iterative quantization: {A}
  procrustean approach to learning binary codes for large-scale image
  retrieval,'' \emph{TPAMI}, vol.~35, no.~12, pp. 2916--2929, 2013.

\bibitem{DBLP:conf/kdd/DizajiWH18}
K.~G. Dizaji, X.~Wang, and H.~Huang, ``Semi-supervised generative adversarial
  network for gene expression inference,'' in \emph{KDD}, 2018, pp. 1435--1444.

\bibitem{DBLP:conf/kdd/LeeGZ18}
M.~Lee, B.~Gao, and R.~Zhang, ``Rare query expansion through generative
  adversarial networks in search advertising,'' in \emph{KDD}, 2018, pp.
  500--508.

\bibitem{DBLP:conf/kdd/YenHDRDX17}
I.~E. Yen, X.~Huang, W.~Dai, P.~Ravikumar, I.~S. Dhillon, and E.~P. Xing,
  ``Ppdsparse: {A} parallel primal-dual sparse method for extreme
  classification,'' in \emph{KDD}, 2017, pp. 545--553.

\bibitem{DBLP:conf/kdd/Tagami17}
Y.~Tagami, ``Annexml: Approximate nearest neighbor search for extreme
  multi-label classification,'' in \emph{KDD}, 2017, pp. 455--464.

\bibitem{DBLP:conf/kdd/JainPV16}
H.~Jain, Y.~Prabhu, and M.~Varma, ``Extreme multi-label loss functions for
  recommendation, tagging, ranking {\&} other missing label applications,'' in
  \emph{KDD}, 2016, pp. 935--944.

\bibitem{DBLP:conf/kdd/XuT016}
C.~Xu, D.~Tao, and C.~Xu, ``Robust extreme multi-label learning,'' in
  \emph{KDD}, 2016, pp. 1275--1284.

\bibitem{DBLP:conf/kdd/KuoWWCYD15}
C.~Kuo, X.~Wang, P.~B. Walker, O.~T. Carmichael, J.~Ye, and I.~Davidson,
  ``Unified and contrasting cuts in multiple graphs: Application to medical
  imaging segmentation,'' in \emph{KDD}, 2015, pp. 617--626.

\bibitem{DBLP:conf/kdd/Papagiannopoulou15}
C.~Papagiannopoulou, G.~Tsoumakas, and I.~Tsamardinos, ``Discovering and
  exploiting deterministic label relationships in multi-label learning,'' in
  \emph{KDD}, 2015, pp. 915--924.

\bibitem{DBLP:conf/kdd/WuZTH014}
B.~Wu, E.~Zhong, B.~Tan, A.~Horner, and Q.~Yang, ``Crowdsourced time-sync video
  tagging using temporal and personalized topic modeling,'' in \emph{KDD},
  2014, pp. 721--730.

\bibitem{DBLP:conf/kdd/ZhaiXW14}
S.~Zhai, T.~Xia, and S.~Wang, ``A multi-class boosting method with direct
  optimization,'' in \emph{KDD}, 2014, pp. 273--282.

\bibitem{DBLP:conf/kdd/KongCY13}
X.~Kong, B.~Cao, and P.~S. Yu, ``Multi-label classification by mining label and
  instance correlations from heterogeneous information networks,'' in
  \emph{KDD}, 2013, pp. 614--622.

\bibitem{DBLP:conf/kdd/WangS13}
X.~Wang and G.~Sukthankar, ``Multi-label relational neighbor classification
  using social context features,'' in \emph{KDD}, 2013, pp. 464--472.

\bibitem{DBLP:conf/nips/HongYHL18}
S.~Hong, X.~Yan, T.~S. Huang, and H.~Lee, ``Learning hierarchical semantic
  image manipulation through structured representations,'' in \emph{NIPS},
  2018, pp. 2713--2723.

\bibitem{DBLP:conf/nips/WydmuchJKBD18}
M.~Wydmuch, K.~Jasinska, M.~Kuznetsov, R.~Busa{-}Fekete, and K.~Dembczynski,
  ``A no-regret generalization of hierarchical softmax to extreme multi-label
  classification,'' in \emph{NIPS}, 2018, pp. 6358--6368.

\bibitem{DBLP:conf/nips/PanYLZZCH18}
B.~Pan, Y.~Yang, H.~Li, Z.~Zhao, Y.~Zhuang, D.~Cai, and X.~He, ``Macnet:
  Transferring knowledge from machine comprehension to sequence-to-sequence
  models,'' in \emph{NIPS}, 2018, pp. 6095--6105.

\bibitem{DBLP:conf/nips/RacahBMKPP17}
E.~Racah, C.~Beckham, T.~Maharaj, S.~E. Kahou, Prabhat, and C.~Pal,
  ``Extremeweather: {A} large-scale climate dataset for semi-supervised
  detection, localization, and understanding of extreme weather events,'' in
  \emph{NIPS}, 2017, pp. 3405--3416.

\bibitem{DBLP:conf/nips/HuHS17}
Y.~Hu, J.~Huang, and A.~G. Schwing, ``Maskrnn: Instance level video object
  segmentation,'' in \emph{NIPS}, 2017, pp. 324--333.

\bibitem{DBLP:conf/nips/JoshiAPIM17}
B.~Joshi, M.~Amini, I.~Partalas, F.~Iutzeler, and Y.~Maximov, ``Aggressive
  sampling for multi-class to binary reduction with applications to text
  classification,'' in \emph{NIPS}, 2017, pp. 4162--4171.

\bibitem{DBLP:conf/nips/NamMKF17}
J.~Nam, E.~Loza~Menc\'{i}a, H.~J. Kim, and J.~F{\"{u}}rnkranz, ``Maximizing
  subset accuracy with recurrent neural networks in multi-label
  classification,'' in \emph{NIPS}, 2017, pp. 5419--5429.

\bibitem{DBLP:conf/nips/RosenfeldG16}
N.~Rosenfeld and A.~Globerson, ``Optimal tagging with markov chain
  optimization,'' in \emph{NIPS}, 2016, pp. 1307--1315.

\bibitem{DBLP:conf/nips/YuRD16}
H.~Yu, N.~Rao, and I.~S. Dhillon, ``Temporal regularized matrix factorization
  for high-dimensional time series prediction,'' in \emph{NIPS}, 2016, pp.
  847--855.

\bibitem{DBLP:conf/nips/BengioVJS15}
S.~Bengio, O.~Vinyals, N.~Jaitly, and N.~Shazeer, ``Scheduled sampling for
  sequence prediction with recurrent neural networks,'' in \emph{NIPS}, 2015,
  pp. 1171--1179.

\bibitem{DBLP:conf/nips/RaiHHC15}
P.~Rai, C.~Hu, R.~Henao, and L.~Carin, ``Large-scale bayesian multi-label
  learning via topic-based label embeddings,'' in \emph{NIPS}, 2015, pp.
  3222--3230.

\bibitem{DBLP:conf/nips/WuPKP14}
A.~Wu, M.~Park, O.~Koyejo, and J.~W. Pillow, ``Sparse bayesian structure
  learning with dependent relevance determination priors,'' in \emph{NIPS},
  2014, pp. 1628--1636.

\bibitem{DBLP:conf/nips/NguyenBRC14}
V.~Nguyen, J.~L. Boyd{-}Graber, P.~Resnik, and J.~Chang, ``Learning a concept
  hierarchy from multi-labeled documents,'' in \emph{NIPS}, 2014, pp.
  3671--3679.

\bibitem{DBLP:conf/nips/HoffmanGTHDGDS14}
J.~Hoffman, S.~Guadarrama, E.~Tzeng, R.~Hu, J.~Donahue, R.~B. Girshick,
  T.~Darrell, and K.~Saenko, ``{LSDA:} large scale detection through
  adaptation,'' in \emph{NIPS}, 2014, pp. 3536--3544.

\bibitem{DBLP:conf/nips/XuJZ13}
M.~Xu, R.~Jin, and Z.~Zhou, ``Speedup matrix completion with side information:
  Application to multi-label learning,'' in \emph{NIPS}, 2013, pp. 2301--2309.

\bibitem{DBLP:conf/nips/CisseUAG13}
M.~Ciss{\'{e}}, N.~Usunier, T.~Arti{\`{e}}res, and P.~Gallinari, ``Robust bloom
  filters for large multilabel classification tasks,'' in \emph{NIPS}, 2013,
  pp. 1851--1859.

\bibitem{DBLP:conf/icml/SibliniMK18}
W.~Siblini, F.~Meyer, and P.~Kuntz, ``Craftml, an efficient clustering-based
  random forest for extreme multi-label learning,'' in \emph{ICML}, 2018, pp.
  4671--4680.

\bibitem{DBLP:conf/icml/YenKYHKR18}
I.~E. Yen, S.~Kale, F.~X. Yu, D.~N. Holtmann{-}Rice, S.~Kumar, and
  P.~Ravikumar, ``Loss decomposition for fast learning in large output
  spaces,'' in \emph{ICML}, 2018, pp. 5626--5635.

\bibitem{DBLP:conf/icml/WehrmannCB18}
J.~Wehrmann, R.~Cerri, and R.~C. Barros, ``Hierarchical multi-label
  classification networks,'' in \emph{ICML}, 2018, pp. 5225--5234.

\bibitem{DBLP:conf/icml/SiZKMDH17}
S.~Si, H.~Zhang, S.~S. Keerthi, D.~Mahajan, I.~S. Dhillon, and C.~Hsieh,
  ``Gradient boosted decision trees for high dimensional sparse output,'' in
  \emph{ICML}, 2017, pp. 3182--3190.

\bibitem{DBLP:conf/icml/JainMR17}
V.~Jain, N.~Modhe, and P.~Rai, ``Scalable generative models for multi-label
  learning with missing labels,'' in \emph{ICML}, 2017, pp. 1636--1644.

\bibitem{DBLP:conf/icml/ZhangZ17}
T.~Zhang and Z.~Zhou, ``Multi-class optimal margin distribution machine,'' in
  \emph{ICML}, 2017, pp. 4063--4071.

\bibitem{DBLP:conf/icml/LiWPA16}
C.~Li, B.~Wang, V.~Pavlu, and J.~A. Aslam, ``Conditional bernoulli mixtures for
  multi-label classification,'' in \emph{ICML}, 2016, pp. 2482--2491.

\bibitem{DBLP:conf/icml/YenHRZD16}
I.~E. Yen, X.~Huang, P.~Ravikumar, K.~Zhong, and I.~S. Dhillon, ``Pd-sparse :
  {A} primal and dual sparse approach to extreme multiclass and multilabel
  classification,'' in \emph{ICML}, 2016, pp. 3069--3077.

\bibitem{DBLP:conf/icml/CisseAB16}
M.~Ciss{\'{e}}, M.~Al{-}Shedivat, and S.~Bengio, ``{ADIOS:} architectures deep
  in output space,'' in \emph{ICML}, 2016, pp. 2770--2779.

\bibitem{DBLP:conf/icml/ParkNZSD15}
D.~Park, J.~Neeman, J.~Zhang, S.~Sanghavi, and I.~S. Dhillon, ``Preference
  completion: Large-scale collaborative ranking from pairwise comparisons,'' in
  \emph{ICML}, 2015, pp. 1907--1916.

\bibitem{DBLP:conf/icml/Hernandez-Lobato15}
D.~Hern{\'{a}}ndez{-}Lobato, J.~M. Hern{\'{a}}ndez{-}Lobato, and Z.~Ghahramani,
  ``A probabilistic model for dirty multi-task feature selection,'' in
  \emph{ICML}, 2015, pp. 1073--1082.

\bibitem{DBLP:conf/icml/HuangWSLC15}
Z.~Huang, R.~Wang, S.~Shan, X.~Li, and X.~Chen, ``Log-euclidean metric learning
  on symmetric positive definite manifold with application to image set
  classification,'' in \emph{ICML}, 2015, pp. 720--729.

\bibitem{DBLP:conf/icml/Yu0KD14}
H.~Yu, P.~Jain, P.~Kar, and I.~S. Dhillon, ``Large-scale multi-label learning
  with missing labels,'' in \emph{ICML}, 2014, pp. 593--601.

\bibitem{DBLP:conf/icml/LinDH014}
Z.~Lin, G.~Ding, M.~Hu, and J.~Wang, ``Multi-label classification via
  feature-aware implicit label space encoding,'' in \emph{ICML}, 2014, pp.
  325--333.

\bibitem{DBLP:conf/icml/LiZ14}
Y.~Li and R.~S. Zemel, ``High order regularization for semi-supervised learning
  of structured output problems,'' in \emph{ICML}, 2014, pp. 1368--1376.

\bibitem{DBLP:conf/icml/BiK13}
W.~Bi and J.~T. Kwok, ``Efficient multi-label classification with many
  labels,'' in \emph{ICML}, 2013, pp. 405--413.

\bibitem{DBLP:conf/icml/KolmogorovT13}
R.~Takhanov and V.~Kolmogorov, ``Inference algorithms for pattern-based crfs on
  sequence data,'' in \emph{ICML}, 2013, pp. 145--153.

\bibitem{DBLP:conf/icml/XiaoG13}
M.~Xiao and Y.~Guo, ``Domain adaptation for sequence labeling tasks with a
  probabilistic language adaptation model,'' in \emph{ICML}, 2013, pp.
  293--301.

\bibitem{brighton2002advances}
H.~Brighton and C.~Mellish, ``Advances in instance selection for instance-based
  learning algorithms,'' \emph{Data mining and knowledge discovery}, vol.~6,
  no.~2, pp. 153--172, 2002.

\bibitem{DBLP:journals/cim/ZhaiOT14}
Y.~Zhai, Y.~Ong, and I.~W. Tsang, ``The emerging ?big dimensionality?''
  \emph{IEEE Computational Intelligence Magazine}, vol.~9, no.~3, pp. 14--26,
  2014.

\bibitem{DBLP:conf/nips/ReedAMTSL16}
S.~E. Reed, Z.~Akata, S.~Mohan, S.~Tenka, B.~Schiele, and H.~Lee, ``Learning
  what and where to draw,'' in \emph{NIPS}, 2016, pp. 217--225.

\bibitem{DBLP:conf/nips/GoodfellowPMXWOCB14}
I.~J. Goodfellow, J.~Pouget{-}Abadie, M.~Mirza, B.~Xu, D.~Warde{-}Farley,
  S.~Ozair, A.~C. Courville, and Y.~Bengio, ``Generative adversarial nets,'' in
  \emph{NIPS}, pp. 2672--2680.

\bibitem{DBLP:conf/iccv/ZhangXL17}
H.~Zhang, T.~Xu, and H.~Li, ``Stackgan: Text to photo-realistic image synthesis
  with stacked generative adversarial networks,'' in \emph{ICCV}, 2017, pp.
  5908--5916.

\bibitem{DBLP:journals/corr/abs-1802-09178}
Z.~Zhang, Y.~Xie, and L.~Yang, ``Photographic text-to-image synthesis with a
  hierarchically-nested adversarial network,'' \emph{CoRR}, vol.
  abs/1802.09178, 2018.

\bibitem{DBLP:journals/corr/abs-1801-07736}
W.~Fedus, I.~J. Goodfellow, and A.~M. Dai, ``Maskgan: Better text generation
  via filling in the {\_}{\_}{\_}{\_}{\_}{\_},'' \emph{CoRR}, vol.
  abs/1801.07736, 2018.

\bibitem{DBLP:conf/nips/ChenL12}
Y.~Chen and H.~Lin, ``Feature-aware label space dimension reduction for
  multi-label classification,'' in \emph{NIPS}, 2012, pp. 1538--1546.

\bibitem{DBLP:conf/nips/HsuKLZ09}
D.~J. Hsu, S.~Kakade, J.~Langford, and T.~Zhang, ``Multi-label prediction via
  compressed sensing,'' in \emph{NIPS}, 2009, pp. 772--780.

\bibitem{DBLP:journals/neco/TaiL12}
F.~Tai and H.~Lin, ``Multilabel classification with principal label space
  transformation,'' \emph{Neural Computation}, vol.~24, no.~9, pp. 2508--2542,
  2012.

\bibitem{DBLP:conf/nips/KapoorVJ12}
A.~Kapoor, R.~Viswanathan, and P.~Jain, ``Multilabel classification using
  bayesian compressed sensing,'' in \emph{NIPS}, 2012, pp. 2654--2662.

\bibitem{DBLP:conf/pkdd/MineiroK15}
P.~Mineiro and N.~Karampatziakis, ``Fast label embeddings via randomized linear
  algebra,'' in \emph{Joint European conference on machine learning and
  knowledge discovery in databases}, 2015, pp. 37--51.

\bibitem{DBLP:conf/icml/JerniteCS17}
Y.~Jernite, A.~Choromanska, and D.~Sontag, ``Simultaneous learning of trees and
  representations for extreme classification and density estimation,'' in
  \emph{ICML}, 2017, pp. 1665--1674.

\bibitem{DBLP:conf/sigir/LiuCWY17}
J.~Liu, W.~Chang, Y.~Wu, and Y.~Yang, ``Deep learning for extreme multi-label
  text classification,'' in \emph{SIGIR}, 2017, pp. 115--124.

\bibitem{DBLP:journals/pr/BoutellLSB04}
M.~R. Boutell, J.~Luo, X.~Shen, and C.~M. Brown, ``Learning multi-label scene
  classification,'' \emph{Pattern Recognition}, vol.~37, no.~9, pp. 1757--1771,
  2004.

\bibitem{DBLP:reference/dmkdh/2010}
O.~Maimon and L.~Rokach, Eds., \emph{Data Mining and Knowledge Discovery
  Handbook, 2nd ed}.\hskip 1em plus 0.5em minus 0.4em\relax Springer, 2010.

\bibitem{DBLP:conf/pkdd/ReadPHF09}
J.~Read, B.~Pfahringer, G.~Holmes, and E.~Frank, ``Classifier chains for
  multi-label classification,'' in \emph{ECML PKDD}, 2009, pp. 254--269.

\bibitem{DBLP:journals/ml/ReadPHF11}
J.~Read, B.~Pfahringer, G.~Holmes, and E.~Frank, ``Classifier chains for
  multi-label classification,'' \emph{Machine Learning}, vol.~85, no.~3, pp.
  333--359, 2011.

\bibitem{DBLP:journals/jmlr/TsochantaridisJHA05}
I.~Tsochantaridis, T.~Joachims, T.~Hofmann, and Y.~Altun, ``Large margin
  methods for structured and interdependent output variables,'' \emph{JMLR},
  vol.~6, pp. 1453--1484, 2005.

\bibitem{DBLP:conf/icml/LaffertyMP01}
J.~D. Lafferty, A.~McCallum, and F.~C.~N. Pereira, ``Conditional random fields:
  Probabilistic models for segmenting and labeling sequence data,'' in
  \emph{ICML}, 2001, pp. 282--289.

\bibitem{DBLP:conf/ecml/TsoumakasV07}
G.~Tsoumakas and I.~P. Vlahavas, ``Random \emph{k} -labelsets: An ensemble
  method for multilabel classification,'' in \emph{ECML}, 2007, pp. 406--417.

\bibitem{DBLP:conf/icml/DembczynskiCH10}
K.~Dembczynski, W.~Cheng, and E.~H{\"{u}}llermeier, ``Bayes optimal multilabel
  classification via probabilistic classifier chains,'' in \emph{ICML}, 2010,
  pp. 279--286.

\bibitem{DBLP:journals/ml/JoachimsFY09}
T.~Joachims, T.~Finley, and C.~J. Yu, ``Cutting-plane training of structural
  svms,'' \emph{Machine Learning}, vol.~77, no.~1, pp. 27--59, 2009.

\bibitem{DBLP:conf/bionlp/BakerK17}
S.~Baker and A.~Korhonen, ``Initializing neural networks for hierarchical
  multi-label text classification,'' in \emph{BioNLP}, 2017, pp. 307--315.

\bibitem{DBLP:conf/cvpr/ChenZDLR17}
S.~Chen, C.~Zhang, M.~Dong, J.~Le, and M.~Rao, ``Using ranking-cnn for age
  estimation,'' in \emph{CVPR}, 2017, pp. 742--751.

\bibitem{DBLP:conf/nips/SutskeverVL14}
I.~Sutskever, O.~Vinyals, and Q.~V. Le, ``Sequence to sequence learning with
  neural networks,'' in \emph{Advances in Neural Information Processing Systems
  27: Annual Conference on Neural Information Processing Systems 2014, December
  8-13 2014, Montreal, Quebec, Canada}, 2014, pp. 3104--3112.

\bibitem{DBLP:journals/ijon/SmithJ14}
C.~Smith and Y.~Jin, ``Evolutionary multi-objective generation of recurrent
  neural network ensembles for time series prediction,'' \emph{Neurocomputing},
  vol. 143, pp. 302--311, 2014.

\bibitem{DBLP:conf/acl/Och03}
F.~J. Och, ``Minimum error rate training in statistical machine translation,''
  in \emph{Association for Computational Linguistics}, 2003, pp. 160--167.

\bibitem{gao2011consistency}
W.~Gao and Z.-H. Zhou, ``On the consistency of multi-label learning,'' in
  \emph{Proceedings of the 24th annual conference on learning theory}, 2011,
  pp. 341--358.

\bibitem{DBLP:journals/jmlr/TewariB07}
A.~Tewari and P.~L. Bartlett, ``On the consistency of multiclass classification
  methods,'' \emph{JMLR}, vol.~8, pp. 1007--1025, 2007.

\bibitem{DBLP:conf/nips/McAllesterK11}
D.~A. McAllester and J.~Keshet, ``Generalization bounds and consistency for
  latent structural probit and ramp loss,'' in \emph{NIPS}, 2011, pp.
  2205--2212.

\bibitem{DBLP:conf/nips/TaskarGK03}
B.~Taskar, C.~Guestrin, and D.~Koller, ``Max-margin markov networks,'' in
  \emph{NIPS}, 2003, pp. 25--32.

\bibitem{DBLP:conf/sigir/YueFRJ07}
Y.~Yue, T.~Finley, F.~Radlinski, and T.~Joachims, ``A support vector method for
  optimizing average precision,'' in \emph{SIGIR}, 2007, pp. 271--278.

\bibitem{DBLP:conf/emnlp/Collins02}
M.~Collins, ``Discriminative training methods for hidden markov models: Theory
  and experiments with perceptron algorithms,'' in \emph{Empirical Methods in
  Natural Language Processing}, 2002.

\bibitem{DBLP:conf/icassp/PoveyKKRSV08}
D.~Povey, D.~Kanevsky, B.~Kingsbury, B.~Ramabhadran, G.~Saon, and
  K.~Visweswariah, ``Boosted {MMI} for model and feature-space discriminative
  training,'' in \emph{ICASSP}, 2008, pp. 4057--4060.

\bibitem{DBLP:conf/naacl/GimpelS10}
K.~Gimpel and N.~A. Smith, ``Softmax-margin crfs: Training log-linear models
  with cost functions,'' in \emph{HLT-NAACL}, 2010, pp. 733--736.

\bibitem{8263148}
W.~Liu, D.~Xu, I.~Tsang, and W.~Zhang, ``Metric learning for multi-output
  tasks,'' \emph{TPAMI}, 2018, \mbox{doi}:\url{10.1109/TPAMI.2018.2794976}.

\bibitem{DBLP:conf/nips/DengSBL11}
J.~Deng, S.~Satheesh, A.~C. Berg, and F.~Li, ``Fast and balanced: Efficient
  label tree learning for large scale object recognition,'' in \emph{NIPS},
  2011, pp. 567--575.

\bibitem{DBLP:conf/iccv/GaoK11}
T.~Gao and D.~Koller, ``Discriminative learning of relaxed hierarchy for
  large-scale visual recognition,'' in \emph{ICCV}, 2011, pp. 2072--2079.

\bibitem{DBLP:conf/aaai/0001LTSO18}
X.~Shen, W.~Liu, I.~W. Tsang, Q.~Sun, and Y.~Ong, ``Compact multi-label
  learning,'' in \emph{AAAI}, 2018, pp. 4066--4073.

\bibitem{DBLP:journals/tnn/ShenLTSO18}
X.~Shen, W.~Liu, I.~W. Tsang, Q.~Sun, and Y.~Ong, ``Multilabel prediction via
  cross-view search,'' \emph{TNNLS}, vol.~29, no.~9, pp. 4324--4338, 2018.

\bibitem{DBLP:conf/ijcai/0001LLOT18}
X.~Shen, W.~Liu, Y.~Luo, Y.~Ong, and I.~W. Tsang, ``Deep discrete prototype
  multilabel learning,'' in \emph{IJCAI}, 2018, pp. 2675--2681.

\bibitem{more2016survey}
A.~More, ``Survey of resampling techniques for improving classification
  performance in unbalanced datasets,'' \emph{arXiv preprint arXiv:1608.06048},
  2016.

\bibitem{mani2003knn}
I.~Mani and I.~Zhang, ``knn approach to unbalanced data distributions: a case
  study involving information extraction,'' in \emph{Proceedings of workshop on
  learning from imbalanced datasets}, vol. 126, 2003.

\bibitem{DBLP:journals/jair/ChawlaBHK02}
N.~V. Chawla, K.~W. Bowyer, L.~O. Hall, and W.~P. Kegelmeyer, ``{SMOTE:}
  synthetic minority over-sampling technique,'' \emph{JAIR}, vol.~16, pp.
  321--357, 2002.

\bibitem{DBLP:journals/corr/abs-1804-10851}
Q.~Dong, S.~Gong, and X.~Zhu, ``Imbalanced deep learning by minority class
  incremental rectification,'' \emph{CoRR}, vol. abs/1804.10851, 2018.

\bibitem{DBLP:journals/corr/abs-1811-10419}
M.~Rezaei, H.~Yang, and C.~Meinel, ``Multi-task generative adversarial network
  for handling imbalanced clinical data,'' \emph{CoRR}, vol. abs/1811.10419,
  2018.

\bibitem{DBLP:journals/corr/abs-1811-08812}
E.~Montahaei, M.~Ghorbani, M.~S. Baghshah, and H.~R. Rabiee, ``Adversarial
  classifier for imbalanced problems,'' \emph{CoRR}, vol. abs/1811.08812, 2018.

\bibitem{DBLP:conf/icml/Romera-ParedesT15}
B.~Romera{-}Paredes and P.~H.~S. Torr, ``An embarrassingly simple approach to
  zero-shot learning,'' in \emph{ICML}, 2015, pp. 2152--2161.

\bibitem{gaure2017probabilistic}
A.~Gaure, A.~Gupta, V.~K. Verma, and P.~Rai, ``A probabilistic framework for
  zero-shot multi-label learning,'' in \emph{The Conference on Uncertainty in
  Artificial Intelligence (UAI)}, vol.~1, 2017, p.~3.

\bibitem{DBLP:conf/emnlp/RiosK18}
A.~Rios and R.~Kavuluru, ``Few-shot and zero-shot multi-label learning for
  structured label spaces,'' in \emph{Conference on Empirical Methods in
  Natural Language Processing}, 2018, pp. 3132--3142.

\bibitem{DBLP:conf/cvpr/LeeFYW18}
C.~Lee, W.~Fang, C.~Yeh, and Y.~F. Wang, ``Multi-label zero-shot learning with
  structured knowledge graphs,'' in \emph{CVPR}, 2018, pp. 1576--1585.

\bibitem{DBLP:conf/cvpr/GanYG16}
C.~Gan, T.~Yang, and B.~Gong, ``Learning attributes equals multi-source domain
  generalization,'' in \emph{CVPR}, 2016, pp. 87--97.

\bibitem{DBLP:conf/cvpr/LiuKS11}
J.~Liu, B.~Kuipers, and S.~Savarese, ``Recognizing human actions by
  attributes,'' in \emph{CVPR}, 2011, pp. 3337--3344.

\bibitem{DBLP:journals/paa/ZhangWXZL15}
Z.~Zhang, C.~Wang, B.~Xiao, W.~Zhou, and S.~Liu, ``Robust relative attributes
  for human action recognition,'' \emph{Pattern Analysis and Applications},
  vol.~18, no.~1, pp. 157--171, 2015.

\bibitem{DBLP:conf/iccv/JainGMS15}
M.~Jain, J.~C. van Gemert, T.~Mensink, and C.~G.~M. Snoek, ``Objects2action:
  Classifying and localizing actions without any video example,'' in
  \emph{ICCV}, 2015, pp. 4588--4596.

\bibitem{DBLP:conf/iccv/MettesS17}
P.~Mettes and C.~G.~M. Snoek, ``Spatial-aware object embeddings for zero-shot
  localization and classification of actions,'' in \emph{ICCV}, 2017, pp.
  4453--4462.

\bibitem{DBLP:journals/pami/ScheirerJB14}
W.~J. Scheirer, L.~P. Jain, and T.~E. Boult, ``Probability models for open set
  recognition,'' \emph{TPAMI}, vol.~36, no.~11, pp. 2317--2324, 2014.

\bibitem{DBLP:conf/eccv/JainSB14}
L.~P. Jain, W.~J. Scheirer, and T.~E. Boult, ``Multi-class open set recognition
  using probability of inclusion,'' in \emph{ECCV}, 2014, pp. 393--409.

\bibitem{DBLP:conf/cvpr/BendaleB16}
A.~Bendale and T.~E. Boult, ``Towards open set deep networks,'' in \emph{CVPR},
  2016, pp. 1563--1572.

\bibitem{DBLP:conf/iccv/LiYSCLL17}
Y.~Li, J.~Yang, Y.~Song, L.~Cao, J.~Luo, and L.~Li, ``Learning from noisy
  labels with distillation,'' in \emph{ICCV}, 2017, pp. 1928--1936.

\bibitem{DBLP:conf/sdm/ChenSWZ08}
G.~Chen, Y.~Song, F.~Wang, and C.~Zhang, ``Semi-supervised multi-label learning
  by solving a sylvester equation,'' in \emph{ICDM}, 2008, pp. 410--419.

\bibitem{DBLP:conf/aaai/SunZZ10}
Y.~Sun, Y.~Zhang, and Z.~Zhou, ``Multi-label learning with weak label,'' in
  \emph{AAAI}, 2010.

\bibitem{DBLP:conf/cvpr/BucakJJ11}
S.~S. Bucak, R.~Jin, and A.~K. Jain, ``Multi-label learning with incomplete
  class assignments,'' in \emph{CVPR}, 2011, pp. 2801--2808.

\bibitem{DBLP:conf/nips/CabralTCB11}
R.~S. Cabral, F.~D. la~Torre, J.~P. Costeira, and A.~Bernardino, ``Matrix
  completion for multi-label image classification,'' in \emph{NIPS}, 2011, pp.
  190--198.

\bibitem{DBLP:conf/aaai/BiK14}
W.~Bi and J.~T. Kwok, ``Multilabel classification with label correlations and
  missing labels,'' in \emph{AAAI}, 2014, pp. 1680--1686.

\bibitem{DBLP:conf/eccv/YangZC16}
H.~Yang, J.~T. Zhou, and J.~Cai, ``Improving multi-label learning with missing
  labels by structured semantic correlations,'' in \emph{ECCV}, 2016, pp.
  835--851.

\bibitem{DBLP:conf/icdm/GongZYT17}
C.~Gong, H.~Zhang, J.~Yang, and D.~Tao, ``Learning with inadequate and
  incorrect supervision,'' in \emph{ICDM}, 2017, pp. 889--894.

\bibitem{DBLP:conf/sspr/BarandelaG00}
R.~Barandela and E.~Gasca, ``Decontamination of training samples for supervised
  pattern recognition methods,'' in \emph{Joint IAPR International Workshops on
  SPR and SSPR}, 2000, pp. 621--630.

\bibitem{DBLP:journals/jair/BrodleyF99}
C.~E. Brodley and M.~A. Friedl, ``Identifying mislabeled training data,''
  \emph{JAIR}, vol.~11, pp. 131--167, 1999.

\bibitem{DBLP:conf/eccv/JoulinMJV16}
A.~Joulin, L.~van~der Maaten, A.~Jabri, and N.~Vasilache, ``Learning visual
  features from large weakly supervised data,'' in \emph{ECCV}, 2016, pp.
  67--84.

\bibitem{DBLP:journals/corr/ReedLASER14}
S.~E. Reed, H.~Lee, D.~Anguelov, C.~Szegedy, D.~Erhan, and A.~Rabinovich,
  ``Training deep neural networks on noisy labels with bootstrapping,''
  \emph{CoRR}, vol. abs/1412.6596, 2014.

\bibitem{DBLP:conf/cvpr/VeitACKGB17}
A.~Veit, N.~Alldrin, G.~Chechik, I.~Krasin, A.~Gupta, and S.~J. Belongie,
  ``Learning from noisy large-scale datasets with minimal supervision,'' in
  \emph{CVPR}, 2017, pp. 6575--6583.

\bibitem{DBLP:conf/icml/MnihH12}
V.~Mnih and G.~E. Hinton, ``Learning to label aerial images from noisy data,''
  in \emph{ICML}, 2012.

\bibitem{DBLP:conf/icdm/JindalNC16}
I.~Jindal, M.~S. Nokleby, and X.~Chen, ``Learning deep networks from noisy
  labels with dropout regularization,'' in \emph{ICDM}, 2016, pp. 967--972.

\bibitem{DBLP:journals/corr/abs-1711-00583}
J.~Yao, J.~Wang, I.~W. Tsang, Y.~Zhang, J.~Sun, C.~Zhang, and R.~Zhang, ``Deep
  learning from noisy image labels with quality embedding,'' \emph{CoRR}, vol.
  abs/1711.00583, 2017.

\bibitem{DBLP:conf/apsipa/MaoCLJ16}
Y.~Mao, G.~Cheung, C.~Lin, and Y.~Ji, ``Joint learning of similarity graph and
  image classifier from partial labels,'' in \emph{Asia-Pacific Signal and
  Information Processing Association Annual Summit and Conference}, 2016, pp.
  1--4.

\bibitem{chai2019tnn}
J.~Chai, I.~W. Tsang, and W.-J. Chen, ``Large margin partial label machine,''
  \emph{TNNLS}, p. to appear.

\bibitem{DBLP:journals/tcyb/GongLTYYT18}
C.~Gong, T.~Liu, Y.~Tang, J.~Yang, J.~Yang, and D.~Tao, ``A regularization
  approach for instance-based superset label learning,'' \emph{TCYB}, vol.~48,
  no.~3, pp. 967--978, 2018.

\bibitem{DBLP:journals/ml/YuZ17}
F.~Yu and M.~Zhang, ``Maximum margin partial label learning,'' \emph{Machine
  Learning}, vol. 106, no.~4, pp. 573--593, 2017.

\bibitem{DBLP:journals/tkde/ZhangYT17}
M.~Zhang, F.~Yu, and C.~Tang, ``Disambiguation-free partial label learning,''
  \emph{TKDE}, vol.~29, no.~10, pp. 2155--2167, 2017.

\bibitem{DBLP:conf/aaai/XieH18}
M.~Xie and S.~Huang, ``Partial multi-label learning,'' in \emph{AAAI}, 2018,
  pp. 4302--4309.

\bibitem{DBLP:conf/cvpr/NamH16}
H.~Nam and B.~Han, ``Learning multi-domain convolutional neural networks for
  visual tracking,'' in \emph{CVPR}, 2016, pp. 4293--4302.

\bibitem{avidan2007ensemble}
S.~Avidan, ``Ensemble tracking,'' \emph{TPAMI}, vol.~29, no.~2, 2007.

\bibitem{DBLP:conf/acml/QuZZQ09}
W.~Qu, Y.~Zhang, J.~Zhu, and Q.~Qiu, ``Mining multi-label concept-drifting data
  streams using dynamic classifier ensemble,'' in \emph{ACML}, 2009, pp.
  308--321.

\bibitem{DBLP:conf/kdd/BifetHPKG09}
A.~Bifet, G.~Holmes, B.~Pfahringer, R.~Kirkby, and R.~Gavald{\`{a}}, ``New
  ensemble methods for evolving data streams,'' in \emph{KDD}, 2009, pp.
  139--148.

\bibitem{DBLP:conf/colcom/KongY11}
X.~Kong and P.~S. Yu, ``An ensemble-based approach to fast classification of
  multi-label data streams,'' in \emph{International Conference on
  Collaborative Computing: Networking, Applications and Worksharing}, 2011, pp.
  95--104.

\bibitem{DBLP:conf/cikm/BuyukcakirBC18}
A.~B{\"{u}}y{\"{u}}k{\c{c}}akir, H.~R. Bonab, and F.~Can, ``A novel online
  stacked ensemble for multi-label stream classification,'' in \emph{CIKM},
  2018, pp. 1063--1072.

\bibitem{DBLP:conf/aaai/MilanRD0S17}
A.~Milan, S.~H. Rezatofighi, A.~R. Dick, I.~D. Reid, and K.~Schindler, ``Online
  multi-target tracking using recurrent neural networks,'' in \emph{AAAI},
  2017, pp. 4225--4232.

\bibitem{DBLP:journals/ml/ReadBHP12}
J.~Read, A.~Bifet, G.~Holmes, and B.~Pfahringer, ``Scalable and efficient
  multi-label classification for evolving data streams,'' \emph{Machine
  Learning}, vol.~88, no. 1-2, pp. 243--272, 2012.

\bibitem{DBLP:journals/tnn/HuangYZ18}
L.~Huang, Q.~Yang, and W.~Zheng, ``Online hashing,'' \emph{TNNLS}, vol.~29,
  no.~6, pp. 2309--2322, 2018.

\bibitem{DBLP:journals/tkde/XuTZ18}
D.~Xu, I.~W. Tsang, and Y.~Zhang, ``Online product quantization,'' \emph{TKDE},
  vol.~30, no.~11, pp. 2185 -- 2198, 2018.

\bibitem{DBLP:conf/cvpr/WangLMBZSX18}
Y.~Wang, W.~Liu, X.~Ma, J.~Bailey, H.~Zha, L.~Song, and S.~Xia, ``Iterative
  learning with open-set noisy labels,'' in \emph{Computer Vision and Pattern
  Recognition}, 2018, pp. 8688--8696.

\bibitem{DBLP:conf/icml/AnavaHZ15}
O.~Anava, E.~Hazan, and A.~Zeevi, ``Online time series prediction with missing
  data,'' in \emph{ICML}, 2015, pp. 2191--2199.

\bibitem{DBLP:conf/cvpr/BendaleB15}
A.~Bendale and T.~E. Boult, ``Towards open world recognition,'' in \emph{CVPR},
  2015, pp. 1893--1902.

\bibitem{DBLP:conf/ijcai/XioufisSTV11}
E.~S. Xioufis, M.~Spiliopoulou, G.~Tsoumakas, and I.~P. Vlahavas, ``Dealing
  with concept drift and class imbalance in multi-label stream
  classification,'' in \emph{IJCAI}, 2011, pp. 1583--1588.

\bibitem{DBLP:conf/sigir/RenPLDR14}
Z.~Ren, M.~Peetz, S.~Liang, W.~van Dolen, and M.~de~Rijke, ``Hierarchical
  multi-label classification of social text streams,'' in \emph{SIGIR}, 2014,
  pp. 213--222.

\bibitem{DBLP:conf/aaai/ShiXPTP19}
Y.~Shi, D.~Xu, Y.~Pan, I.~W. Tsang, and S.~Pan, ``Label embedding with partial
  heterogeneous contexts,'' in \emph{AAAI}, 2019, pp. 4926--4933.

\bibitem{DBLP:conf/3dim/MousavianPK16}
A.~Mousavian, H.~Pirsiavash, and J.~Kosecka, ``Joint semantic segmentation and
  depth estimation with deep convolutional networks,'' in \emph{3DV}, 2016, pp.
  611--619.

\bibitem{van2018fast}
W.~Van~Ranst, F.~De~Smedt, J.~Berte, T.~Goedem{\'e}, and T.-Z. RoboVision,
  ``Fast simultaneous people detection and re-identification in a single shot
  network,'' in \emph{AVSS}, 2018.

\end{thebibliography}
% argument is your BibTeX string definitions and bibliography database(s)
% Generated by IEEEtran.bst, version: 1.13 (2008/09/30)

\vspace{-0.5in}

\begin{IEEEbiography}[{\includegraphics[width=1in,height=1.25in,clip,keepaspectratio]{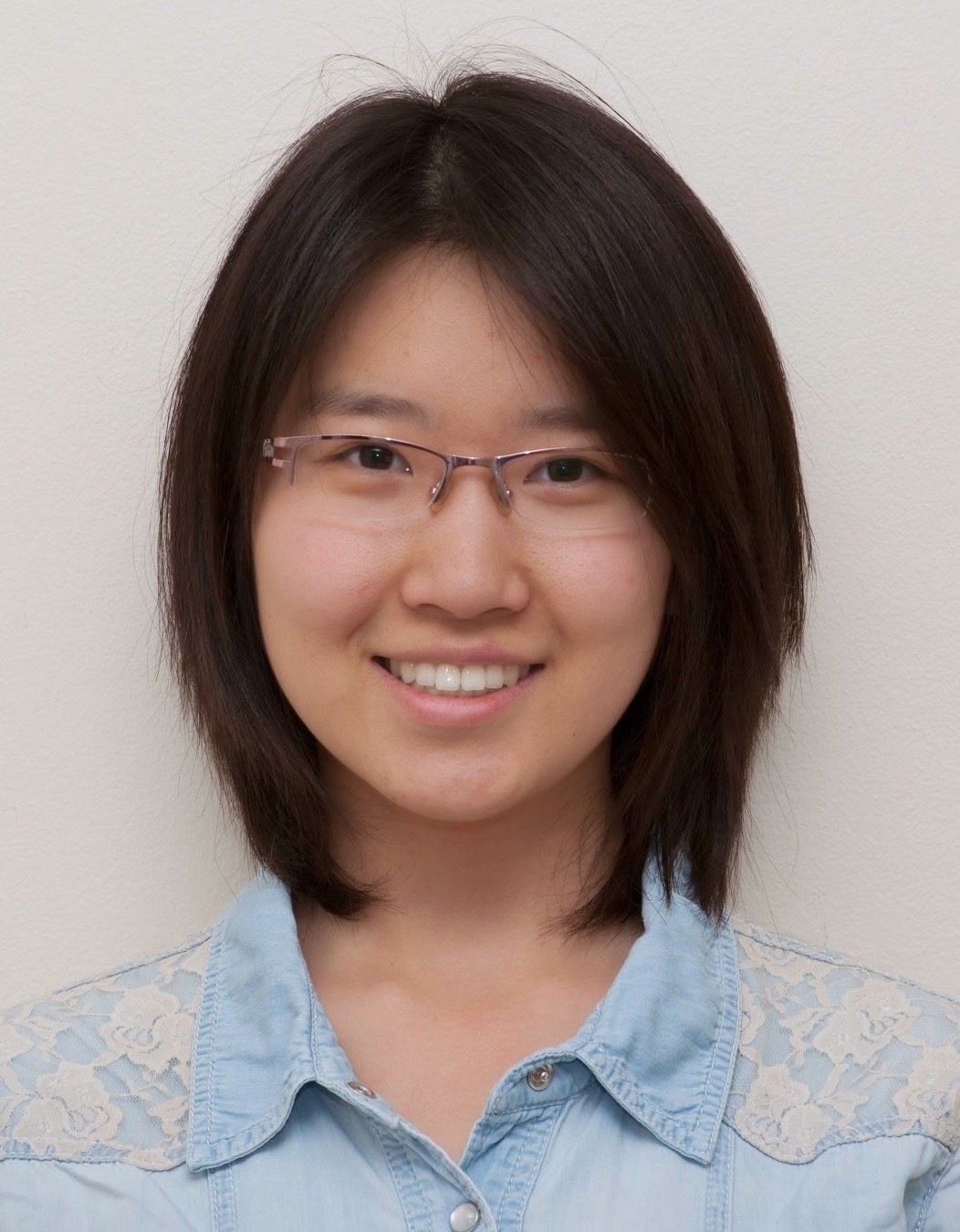}}]{Donna Xu} received the BCST (Honours) in computer science from the University of Sydney in 2014, and the PhD degree from the Centre for Artificial Intelligence, FEIT, University of Technology Sydney, NSW, Australia. Her research interests include multiclass classification, online hashing and information retrieval. 
\end{IEEEbiography}
\vspace{-0.5in}

 \begin{IEEEbiography}[{\includegraphics[width=1in,height=1.25in,clip,keepaspectratio]{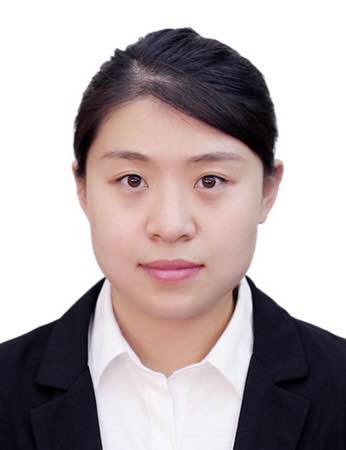}}]{Yaxin Shi} received her M.E.degree in computer science from Ocean University of China in 2017. She is currently pursuing a Ph.D.degree under the supervision of Prof. Ivor W. Tsang at the Centre for Artificial Intelligence, University of Technology Sydney, Australia. Her research interests include including multi-view learning, structure learning and deep generative networks.  
\end{IEEEbiography}

\vspace{-0.5in}

\begin{IEEEbiography}[{\includegraphics[width=1in,height=1.25in,clip,keepaspectratio]{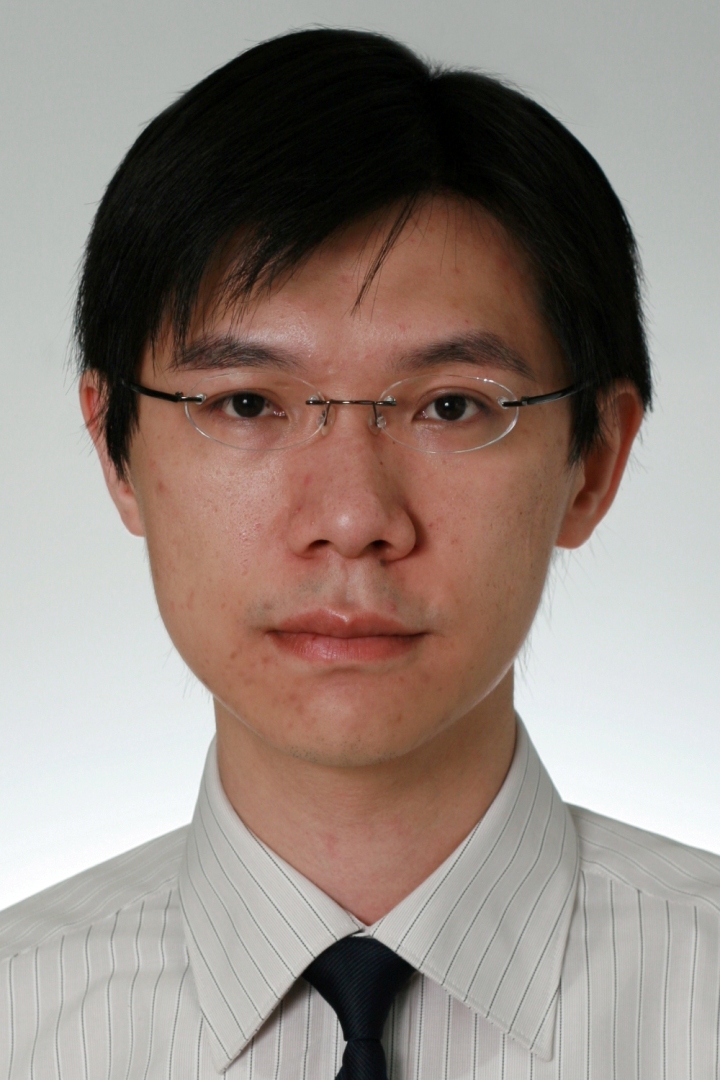}}]{Ivor W. Tsang} received his PhD degree from the Hong Kong University of Science and Technology in 2007. He is a professor with the University of Technology Sydney. He is also the research director of the UTS Priority Research Centre for Artificial Intelligence. He was conferred the 2008 Natural Science Award (Class II), Australian Research Council Future Fellowship in 2013, IEEE Transactions on Neural Networks Outstanding 2004 Paper Award in 2007, the 2014 IEEE Transactions on Multimedia Prize Paper Award, and the Best Student Paper Award at CVPR 2010. He serves as AE for IEEE Transactions on Emerging Topics in Computational Intelligence, IEEE Transactions on Big Data and Neurocomputing. He also serves as Area Chair/SPC for NeurIPS, AAAI and IJCAI.
\end{IEEEbiography}

\vspace{-0.5in}

\begin{IEEEbiography}[{\includegraphics[width=1in,height=1.25in,clip,keepaspectratio]{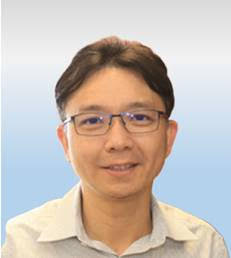}}]{Yew Soon Ong} received the PhD degree for his work on Artificial Intelligence in complex design from the University of Southampton, UK in 2003. He is a President Chair Professor of Computer Science at the Nanyang Technological University (NTU), and holds the position of Chief Artificial Intelligence Scientist at the Agency for Science, Technology and Research in Singapore. At NTU, he serves as Director of the Singtel-NTU Cognitive \& Artificial Intelligence Joint Lab. His research interest lies in artificial and computational intelligence. He is the founding Editor-in-Chief of the IEEE Transactions on Emerging Topics in Computational Intelligence and AE of IEEE Transactions on Neural Networks \& Learning Systems, the IEEE Transactions on Cybernetics, and others.
\end{IEEEbiography}

\vspace{-0.5in}

\begin{IEEEbiography}[{\includegraphics[width=1in,height=1.25in,clip,keepaspectratio]{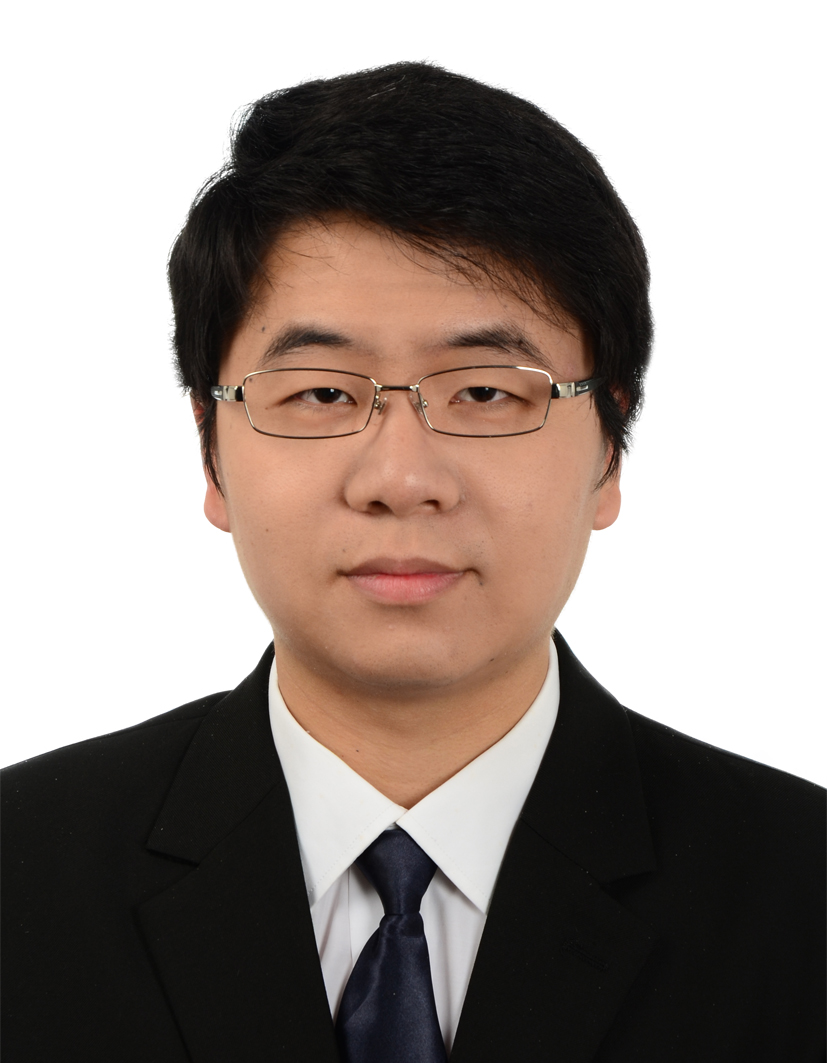}}]{Chen Gong} (M'16) received his dual doctoral degree from Shanghai Jiao Tong University (SJTU) and University of Technology Sydney (UTS) in 2016 and 2017, respectively. Currently, he is a full professor in the School of Computer Science and Engineering, Nanjing University of Science and Technology. His research interests mainly include machine learning, data mining, and learning-based vision problems. He has published more than 60 technical papers at prominent journals and conferences such as IEEE T-PAMI, IEEE T-NNLS, IEEE T-IP, IEEE T-CYB, IEEE T-CSVT, IEEE T-MM, IEEE T-ITS, ACM T-IST, NeurIPS, CVPR, AAAI, IJCAI, ICDM, etc. He also serves as the reviewer for more than 20 international journals such as AIJ, IEEE T-PAMI, IEEE T-NNLS, IEEE T-IP, and also the SPC/PC member of several top-tier conferences such as ICML, NeurIPS, AAAI, IJCAI, ICDM, AISTATS, etc. He received the “Excellent Doctorial Dissertation” awarded by Shanghai Jiao Tong University (SJTU) and Chinese Association for Artificial Intelligence (CAAI). He was also enrolled by the “Young Elite Scientists Sponsorship Program” of Jiangsu Province and China Association for Science and Technology.
\end{IEEEbiography}

\vspace{-0.5in}

\begin{IEEEbiography}[{\includegraphics[width=1in,height=1.25in,clip,keepaspectratio]{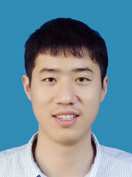}}]{Xiaobo Shen} received his BSc and PhD from School of Computer Science and Engineering, Nanjing University of Science and Technology in 2011 and 2017 respectively. He is currently a Professor with the School of Computer Science and Engineering, Nanjing University of Science and Technology, China. He has authored over 30 technical papers in prominent journals and conferences, such as IEEE TNNLS, IEEE TIP, IEEE TCYB, NIPS, ICML, ACM MM, AAAI, and IJCAI. His primary research interests are Multi-view Learning, Multi-label Learning, Network Embedding and Hashing.
\end{IEEEbiography}

% You can push biographies down or up by placing
% a \vfill before or after them. The appropriate
% use of \vfill depends on what kind of text is
% on the last page and whether or not the columns
% are being equalized.

%\vfill

% Can be used to pull up biographies so that the bottom of the last one
% is flush with the other column.
%\enlargethispage{-5in}

% that's all folks
\end{document}